\documentclass[12pt, twoside, openright]{thesis}

\usepackage[T1]{fontenc}
\usepackage{lmodern}

\usepackage[acronym, nonumberlist]{glossaries}
\makeglossaries

\newacronym{rl}{RL}{Reinforcement Learning}
\newacronym{mdp}{MDP}{Markov Decision Process}
\newacronym{ea}{EA}{Evolutionary Algorithm}
\newacronym{ns}{NS}{Novelty Search}
\newacronym{bs}{BS}{Behaviour Space}
\newacronym{qd}{QD}{Quality-Diversity}
\newacronym{ae}{AE}{autoencoder}
\newacronym{taxons}{TAXONS}{Task Agnostic eXploration of Outcome space through Novelty and Surprise}
\newacronym{dmp}{DMP}{Dynamic Movement Primitive}
\newacronym{10d}{10D}{10-dimensional}
\newacronym{cnn}{CNN}{Convolutional Neural Network}
\newacronym{bc}{BC}{Behaviour characterization}
\newacronym{vae}{VAE}{Variational Autoencoder}
\newacronym{me}{ME}{MAP-Elites}
\newacronym{nn}{NN}{neural network}
\newacronym{serene}{SERENE}{SparsE Reward Exploration via Novelty and Emitters}
\newacronym{dof}{DoF}{degrees of freedom}
\newacronym{nslc}{NSLC}{Novelty Search with Local Competition}
\newacronym{es}{ES}{Evolution Strategies}
\newacronym{lstm}{LSTM}{Long Short-Term Memory}
\newacronym{name}{STAX}{SERENE augmented TAXONS}
\newacronym{mab}{MAB}{Multi-Armed Bandit}
\newacronym{her}{HER}{Hindsight Experience Replay}
\newacronym{gan}{GAN}{Generative Adversarial Network}
\newacronym{im}{IM}{Intrinsic Motivation}
\newacronym{imgep}{IMGEP}{Intrinsically Motivated Goal Exploration Processes}
\newacronym{gep}{GEP}{Goal Exploration Processes}
\newacronym{rnd}{RND}{Random Network Distillation}
\newacronym{moo}{MOO}{Multi-objective optimization}

\usepackage[ruled]{algorithm2e}
\usepackage{caption}
\usepackage{wrapfig}
\usepackage[nice]{nicefrac}
\usepackage{amssymb}
\usepackage{breqn}
\usepackage{amsmath}
\usepackage{csquotes}
\usepackage{minitoc}
\usepackage{enumitem}
\usepackage{pdfpages} 
\usepackage{hhline}
\usepackage{hyphenat}

\newcommand{\centered}[1]{\begin{tabular}{@{}l@{}} #1 \end{tabular}}
\usepackage{array}
\usepackage{vcell}

\usepackage{xcolor}
\usepackage[draft]{changes}  
\definecolor{color-alban}{rgb}{0.0, 0.4, 0.9}
\definechangesauthor[name={Alban}, color=color-alban]{A}

\definecolor{color-giuseppe}{rgb}{0.8, 0.05, 0.05}
\definechangesauthor[name={Giuseppe}, color=color-giuseppe]{G}

\definecolor{color-stephane}{rgb}{0.9, 0.1, 0.1}
\definechangesauthor[name={Stephane}, color=color-stephane]{S}

\graphicspath{{./figures/}}

\unilogo{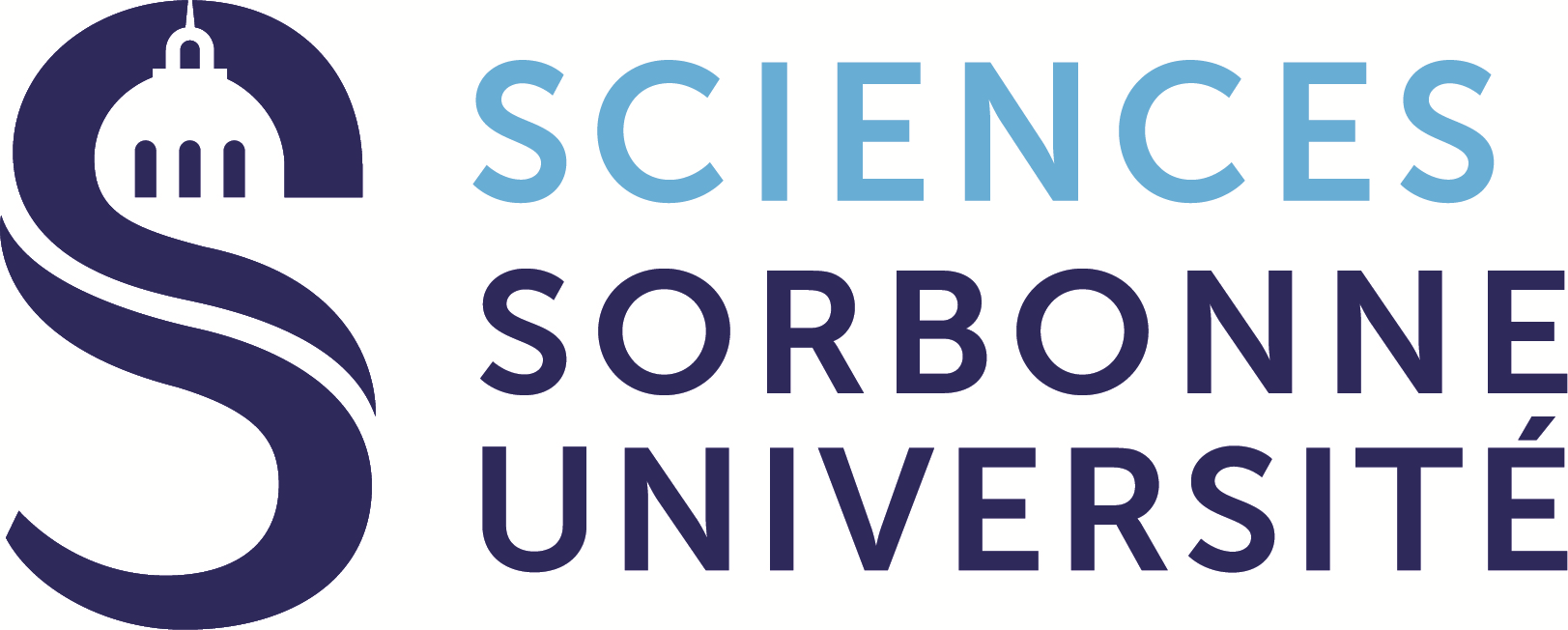}
\lablogo{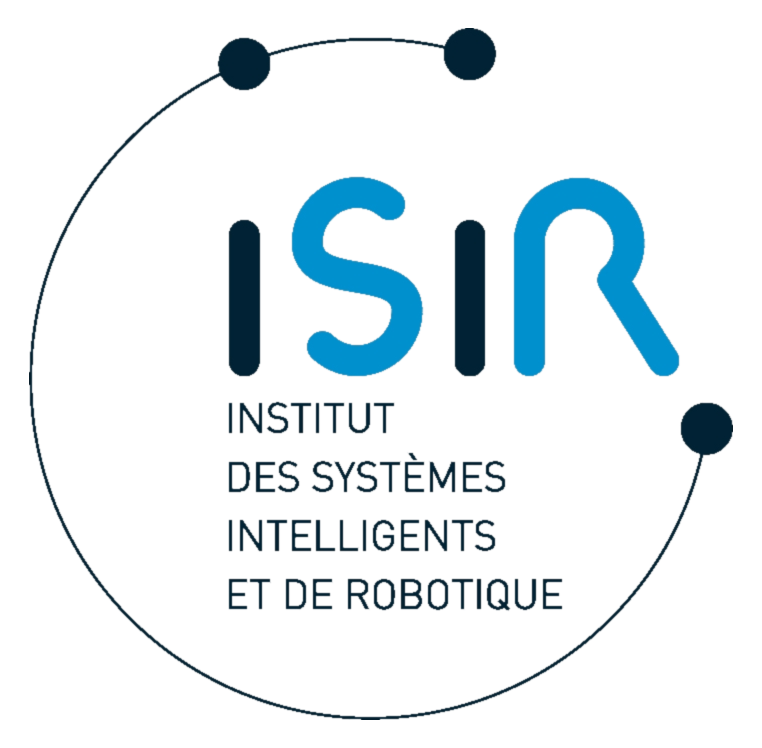}
\companylogo{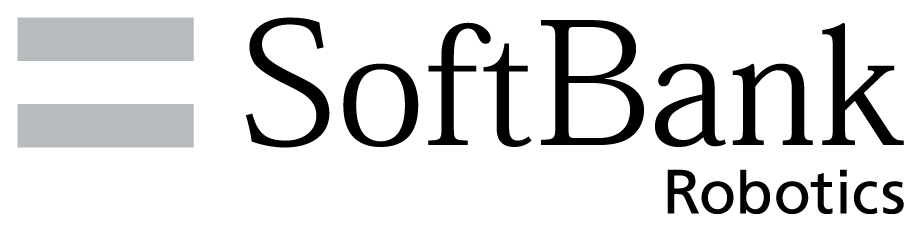}
\cnrslogo{figures/cnrs.jpg}
\title{Learning in Sparse Rewards settings through Quality Diversity algorithms}
\author{Giuseppe Paolo}
\university{Sorbonne Université}
\docschool{\'Ecole Doctorale Informatique, Télécommunications et \'Electronique}
\lab{Institut des Systèmes Intelligents et de Robotique}
\company{Softbank Robotics Europe}
\defensedate{16 January 2020}

\njurymembers{5}
\addjurymember{1}{M.}{Stéphane Doncieux}{Directeur de thèse}
\addjurymember{2}{M.}{Alban Laflaquière}{Coencadrant}
\addjurymember{3}{M.}{Alexander Coninx}{Coencadrant}
\addjurymember{4}{M.}{David Filliat}{Rapporteur}
\addjurymember{5}{M.}{Olivier Sigaud}{Rapporteur}

\begin{document}

\frontmatter
\begin{titlepage}
\includepdf{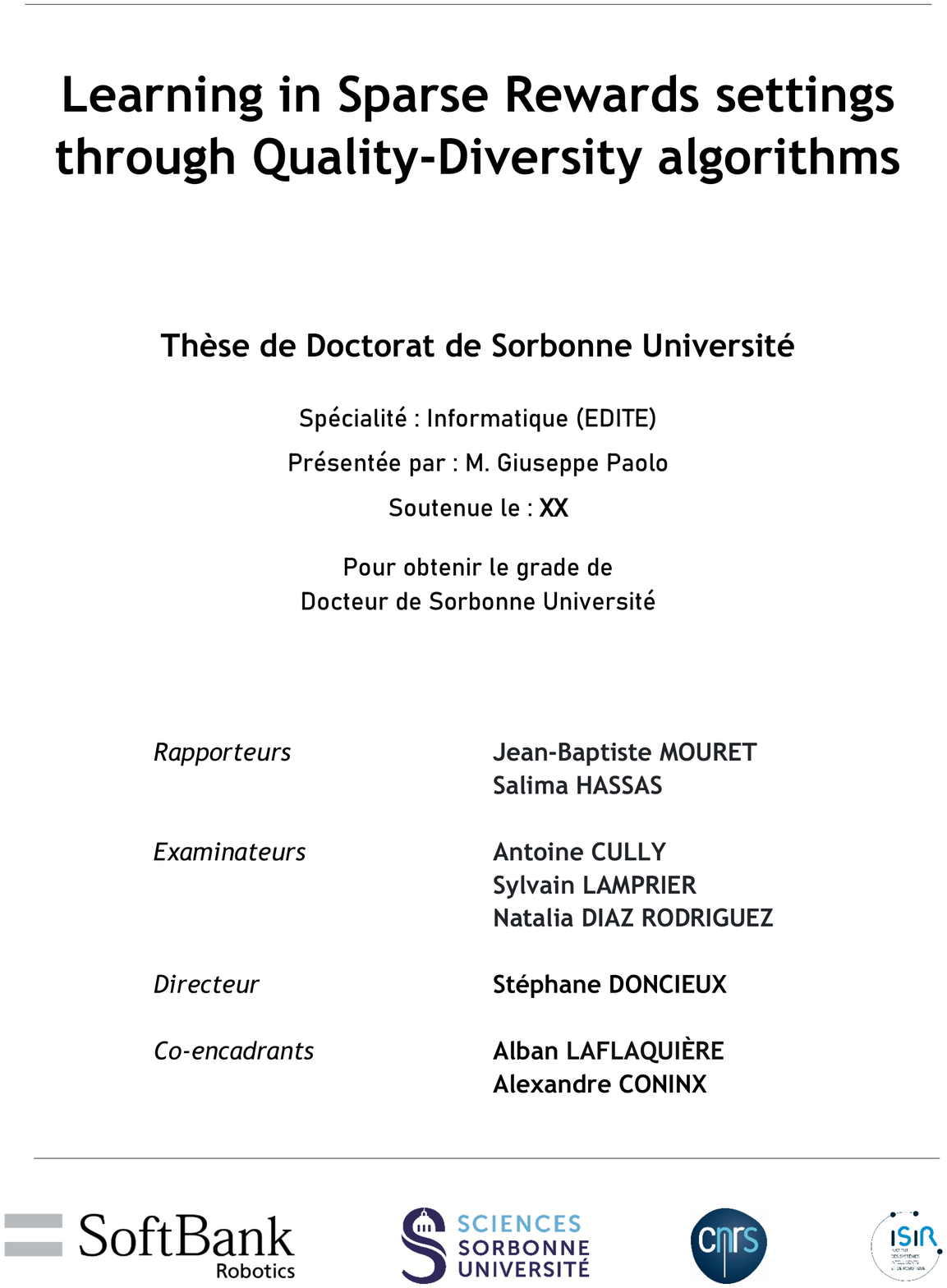}
\end{titlepage}
\dominitoc 

\cleardoublepage
\section*{\textsc{\huge English Abstract}}
\addstarredchapter{English Abstract}
Embodied agents, both natural and artificial, can learn to interact with the environment they are in through a process of trial and error.
This process can be formalized through the Reinforcement Learning framework, in which the agent performs an action in the environment and observes its outcome through an observation and a reward signal.
It is the reward signal that tells the agent how good the performed action is with respect to the task.
This means that the more often a reward is given, the easier it is to improve on the current solution.
When this is not the case, and the reward is given sparingly, the agent finds itself in a situation of sparse rewards.
This requires a big focus on exploration, that is on testing different things, in order to discover which action, or set of actions leads to the reward.
RL agents usually struggle with this.
Exploration is the focus of Quality-Diversity methods, a family of evolutionary algorithms that searches for a set of policies whose behaviors are as different as possible, while also improving on their performances.
In this thesis, we approach the problem of sparse rewards with these algorithms, and in particular with Novelty Search.
This is a method that, contrary to many other Quality-Diversity approaches, does not improve on the performances of the discovered rewards, but only on their diversity.
Thanks to this it can quickly explore the whole space of possible policies behaviors.

The first part of the thesis focuses on autonomously learning a representation of the search space in which the algorithm evaluates the discovered policies.
In this regard, we propose the \emph{Task Agnostic eXploration of Outcome spaces through Novelty and Surprise (TAXONS)} algorithm.
This method learns a low-dimensional representation of the search space in situations in which it is not easy to hand-design said representation.
TAXONS has proven effective in three different environments but still requires information on when to capture the observation used to learn the search space.
This limitation is addressed by performing a study on multiple ways to encode into the search space information about the whole trajectory of observations generated during a policy evaluation.
Among the studied methods, we analyze in particular the mathematical transform called \emph{signature} and its relevance to build trajectory-level representations.

The manuscript continues with the study of a complementary problem to the one addressed by TAXONS: how to focus on the most interesting parts of the search space.
Novelty Search is limited by the fact that all information about any reward discovered during the exploration process is ignored.
In our second contribution, we introduce the \emph{Sparse Reward Exploration via Novelty Search and Emitters (SERENE)} algorithm.
This method separates the exploration of the search space from the exploitation of the reward through a two-alternating-steps approach.
The exploration is performed through Novelty Search, but whenever a reward is discovered, it is exploited by instances of reward-based methods - called emitters - that perform local optimization of the reward.
Experiments on different environments show how SERENE can quickly obtain high rewarding solutions without hindering the exploration performances of the method.

In our third and final contribution, we combine the two ideas presented with TAXONS and SERENE into a single approach: \emph{SERENE augmented TAXONS (STAX)}.
This algorithm can autonomously learn a low-dimensional representation of the search space while quickly optimizing any discovered reward through emitters.
Experiments conducted on various environments show how the method can i) learn a representation allowing the discovery of all rewards and ii) quickly exploit those rewards thanks to the emitters.

Throughout this thesis, we introduce methods that, while dealing with sparse rewards situations, lower the amount of prior information needed at design time.
These contributions are a promising step towards the development of methods that can autonomously explore and find high-performance policies in a variety of sparse rewards settings.
This could increase the range of applicability of existing approaches leading to more autonomous embodied agents.

\cleardoublepage
\section*{\textsc{\huge French Abstract}}
\addstarredchapter{French Abstract}
Les agents incarnés, qu'ils soient naturels ou artificiels, peuvent apprendre à interagir avec l'environnement dans lequel ils se trouvent par un processus d'essais et d'erreurs.
Ce processus peut être formalisé dans le cadre de l'apprentissage par renforcement, dans lequel l'agent effectue une action dans l'environnement et observe son résultat par le biais d'une observation et d'un signal de récompense.
C'est le signal de récompense qui indique à l'agent la qualité de l'action effectuée par rapport à la tâche.
Cela signifie que plus une récompense est donnée, plus il est facile d'améliorer la solution actuelle.
Lorsque ce n'est pas le cas, et que la récompense est donnée avec parcimonie, l'agent se retrouve dans une situation de récompenses éparses.
Cela nécessite de se concentrer sur l'exploration, c'est-à-dire de tester différentes choses, afin de découvrir quelle action ou quel ensemble d'actions mène à la récompense.
Les agents RL ont généralement du mal à le faire.
L'exploration est le point central des méthodes de Qualité-Diversité, une famille d'algorithmes évolutionnaires qui recherche un ensemble de politiques dont les comportements sont aussi différents que possible, tout en améliorant leurs performances.
Dans cette thèse, nous abordons le problème des récompenses éparses avec ces algorithmes, et en particulier avec Novelty Search.
Il s'agit d'une méthode qui, contrairement à de nombreuses autres approches Qualité-Diversité, n'améliore pas les performances des récompenses découvertes, mais uniquement leur diversité.
Grâce à cela, elle peut explorer rapidement tout l'espace des comportements possibles des politiques.

La première partie de la thèse se concentre sur l'apprentissage autonome d'une représentation de l'espace de recherche dans lequel l'algorithme évalue les politiques découvertes.
A cet égard, nous proposons l'algorithme \emph{Task Agnostic eXploration of Outcome spaces through Novelty and Surprise (TAXONS)}.
Cette méthode apprend une représentation à faible dimension de l'espace de recherche dans des situations où il n'est pas facile de concevoir manuellement cette représentation.
TAXONS s'est avéré efficace dans trois environnements différents mais nécessite encore des informations sur le moment où il faut saisir l'observation utilisée pour apprendre l'espace de recherche.
Cette limitation est abordée en réalisant une étude sur les multiples façons d'encoder dans l'espace de recherche des informations sur la trajectoire complète des observations générées pendant une évaluation de politique.
Parmi les méthodes étudiées, nous analysons en particulier la transformation mathématique appelée \emph{signature} et sa pertinence pour construire des représentations au niveau de la trajectoire.

Le manuscrit se poursuit par l'étude d'un problème complémentaire à celui abordé par TAXONS : comment se concentrer sur les parties les plus intéressantes de l'espace de recherche.
Novelty Search est limitée par le fait que toute information sur une récompense découverte au cours du processus d'exploration est ignorée.
Dans notre deuxième contribution, nous présentons l'algorithme \emph{Sparse Reward Exploration via Novelty Search and Emitters (SERENE)}.
Cette méthode sépare l'exploration de l'espace de recherche de l'exploitation de la récompense par une approche en deux étapes alternées.
L'exploration est effectuée par Novelty Search, mais lorsqu'une récompense est découverte, elle est exploitée par des instances de méthodes basées sur la récompense - appelées émetteurs - qui effectuent une optimisation locale de la récompense.
Des expériences sur différents environnements montrent comment SERENE peut obtenir rapidement des solutions à forte récompense sans nuire aux performances d'exploration de la méthode.

Dans notre troisième et dernière contribution, nous combinons les deux idées présentées avec TAXONS et SERENE en une seule approche : \emph{TAXONS augmentés par SERENE (STAX)}.
Cet algorithme peut apprendre de manière autonome une représentation à faible dimension de l'espace de recherche tout en optimisant rapidement toute récompense découverte grâce à des émetteurs.
Des expériences menées sur différents environnements montrent comment la méthode peut i) apprendre une représentation permettant la découverte de toutes les récompenses et ii) exploiter rapidement ces récompenses grâce aux émetteurs.

Tout au long de cette thèse, nous introduisons des méthodes qui, tout en traitant des situations de récompenses éparses, réduisent la quantité d'informa-tions préalables nécessaires au moment de la conception.
Ces contributions constituent une étape prometteuse vers le développement de méthodes capables d'explorer et de trouver de manière autonome des politiques performantes dans une variété de situations de récompenses éparses.
Cela pourrait augmenter le champ d'application des approches existantes et conduire à des agents incarnés plus autonomes.
\clearpage\null\thispagestyle{empty}

\section*{\textsc{\huge Acknowledgement}}

Completing a Ph.D. thesis is a huge task and the whole process is a though but incredible journey.
Doing all of this just by myself would have been impossible, even more considering the situation created by COVID in these past years.
I want to thank here all the people that, with their help and support, made reaching the end of this journey possible.\\

First and foremost, I want to thank my supervisors.
They believed in me, giving me the opportunity to freely develop my research ideas.
If I am here as a young scientist, it is without any doubt thanks to them.\\

I am incredibly proud of having done this under the supervision of Stéphane Doncieux.
The support he gave me, and the patience shown in the most stressful moments, meant a lot.
His advice has always been incredibly helpful and allowed me to grow as a scientist while keeping me on track towards the final goal.\\

A great part of the supervision came also from Alban Laflaquière.
He proved to be a great manager and allowed me to have all the freedom I needed within SBRE while keeping a close eye on my work.
His suggestions have always been of the highest quality, and I really enjoyed our discussions on the most disparate topics.\\

A big thanks also to Alexandre Coninx, whose supervision and suggestions helped me to get out of the many “local minima” that a Ph.D. student finds in its path.\\

Both ISIR and SBRE provided me with the right working environment and the resources needed to perform my research. Being part of these two institutions allowed me to meet many great colleagues and friends. 

In particular, I want to thank the members of the AMAC equipe, the AI Lab, the Proto Lab, and the Expressivity team. The lunches, discussions, and activities we did together were a great part of the experience of my Ph.D. and I will always cherish these memories.

An important part of the life in the lab was also the open-ended learning group, whose immensely interesting and stimulating discussion never ceased to inspire me. I hope the spirit of discovery and discussion of the group will continue to inspire future PhDs and members of the lab.\\

I want to thank my family. Even if far away, and even if they did not always understand what I was doing, they never ceased to support me in any possible way. I wouldn’t be here if it were not for them.\\

Living abroad, in the fantastic city that is Paris, I also made another family: Ginevra, Sara, Maura, Michel, Laura, Carmine, Marwen, Hugo, Alessandro, Helena and all the amazing friends that made this experience incredible.
The list would be too long to mention everyone, but you’re all in my heart, and the time and “adventures” we had together will always be very fond memories for me.\\

Also thanks to Lisa, Désirée and Gabriele for the support (and the wifi hotspot) in the last moments of this manuscript redaction.\\

Finally, I want to thank the members of the Jury of my Ph.D. defense for accepting being here. Their opinion and judgment of my work will surely be a good conclusion of these past 3 years and help me in defining my future career.\\

Sometimes I feel this went all too fast.

And as someone wiser than me once said: “Dovrò soltanto reimparare a camminare”\\

\textbf{Thanks for everything!}
\clearpage\null\thispagestyle{empty}

\hypertarget{contents}{}
\addstarredchapter{Contents}
\tableofcontents

\printglossary[type=\acronymtype,title=Acronyms, toctitle=List of terms]
{%
\let\oldnumberline\numberline%
\renewcommand{\numberline}{\figurename~\oldnumberline}%
\listoffigures%
}


\mainmatter
\chapter{Introduction}
An embodied agent is any agent situated in an environment with which it interacts.
The natural world around us is full of this kind of agents: not only humans, but animals, plants, fungi, bacteria can all be considered embodied agents.
They all act and interact with the world they are in, the environment, by following some kind of policy.
This policy can be either innate, in which case it is usually referred as instinct \cite{guillaume1950manuel, eibl1984ethologie}, or learned during the life of the agent itself.
In both situations, the policy dictates which actions the agent should perform on the environment as a reaction to the state the environment or the agent itself are in. 
Up until very recently, the only existing type of embodied agents were biological beings, evolved through natural evolution in the enormous variety of living beings known today \cite{darwin1859origin}.
This is not the case anymore, thanks to the many research advancements that have lead to the development of artificial - albeit still rudimentary - embodied agents.

Humans have long dreamt of creating artificial agents capable of interacting with the world they are in.
The first accounts of these ideas date back to ancient Greece: in Homer's Iliad are described the creation of artificial agents by both gods and humans \cite{silk2004homer}.
Similar themes are present also in Chinese and Egyptian mythology \cite{needham1978shorter}.
The appearance of stories like this in cultures so distinct and far apart demonstrates how great for humans is the desire to create artificial agents.
A desire that continued through the Middle Ages and the Renaissance, when many "embodied agents" were built and displayed by scientists and engineers all around the world \cite{riskin2010genesis}.
These machines were known as \emph{automata}, a word coming from ancient Greek meaning "acting on one's own will", a name highlighting how they could interact with the world by following a policy, their "will".
Notwithstanding the promise given by the name, and the fascination these machines were provoking in people at the time, automata were nothing more than complex mechanism, much more similar to a clock than to an agent capable of reacting and adjusting to the state of the world.
The policy governing them was designed to perform only a limited set of actions in a way that could give the illusion of will.
Moreover, due to said policy being part of the hardware design, it could not be modified without rebuilding the whole machine.

Subsequent technological developments, an in particular the invention of the computer, allowed the creation of ever more sophisticated agents, capable of interacting with the world in multiple and better ways.
Agents of this kind are known today as \emph{robots}, a word coming from the Slavic languages expression for forced labor \cite{zunt2002did}.
Contrary to automata, the word robot stresses the fact that these are machines, deprived of any will, that just follow a set of instructions, the policy.
At the same time, robots have an important advantage on ancient automata in the fact that they have sensors.
This enables the creation of agents capable of dealing with a much bigger range of situations.
Thanks to this, the development of robots allowed the automation of many labor-intense tasks.
An example of this is the introduction of robots in assembly lines, warehouses and other environments requiring heavy and repetitive tasks, allowing the automation and the increase in production efficiency of these systems \cite{wallen2008history, singh2013evolution, hagele2016industrial}.
In the future, even more advantages for human societies are predicted to come thanks to robots \cite{wired2020robots}.
At the same time, these are very controlled and well defined systems, for which relatively simple policies can be hand-designed by the engineers.
This is not the case for more complex and difficult to control settings, for which the policy designer must take into account all possible interactions between the robot and the elements of the environment.
A robot performing inspection of an industrial plant \cite{kroll2008survey}, has to deal with uneven terrain, doors to open, objects to move or navigate around.
There is an almost infinite amount of different situations to deal with.
For this kind of problems, learning the controller policy is an effective alternative approach to hand-designing it.
This allows the agent to adapt its strategy to the task at hand, while moving the engineer's design effort on the training process.
The advantage of this approach is its generalizability: the same training process can be used to learn policies for different tasks.
There are multiple ways in which a policy can be learned, but they typically consist in algorithms optimizing a performance metric measuring how well the learned policy can accomplish the desired goal.
Algorithms of this kind belong to the field of Artificial Intelligence and, as many of the methods in this discipline, they take huge inspiration from natural processes and the way animals learn.
Depending on which natural mechanics they are based on, policy learning methods can be grouped in two families: \gls{rl} and \glspl{ea}.

\section*{Reinforcement Learning}
\gls{rl} is a framework that can be used for learning policies able to solve a given task through a process of "trial-and-error" \cite{sutton2018reinforcement}.
It is inspired by the way animals learn how to solve tasks: try something and according to the outcome, learn to repeat or to avoid the same action in similar situations \cite{ludvig2012reinforcement}.
The origin of \gls{rl} can in fact be traced back to Thorndike's Law of Effect \cite{thorndike1927law} and Pavlov's studies on conditioned reflexes \cite{pavlov1927conditioned}.
Both scientists studied how behaviors that were followed by positive outcomes were more likely to become established and be repeated in similar situations.
A famous experiment in this regard is Pavlov's dog study \cite{pavlov1927conditioned, britannica2008britannica}.
The experiment consisted in placing a dog in a room, in which some food is delivered every time a bell is rang.
After few of these interactions, the scientist observed that the dog's salivation would increase whenever the bell was rang, as if the animal was in presence of food, even if no food was delivered anymore.
This shows how the dog learned a behavior, the salivation, whenever a given conditioning, the sound of the bell, happened; even if no food was given anymore.
\gls{rl} algorithms imitate the same mechanism in order to learn a policy, rendering them extremely flexible on the range of problems they can solve \cite{mao2016resource, arel2010reinforcement, levine2016end, zhou2017optimizing}.
A fundamental component of any \gls{rl} system is the reward function, used to drive the training process towards learning a good policy. 
It is through this function that the engineer "communicates" to the agent the task it needs to solve.
This means that a great part of the design effort has to be directed towards the definition of that function.
In general, \gls{rl} methods require the reward to be dense.
This means that the reward function should give a feedback on every action the agent performs.
Unfortunately, this is not always the case.
When the reward is given only after multiple actions have been performed, or if a specific situation is met, the agent has to deal with a \emph{sparse reward system}.
Sparse rewards mean that in many of the states of the system there is no clear signal of what is the best course of action.
In these settings, standard \gls{rl} algorithms struggle to learn good policies, leading to poor performances or to no solution at all.
Nonetheless, given the ubiquity of sparse rewards systems, being able to deal with them is fundamental.
Even more so when the agent is in the real world.
There are many factors rendering the design of a dense reward function difficult for real world problems.
The task could be not well defined, too broad or complex, or require too much information in order to calculate a reward after each action.
An example of this is a search-and-rescue mission \cite{murphy2008search}.
While the task is well defined - explore the area and look for people in need of help - the design of a dense reward function is not easy.
Awarding the agent for every discovered person would be simple, but such a function is incredibly sparse. 
To have a denser reward, the position of the dispersed people would have to be known in advance, but this, other than requiring too much information, would remove the search part from the search-and-rescue mission.
This is an extreme scenario, but it is a good example of how agents capable of dealing with sparse rewards could help in addressing many difficult and dangerous situations.

Recently, many scientists focused their research efforts on proposing algorithms capable of solving sparse reward problems.
This ranges from reward shaping, in which the reward function is modified in a way that can lead the agent to solve the task, \cite{mataric1994reward}, to intrinsic motivation, where the agent generates the reward by itself by maximizing another metric \cite{pathak2017curiosity, burda2018large}.
Other methods include the agent learning from past experiences to reach self assigned goals \cite{Andrychowicz2017HER}, or solving of auxiliary tasks that can help in learning how to reach the main goal \cite{jaderberg2016reinforcement}.

In general, when the reward is sparse the agent has no clue where to start to solve the task.
In these situations a good strategy is to focus on \emph{exploration}, to discover all the possible things that can be done in the environment.
This strategy also applies to the previous example of a search-and-rescue mission: by focusing on exploring the environment, the agent can discover the missing people and thus be able to maximize its reward.
In this thesis, we approach the problem of sparse rewards with this idea in mind, studying and proposing a set of algorithms capable of performing efficient exploration in this kind of settings.
Contrary to the methods discussed until now, this is done through the other family of policy learning methods mentioned before: \glspl{ea} \cite{vikhar2016evolutionary}.
The reason behind this is the greater versatility of \glspl{ea} algorithms thanks to them being gradient-free.
Not having to calculate any gradient allows for their applications in a much wider range of situations, without the requirement to have a differentiable policy parametrization.
At the same time, this comes at a price: \glspl{ea} are slower than gradient-based methods in their optimization process.

\section*{Evolutionary Algorithms}
As the name indicates, \gls{ea} are directly inspired by the natural evolution process described by Darwin \cite{darwin1859origin}.
In his work, Darwin described how, given a population of living beings in an environment, the elements better adapted to the environment have higher chances of reproduction, while the ones less suited are more likely to die prematurely.
In time, the natural selection of fitter elements will lead to the development of a population of agents highly adapted to live in the environment in which it finds itself \cite{darwin1859origin, de2016evolutionary}.

A similar selection process is also applied by \glspl{ea} when performing the search for policies.
These algorithms work with a population of policies that are used to generate new policies through two operators: \emph{mutation} and \emph{crossover}.
The first, inspired by the generic mutation happening in nature, randomly mutates some of the parameters of a policy to generate a new one.
The crossover is instead inspired by sexual reproduction: the parameters of two or more policies are mixed to generate a new set of policies.
The newly generated policies according to these operators are then evaluated in the environment.
Among them, only the best ones are selected to form the next generation population, according to a given performance metric, usually called \emph{fitness function}.
Notwithstanding the different name, \glspl{ea}' fitness function has the same role the reward function has in \gls{rl}.
For this reason, and given that in the literature the problem of sparse rewards is mainly defined with respect to \gls{rl}, throughout this thesis we will refer to the fitness of a policy as to its reward.
Contrary to \gls{rl} algorithms, expecting a reward for every action performed, the performance evaluation done by \glspl{ea} on their policies happens only at the end of their execution, rendering them better suited for sparse rewards systems.
Moreover, thanks to the fact that these algorithms do not make any assumption about the fitness landscape they are in, they have proven useful in many domains, from circuit design \cite{cohoon2003evolutionary} to the evolution of other artificial intelligence algorithms \cite{gent2020artificial}.
Nonetheless, researchers have been wondering why standard \glspl{ea} cannot generate the same amount of diversity generated by the natural evolution process.
These algorithms are very prone to converge to a single local minima, with the whole population being the same, a phenomenon known as \emph{population collapse} \cite{de2003multi, badran2007roles}.
One of the possible culprits for this issue has been identified in the fitness function \cite{Lehman2008NS}.
If not properly designed, this function can lead the search astray or towards dead ends.
Moreover, relying on the fitness function to drive the search can be problematic in situations in which this function is extremely sparse.
In these scenarios it can happen that no policy in a whole population can get any reward at the end of its evaluation, rendering the search for a solution difficult.
To address these problems, Lehman and Stanley proposed a novel take on the design of \glspl{ea} by introducing the \gls{ns} algorithm \cite{Lehman2008NS, lehman2011abandoning}.
This algorithm works by completely ignoring the reward and just focusing on exploration, looking for \emph{novel behaviors}.
The idea behind the \gls{ns} approach is that while the amount of behavior of a certain complexity is limited, many policies in the search space can express the same behavior.
By only focusing on novel behaviors the search can thus discover more complex and diverse ways of acting, possibly finding a solution to the task.
This allows the algorithm to return a whole collection of policies, each one with a significantly different behavior from the others.
According to how the policies and their behaviors are defined, this collection can then be used in many different ways \cite{gomes2018approach, chatzilygeroudis2018reset, duarte2017evolution, cully2016evolving}.
The authors have shown that this way of performing the search for policies, rather than optimizing a reward function, allows the discovery of solutions for problems in which many other reward-based algorithms get stuck \cite{lehman2011abandoning}.

\subsubsection{Quality-Diversity algorithms}
The introduction of \gls{ns} extended the range of problems to which \glspl{ea} can be applied, sparking a renewed interest in using methods from this field for learning policies.
This lead to the development of many similar algorithms that, rather than looking for the best solution to a problem, can perform \emph{divergent search} and find a set of many different solutions \cite{lehman2011evolving, pugh2016qdfontier, cully2017quality, mouret2015illuminating, Eysenbach2018DIAYN, Gravina2016Surprise}.
Some of these algorithms are designed to not only optimize the diversity of the discovered set of policies, but also their quality towards the given goal.
Due to this characteristic these methods are usually referred to as \gls{qd} algorithms \cite{pugh2016qdfontier, cully2017quality}.
By discovering a whole set of policies rather than a single one, these algorithms make the embodied agent more adaptable to different situations and tasks.
This has been shown by using \gls{me} \cite{mouret2015illuminating}, a well known \gls{qd} methods, to train an hexapod to walk and to quickly adapt to damages to its legs, even if no damage was present at training time \cite{CUlly2015MAPElites}.

The generation of multiple policies, and the great exploration ability divergent search algorithms have made us choose them as an approach to perform policy search for sparse rewards.
An overview of the literature on these methods and a detailed description of how both \gls{rl} and \glspl{ea} work will be given in Chapter \ref{chap:related}.

Notwithstanding their advantages, divergent search algorithms are still limited in many ways.
The most notable limitation is the way the diversity of the policies' behaviors is calculated.
This is done in a space, called the \gls{bs}, in which the behaviors are represented and that is usually hand-designed by the engineer setting up the learning system.
While this can help tailor the solution to the problem at hand, it requires a significant amount of prior knowledge about the features of the system, the robot and the task itself.
The search will also be constrained by the biases of the designer's choices or by the need to have access at runtime to informations difficult to extract.
For example, in the case of a robot learning to throw a ball in different ways \cite{kim2017learning}, the ball position needs to be properly tracked in order to estimate where it touches the ground.
Not doing so would make it difficult to distinguish between different behaviors.
This can strongly limit the range of application of \gls{qd} algorithms.

\section*{Learning the behavior space}
In light of the problems just discussed, Chapters \ref{chap:taxons} and \ref{chap:signatures} explicitly address the following question:\\

\begin{center}
\noindent\emph{How to remove the requirement of hand designing the \gls{bs} for \gls{ns}, \\ and \gls{qd} algorithms in general?}\\
\end{center}

Having an algorithm capable of autonomously learning the \gls{bs} in which the search is performed could greatly help in this direction.
A way to address the issue is proposed in Chapter \ref{chap:taxons}, with the introduction of the \glsfirst{taxons} algorithm \cite{paolo2019unsupervised}.
This is a divergent search algorithm based on \gls{ns} and designed to build in parallel, and in an unsupervised way, both a collection of diverse policies and the space in which the behavior of said policies is represented.
It does so by encoding the high-dimensional observation of the final outcome generated by a policy evaluation into a lower dimensional representation through an \gls{ae} \cite{Hinton1993Autoencoder}.
The conducted experiments show that \gls{taxons} can efficiently explore the whole space of possible behaviors in the tested environments.

The evaluation of a policy behavior in \gls{taxons} is done with respect to its final outcome.
To reach this outcome, the system traverses a whole trajectory of states during the evaluation of a policy, that are ignored by the algorithm.
However, in some situations, considering only the final outcome is not enough to properly explore.
Let us consider a robot learning how to throw an object against a wall.
A good way to distinguish between the behaviors of different policies is by measuring the different positions where the object hits the wall.
At the same time, without knowing the exact moment in which this happens, it is difficult to determine which state - from the trajectory of traversed states - to use to perform this measurement.
In Chapter \ref{chap:signatures}, a possible way of dealing with this problem is presented: the signature transform.
This is a mathematical object consisting of an infinite series of integrals encoding a stream of data into a vector representing the geometrical properties of this stream \cite{chen1954iterated, chen1957integration, chen1958integration}.
Thanks to its many positive properties, the signature transform has been used in the field of machine learning to encode sequences of data in a fast and easy way \cite{bonnier2019deep, fermanian2021embedding}. 
Given that the signature can encode a sequence of data into a single vector, it is a good candidate to encode the whole trajectory of states traversed by the system during the evaluation of a policy.
This method is compared against simpler approaches that can help in removing the limitation of working only with the final outcome.
The results show that these simpler methods are as effective as the signature in this regard.
For this reason, we choose to use one of those simpler approaches in Chapter \ref{chap:stax}.

\section*{Exploiting sparse rewards}
After addressing the issue of reducing the amount of prior information about the search space and removing the limitation of hand-designing the \gls{bs}, the manuscript continues by focusing on another important question:\\

\begin{center}
\noindent\emph{How to take advantage of the rewards discovered \\
during the search performed by \gls{ns}?}\\
\end{center}

\gls{ns} is limited by the fact that all information about any reward discovered during the exploration process is discarded.
Even if sparse, the reward is extremely useful in steering the search towards more interesting areas of the search space.
To address this problem, the \gls{serene} algorithm is introduced in Chapter \ref{chap:serene}.
This method combines the exploration abilities of \gls{ns} with the exploitation of the reward provided by \emph{emitters}, instances of reward based \glspl{ea} performing local exploration \cite{fontaine2020covariance, cully2020multi}.
By combining these methods, \gls{serene} can separate the exploration from the reward exploitation in an alternating two-steps process.
The algorithm can then seamlessly adapt to a wide range of situations, from settings in which the reward is present in multiple areas of the search space to ones in which it is not present at all.
Similarly to other divergent search algorithms, \gls{serene} returns a collection of policies.
This collection is divided in two groups: one containing the high performing policies optimized by the emitters and one containing the set of diverse policies usually returned by \gls{ns}.
The conducted experiments show how this approach can quickly explore the search space and optimize policies in settings of sparse rewards; even when multiple reward areas are present in the search space.

The \gls{taxons} and \gls{serene} methods proposed in Chapters \ref{chap:taxons} and \ref{chap:serene} address complementary problems of \gls{ns}: the hand-design of the \gls{bs} and the exploitation of possible rewards found during the search.
The final contribution of this manuscript is presented in Chapter \ref{chap:stax}: the \gls{name} algorithm.
This is a method merging ideas from previous chapters to address both problems at the same time.
It does so by taking advantage of the representation learning abilities of \gls{taxons} to drive the search in the two-steps process of \gls{serene}.
At the same time, the limitation present in \gls{taxons} due to only observing the outcome of a policy to describe its behavior is removed thanks to the lessons learned in Chapter \ref{chap:signatures}.
Experiments show that \gls{name} can properly learn a good representation of the behavior space, discovering and quickly exploiting all the rewards present in the environment.
This allows to have an algorithm capable of dealing with sparse reward environments, with minimal prior information about the task and the environment in which the embodied agent operates.

\gls{name} represents the culmination of the work conducted in this thesis, that started with the identification of the sparse rewards problem and the possible ways of dealing with it.
The advantages and the shortcomings of the approaches introduced throughout this manuscript are discussed in Chapter \ref{chap:discussion}, highlighting the new and exciting possible research directions arising.
The manuscript concludes with Chapter \ref{chap:conclusion}, providing a final overview of the work developed during this thesis.

As a recap, the work in this manuscript focuses on \emph{how to deal with sparse rewards settings.}
A good strategy to use in these situations is to focus on \emph{exploration}, which \gls{qd} algorithms, and \gls{ns} in particular, are explicitly designed to do.
At the same time, these methods perform the search in an \emph{hand-defined} space, which can be a strong limitation in certain situations.
The first presented contributions address this issue by \emph{autonomously building a behavior space}, in order to reduce as much as possible the amount of prior information given at design time.
The manuscript continues by highlighting how \gls{ns} discards all informations about the most interesting parts of the search space by ignoring all rewards.
The second presented contribution focuses on this limitation by introducing a method that augments \gls{ns} with the ability to \emph{exploit any reward discovered in the search space}.
This is done without hindering the exploration performed by \gls{ns}.
Finally, these two contributions are merged by introducing a method that can \emph{autonomously build a behavior space} while also \emph{focusing on any discovered reward without hindering exploration}.
\label{chap:intro}
\clearpage\null\thispagestyle{empty}

\chapter{Background and related work}
{\hypersetup{linkcolor=black}\minitoc}
\label{chap:related}
This chapter introduces the different research topics and algorithms on which this thesis builds.
Along with the overview of the theory, there will be a discussion of the related approaches and the current state of the art.
The chapter starts with an overview of \glsfirst{rl} and the framework in which it operates when performing policy search for embodied agents.
This will help in properly framing the issue of sparse rewards and describing some of the methods used to approach the problem.
From there the chapter will describe how an \glsfirst{ea} works and how, thanks to \glsfirst{ns}, it is possible to overcomes some of the limitations of classical \glspl{ea} when optimizing policies.
As said in Chapter \ref{chap:intro}, \gls{ns} sparked the development of a new family of \glspl{ea} focusing on generating a set of diverse solutions, rather than a single optimal one.
An overview of these methods will be given, with a detailed description of the most widely used algorithms in the domain.

\section{Reinforcement Learning}
\label{sec:related_rl}
\begin{figure}[!h]
    \centering
    \includegraphics[width=.7\textwidth]{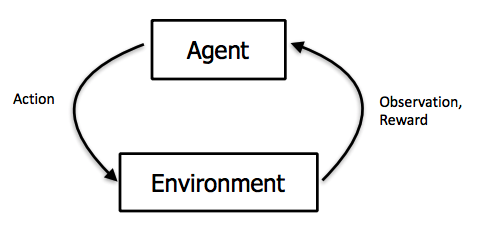}
    \caption[Reinforcement Learning cycle]{Reinforcement Learning cycle}
    \label{fig:rl}
\end{figure}

Chapter \ref{chap:intro} discussed how an embodied agent can act in the environment it is in by following a policy.
The policy is what governs the agent and defines how it acts in different situations.
The goal of many learning algorithms for embodied agents is to learn a policy allowing the agent to solve a given task \cite{kober2014policy, kober2013reinforcement, sigaud2019policy}.
\glsfirst{rl} \cite{sutton2018reinforcement} is a branch of machine learning that can be used for this.
The focus of the \gls{rl} framework is in fact to learn what actions an agent has to perform in an environment in order to maximize a reward function.
The learning is done through a trial-and-error approach in which, at each time step, the agent performs an action in the environment and observes the outcome of said action, expressed through an observation and a reward.
The agent will then use this observation and reward to select the next action.
The process is represented in Fig. \ref{fig:rl}.

\gls{rl} has proven to be a powerful and generic framework to model and address problems in many different fields, from resource management \cite{mao2016resource} and cross-light control \cite{arel2010reinforcement}, to robotics \cite{levine2016end} and even chemistry \cite{zhou2017optimizing}.
In order to properly address a problem through an \gls{rl} algorithm, the problem needs to be formulated as a \gls{mdp}.

\subsection{Markov Decision Process}
\label{sec:mdp}
A \glsfirst{mdp} is a time-discrete stochastic control model that can be used to describe a decision process with possibly random outcomes \cite{Bellman1957mdp}.
Thanks to its generality, it can be used to model problems in many fields, from robotics to economy.
For this reason it has been widely studied and multiple approaches capable of solving an \gls{mdp} have been proposed \cite{howard1960dynamic, sutton2018reinforcement, katehakis1987multi}.

The main components of an \gls{mdp} that need to be specified in order to formulate a problem in this framework are the following:
\begin{itemize}
    \item the set of states $S$. It represents all the possible states $s \in S$ in which the system can be in. The set can be either discrete or continuous;
    \item the set of actions $A$. It contains all the actions $a \in A$ that the agent can perform on the environment. As with $S$, $A$ can also be either discrete or continuous;
    \item the state transition probability function $P_a(s', s)$. This function determines the probability of the system transitioning from state $s$ at time $t$ to state $s'$ at time $t+1$ due to the agent performing action $a$: $P_a(s', s) = Pr(s_{t+1} = s'|a_t = a, s_t = s)$;
    \item the reward function $R_a(s', s)$. This function determines the immediate reward the agent receives when transitioning in state $s'$ from state $s$ due to performing action $a$.
\end{itemize}
An important aspect of a problem formulated through an \gls{mdp} is the fact that the state transition function has to respect the \emph{Markov property}.
It states that given a state $s_t$ and an action $a_t$ the probability that the system moves to state $s_{t+1}$ is independent from all previous states and actions.
Respecting the Markov property allows to greatly simplify the problem and its solution, thus many algorithms dealing with \glspl{mdp} assume that the system respects it.

Finally, the \emph{policy function} can be defined as $\pi(s_t) = a_t$.
This function determines the action $a_t$ to perform at time $t$ when the system is in state $s_t$.
The goal of a learning system acting on an \gls{mdp} is to find the best policy $\pi(\cdot)$ such that it maximizes the expected discounted sum over a potentially infinite horizon:
\begin{equation}
    \mathbb{E}\left[ \sum_{t_0}^{\infty} \gamma^t R_{a_t}(s_t, s_{t+1}) \right],
\end{equation}
where $\gamma^t \in [0,1]$ is a discount factor for future rewards.
The closer $\gamma$ is to 1, the more importance is assigned to the future, the closer it is to 0, the more shortsighted the agent will be with respect to the reward.

\begin{wrapfigure}{r}{.5\textwidth}
    \centering
    \includegraphics[width=.45\textwidth]{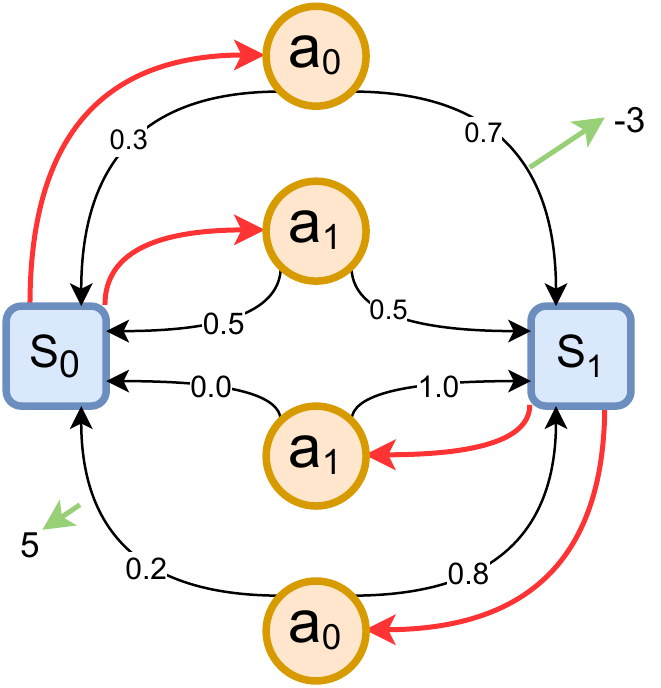}
    \caption[MDP example]{Simple \gls{mdp}. The two states in which the system can be are represented in blue, while the two possible actions are in orange. In each state the agent can select either one of the two actions by following the red arrow. This will cause the system to transition to another state according to the probabilities indicated over each black arrow. The two possible rewards are indicated by the green arrows.}
    \label{fig:mdp}
\end{wrapfigure}
Fig. \ref{fig:mdp} shows a simple \gls{mdp}.
In this example, the system can find itself in two possible states, shown in blue: $[s_0, s_1]$.
At the same time, the agent can perform two actions, depicted in orange: $[a_0, a_1]$.
The selection of an action while being in a state is represented by the red arrows.
Each action makes the system transition to either one of the two states according to the probability indicated over each black arrow.
In this system, the agent can obtain two rewards, indicated by the green arrows.
Let us consider the case in which the system is in state $s_1$.
The agent can perform either action $a_0$ or action $a_1$.
By performing action $a_0$, the system can go either back to state $s_1$, with probability 0.8, or go to state $s_0$, with probability 0.2.
In this last case, the agent will receive a reward of 5.

Finding optimal policies for \glspl{mdp} is not easy, even for the simple example shown in Fig. \ref{fig:mdp}.
For this reason, a lot of research has addressed these problems, leading to the introduction of multiple methods capable of solving them \cite{sutton2018reinforcement, sigaud2019policy, slivkins2019introduction, bellman1953introduction}.
These methods work by calculating the value of a state $s$ through the value function $V(s)$ and updating the policy $\pi(s)$ with respect to $V(s)$ \cite{sutton2018reinforcement}.
The value function is the expected return when starting from state $s$ and following policy $\pi$ and describes how good it is to be in a given state:
\begin{equation}
    V(s) = \mathbb{E}_{\pi, P_a} \left[ R(s'|s) + \gamma V(s') \right].
    \label{eq:value_func}
\end{equation}

The policy is updated with respect to the value of a state by selecting the action leading to the highest valued next state:
\begin{equation}
    \pi(s) = \text{argmax}_a \left\{ \sum_{s'} P(s'|s, a) \left( R(s'|s,a) + \gamma V(s') \right) \right\}.
    \label{eq:policy_opt}
\end{equation}
As it can be seen from Fig.\ref{fig:mdp}, how good an action is strongly depends from the state the system is in.
The reward obtained by performing action $a_0$ in state $s_0$ is different than the one obtained if the system was in $s_1$.
This important aspect is evaluated through the Q-value function $Q(a, s)$.
This function represents the \emph{state-action value}, that is the value of performing an action $a$ in a state $s$ and is defined as:
\begin{equation}
    Q_{\pi, P_a}(s, a) = \mathbb{E}\left[R(s'|s) + \gamma \mathbb{E}_{a \sim \pi} Q(s', a|s) \right].
    \label{eq:q_func}
\end{equation}.

Different algorithms use these functions in different ways, but the final goal remains the same: discover the policy leading to the highest reward.


\subsection{Exploration-exploitation trade-off}
Always selecting the action according to Eq. \eqref{eq:policy_opt}, can easily lead the algorithm to get stuck in local minima.
This would prevent it to find an optimal solution to the problem.
The reason behind this is that, by always choosing the action that leads to the highest reward, the agent is only \emph{exploiting} the knowledge it already has, not collecting new informations about the environment.
This strategy is called a \emph{greedy} strategy and it can prevent the discovery of other states and action combinations that can be more rewarding.

To discover new situations it is important to \emph{explore} by testing different actions and visiting different states.
At the same time, it is not given that higher rewarding situations can be found, so focusing too much on exploration rather than exploitation can be a waste of time and resources.
It is then important to find a good balance between exploration and exploitation.
This problem is usually referred as \emph{the exploration-exploitation trade-off} \cite{sutton2018reinforcement}.
There are multiple strategies that have been proposed to deal with it \cite{sykulski2011exploration, sutton2018reinforcement} due to the fact that it is a fundamental problem in many learning settings.
Among them, a very simple but well known one is the \emph{$\epsilon$-greedy} algorithm \cite{sutton2018reinforcement}.

The idea behind this method is simple: each time the agent has to choose an action, it selects the best action with probability $1-\epsilon$ or a random one among the possible actions with probability $\epsilon$.
The balance between the exploration and exploitation can be easily decided by changing the value of $\epsilon$.
To only focus on exploration is enough to set $\epsilon = 1$, while a value of $\epsilon=0$ makes the agent completely greedy.
By carefully setting $\epsilon$, this strategy allows to easily exploit what the agent already knows, but also to explore the environment and gather new information.

Given what has been discussed until now, it is possible to notice how the reward function $R(\cdot)$ plays a fundamental role in the way \gls{rl} algorithms operate.
It is this function that is used to calculate both the value of each state $s$ and to select the action to perform at each given step.
This means that the reward function can be used by the designer of the problem to communicate to the algorithm which goal it needs to achieve.
At the same time, to learn a good policy capable of obtaining the maximum reward possible, most \gls{rl} methods expect a dense reward function.
This implies that $R(\cdot)$ needs to provide a relevant feedback on every single action the agent can perform on the environment.
If the reward function rarely provide its feedback, it is defined as being a \emph{sparse reward} \cite{sutton2018reinforcement, hare2019dealing}.

\subsection{Sparse rewards}
\label{sec:related_sparse}
Given the importance of the role played by the reward function in learning a good policy, situations of sparse rewards can be extremely difficult to deal with \cite{hare2019dealing}.
The \emph{ideal situation} in which to apply a \gls{rl} algorithm is one in which the reward is well defined for each state and proportional to how beneficial that state is to reach the goal.
Unfortunately, for many problems and in many real world settings this is not the case.
Often the reward tends to be extremely sparse.
In these situations, it can happen that the agent never experiences any reward at all, making learning a good policy impossible.

\begin{figure}
    \centering
    \includegraphics[width=.8\textwidth]{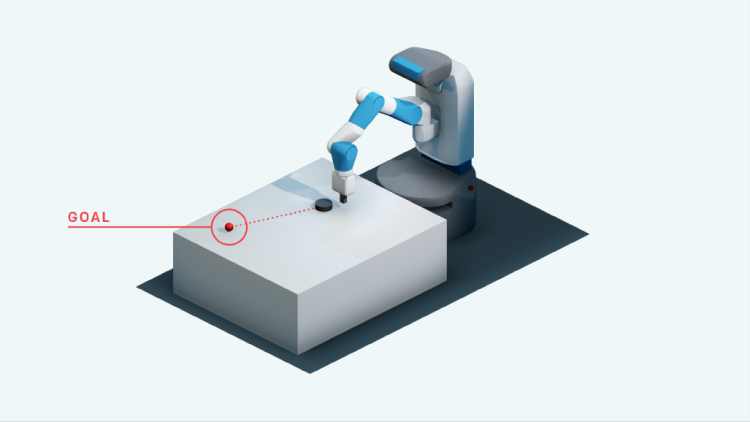}
    \caption[Sparse Reward environment]{Sparse reward environment from OpenAI Gym \cite{openaigym}.}
    \label{fig:sparse_rewards}
\end{figure}

An example of a sparse reward system is given in Fig. \ref{fig:sparse_rewards}, where a robotic arm has to push the black puck towards the goal, in red.
The reward is given only if the puck reaches the target.
This makes it very sparse: only one state among all the possible ones the system can be in would generate a reward.
Another possible approach is to give the reward proportionally to how close the puck is to the goal.
At the same time, this would require a precise measurement of its position at each time-step, requiring a more complex setup.
Moreover, there can be other situations in which simple workarounds like this are not possible; the search-and-rescue example given in Chapter \ref{chap:intro} is one of those.
To have embodied agents capable of efficiently dealing with real world problems, approaches that can overcome sparse rewards problems needs to be developed.

For these reasons, many researchers have focused on sparse rewards situations, leading to the development of many different approaches.

\subsubsection*{Reward shaping}
The most simple way of dealing with a sparse rewards situation is by performing \emph{reward shaping} \cite{mataric1994reward}.
This consists in augmenting the original reward function with additional features that can help the agent solve the task.
Giving the reward depending on the distance between the puck and the goal, as discussed for the robot arm example, is a form or reward shaping.
This approach has proven useful in many situations \cite{hu2020learning, badnava2019new, berner2019dota, trott2019keeping,jonschkowski2014state}.
Notable is the work conducted by OpenAI on the Dota2 \cite{dota2} game, in which the authors managed to train an agent to reach superhuman performances \cite{berner2019dota}.
Nonetheless, reward shaping comes with multiple shortcomings.
The new features added to the reward require huge amount of prior knowledge about the system and the task.
Moreover, if not properly designed, this reward can introduce bias into the problem, leading the agent astray and preventing it to efficiently solve the problem.

\subsubsection*{Self-assigning goals}
Another approach is to have the agent self assign goals in order to train itself.
This can be done by taking advantage of previously encountered situations, as done with \gls{her} \cite{Andrychowicz2017HER}.
The main idea of the method is to store the state transitions in a \emph{replay buffer}, even if no reward has been achieved.
These stored transitions then are used by the algorithm to learn how to reach a goal, even if the given goal is not the one needed to solve the task.
A similar goal relabeling approach is also used in the works from Levy et al. \cite{levy2017learning} and Nair et al. \cite{Nair2018ImaginedGoal}.
The first method takes advantage of hierarchical \gls{rl}, an approach that learns a hierarchy of policies in which at each step high-level policies select low-level policies to perform the task \cite{levy2017learning}.
As for \gls{her}, the self-assigned goals are sampled from the collection of already visited states.
On the contrary, the second approach takes advantage of an unsupervised representation learning algorithm to generate the targets to reach \cite{Nair2018ImaginedGoal}.
In this method the visited states are not collected into a buffer, but are used to learn a compressed representation through a \gls{vae} \cite{kingma2013auto} from which the goals are then sampled.
This approach increases the exploration abilities of the algorithm, allowing it to reach states not yet visited by the agent.

A similar strategy to deal with sparse rewards is the one used by Florensa and his colleagues \cite{florensa2018automatic}, sampling the goals from a \gls{gan} \cite{goodfellow2014generative} rather than from the latent space of a \gls{vae}.
The adversarial approach allows the agent to assign itself goals that become more complex with time.
This strategy creates a sort of curriculum that continuously pushes the boundaries of the agent's abilities.
The strategy of using a curriculum to progressively increase the difficulty of the tasks to solve is also used by Riedmiller and his colleagues \cite{riedmiller2018learning}.
In their work, the agent tries to solve auxiliary tasks that start very simple and become more complex with time.
This is repeated until the agent can solve the main task.
The capacity to solve auxiliary tasks grants the agent a lot of flexibility in the range of goals that can be achieved.
However, in order for the curriculum of tasks to be meaningful towards reaching the desired goal, the simpler tasks need to be selected by the engineer before training the agent.

\subsubsection*{Intrinsic Motivation}
A completely different idea for approaching sparse rewards problems is \gls{im} \cite{oudeyer2009intrinsic, aubret2019survey}.
Directly inspired from developmental psychology, this approach consists in the agent generating its own learning signal, without expecting any reward from the environment.
This is similar to how kids learn something driven by their own curiosity rather than expecting some external reward \cite{gopnik1999scientist}.
An action can be defined intrinsically motivated if it is performed with the goal of collecting more information about the outcome of the action itself, rather than with the expectation of obtaining a reward  \cite{oudeyer2009intrinsic, aubret2019survey}.
Given this definition, there are multiple ways to provide \gls{im} to an agent \cite{singh2009rewards, oudeyer2009intrinsic}.
\paragraph{Novelty}
The agent can be rewarded if it reaches more novel states, where the novelty is obtained by calculating an estimation of how often a state has been visited.
With discrete state spaces, this estimation can be calculated by simply counting the number of times the agent visited a given state \cite{brafman2002r, kearns2002near}.
For state spaces that are too big or continuous, this approach is not possible, requiring different strategies.
One of those strategies is the use of a density estimation model over the state space to generate what the authors call \emph{pseudo-count} \cite{bellemare2016unifying}.
For instance, the \gls{rnd} method uses a couple of \glspl{nn} to estimate the novelty of each visited state \cite{burda2018exploration}.
One of the two \glspl{nn} remains untrained with random weights, while the other is trained to reproduce the output of the untrained one.
The novelty of a state is then assumed to be proportional to the difference between the output of the two models.
The idea behind this is that the more often a state has been visited, the more the trained network has been trained to reproduce the same output of the random \gls{nn} with respect to that state.
This leads to a lower difference between the outputs of the two models.
On the contrary, for a state that has been less visited, the trained model will return an output that is less similar to the one of the non trained \gls{nn}.
This leads to the identification of a more novel state.

\paragraph{Empowerment}
Another possible way of providing \gls{im} to an embodied agent is through \emph{empowerment} \cite{klyubin2005empowerment, klyubin2005all, salge2014empowerment}.
Based on Information Theory \cite{stone2015information}, this approach pushes the agent to maximize its control over the environment.
This can be done by maximizing the entropy of future states the system can move into while also minimizing the entropy of the future states the system can reach given the action \cite{klyubin2005all}.
The rationale behind this being that the more an agent can influence the system to move to different states, the more control it has on the environment.
There have been other approaches to empowerment, but the method is still hindered by the complexity of calculating the empowerment metric \cite{aubret2019survey}.
Simplifying this calculation would require a model of the environment itself, as done by Mohamed and Rezende \cite{mohamed2015variational}, but in many situations this is not feasible.

\paragraph{Curiosity}
In a different direction goes the notion of \emph{curiosity} \cite{oudeyer2004intelligent, silvia2012curiosity, oudeyer2018computational}.
This consists in providing the agent with a transition model on what can happen in the environment.
The error between the predictions of the agent and what actually happens in the environment is then used as a reward signal by the agent.
This will push the agent to explore more and look for situations in which it does not know what will happen, hopefully discovering a solution for the task.
Curiosity as an \gls{im} approach has been systematically studied in \cite{burda2018large}, showing an alignment between the curiosity objective and the hand-designed rewards present in many game environments.
This leads to good performances for curiosity-based methods in the tested situations.
The agent's error can also be used when predicting the consequences of its own actions to drive the learning process \cite{pathak2017curiosity}.
The error is calculated in a feature space learned through a self-supervised inverse dynamics module.
This approach offers the advantage of not having to deal with the many unimportant informations present in pixel space that can keep the error high, e.g. leaves moving in the background or small changes in luminosity.
The idea of curiosity can also be mixed with the one of the agent self-assigning goals as done with \gls{imgep} \cite{baranes2013active, Forestier2017IMGEP, laversanne2018curiosity}.
\glspl{imgep} work by having the agent sample its own desired goals from a given goal space according to a given strategy.
Among the possible sampling strategies, a powerful one is to have a \gls{mab} \cite{slivkins2019introduction} with the goal to maximize the competence of the agent in reaching the sampled goals.
This approach allows the creation of a curriculum, starting from simpler targets and moving toward more complex ones.
Contrary to other goal self-assigning methods, \glspl{imgep} are usually based on a population approach, in which multiple policies are tested at the same time.
This allows more flexibility and an easier recovery of discovered abilities compared to other approaches that do not use a population \cite{laversanne2018curiosity}.

Other than through curiosity, \gls{gep} help in addressing sparse rewards settings thanks to the separation between the exploration of the search space and the exploitation of any possible discovered reward.
Forestier et al. \cite{Forestier2017IMGEP} use such a method by firstly learning a goal-parametrized policy capable of reaching any goal from any state and then using this policy to solve the task.
This method has also been extended in IMGEP-UGL \cite{Pere2018IMGEP_UGL}, in which the searched goal space is learned in an unsupervised fashion from the environment observations thanks to a dimensionality reduction algorithm.
Another method, proposed by Colas and his colleagues \cite{colas2018gep}, performs a task agnostic exploration phase before learning an inverse policy on the task to solve.
Similarly, GoExplore \cite{Ecoffet2019GO_Explore} deals with the problem by using a two-phases strategy.
It starts by exploring as much as possible, without caring about the reward.
This leads to the building of a set of interesting states and trajectories leading to these states. 
Then in the second phase, it chooses one of the trajectories to turn it into a robust policy.

As stated in Sec. \ref{sec:mdp}, \gls{rl} methods work best in situations of dense rewards, rendering the problem of sparse rewards complex to address.
Another way to deal with the issue is to rely on a different family of policy search algorithms, requiring the reward signal to be provided only at the end of the policy evaluation: \glsreset{ea}\glspl{ea}.

\section{Evolutionary Algorithms}
\label{sec:related_ea}
A different approach in learning policies for embodied agents consists in using a \glsfirst{ea} \cite{Mitchell1998EA}.
\glspl{ea} are a family of optimization algorithms inspired by the theory of natural evolution described by Darwin in its work ''On the origin of species'' \cite{darwin1859origin}.
These algorithms take advantage of the concept of \emph{survival of the fittest} to perform \emph{direct policy search}.
They work with a population of individuals - each one of the corresponding to a parametrized policy - that is randomly initialized at the beginning of the search.
This population is then evaluated in the environment and the performance of the individuals in it is measured through a given \emph{fitness function}.
Notwithstanding the different name, this function has a similar role than one of the reward function in \gls{rl} methods.
The main difference between the two functions is that the fitness function returns the reward for the whole evaluation episode, rather than evaluating the time-steps.
Similarly to what happens in natural settings, the individuals whose fitness is too low will be discarded.
At the same time, the agents deemed fit to survive are selected and allowed to reproduce through some variation operators, generating a new population of individuals to evaluate.
Each iteration of this process is called a \emph{generation}.

The selection of the individuals that are used for the generation of the new population can be performed in multiple ways \cite{Mitchell1998EA, eiben2015evolutionary}.
The most common selection strategies are:
\begin{itemize}
    \item \textbf{Roulette wheel selection}: the probability of selecting an individual is proportional to its fitness; the better the fitness, the higher chance for that individual to be chosen;
    \item \textbf{Tournament selection}: multiple tournaments are performed between individuals sampled from the population. The winner of each tournament is selected for reproduction. This method can be improved by performing, at the end of each tournament, a probabilistic selection of the individual to reproduce: select the first with probability $p$, the second with probability $p*(1-p)$, the third with probability $p*(1-p)^2$, etc. \cite{miller1995genetic};
    \item \textbf{Elitist selection}: the new population is composed not only by the new individual generated through reproduction, but also by the best elements from the previous generation. This allows to preserve particularly good set of parameters.
\end{itemize}

\begin{figure}[!h]
    \centering
    \includegraphics[width=.6\textwidth]{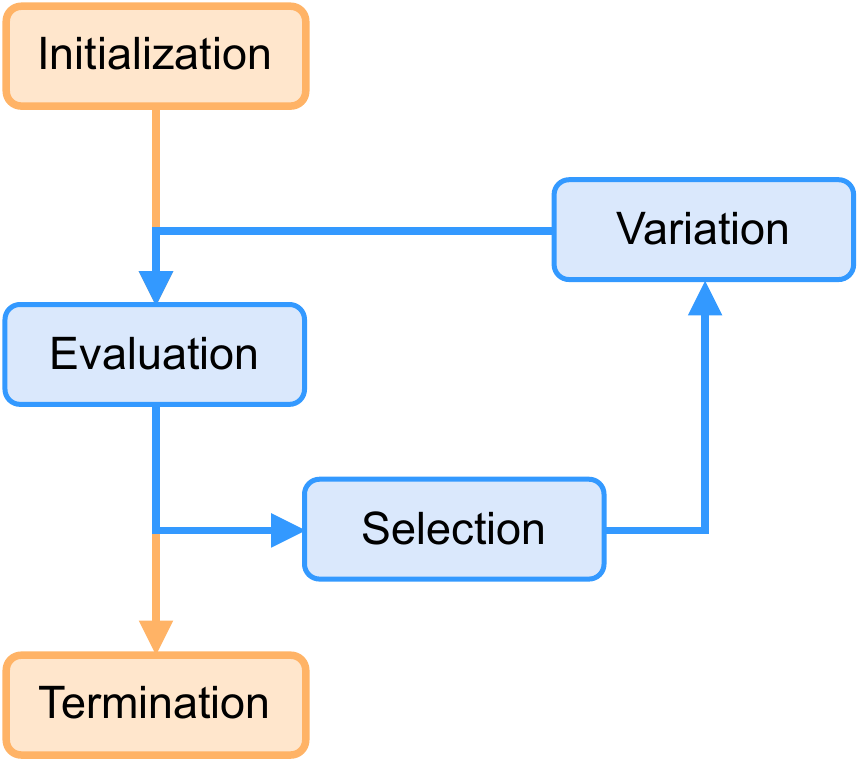}
    \caption[Evolutionary algorithms cycle]{Evolutionary algorithms cycle \cite{eiben2015evolutionary}}
    \label{fig:evo}
\end{figure}

By continuously selecting the fittest agents the algorithm can generate agents that reach higher and higher fitness scores, thus discovering better ways to solve the task.
The evolution cycle, represented in Fig. \ref{fig:evo}, is repeated until a given termination condition is reached.

\subsubsection*{Genotype and Phenotype}
Before defining how the creation of new individuals from the selected ones in the current population is performed, it is useful to define two important concepts for \glspl{ea}: the \emph{genotype} and the \emph{phenotype}.

The \emph{genotype} corresponds to any set of parameters used to define an individual.
In the case of natural living beings the genotype is the DNA, encoding information about any aspect of the being.
On the contrary, the \emph{phenotype} is the expression of the information contained in the genotype.
Keeping the parallel with the natural world, the phenotype expressed by the DNA can be the shape of the body, the color of the eyes or of the hair, or even the behavior of a living being.
The mapping between the genotype and the phenotype of an agent strongly depends on the kind of setting and environment the algorithm is being applied to \cite{eiben2015evolutionary}.

\subsubsection{Variation operators}
In general, \glspl{ea} act on the genotype in order to obtain and observe different phenotypes. 
This is done through two operators, used during the variation step of Fig. \ref{fig:evo}, to generate a new set of individuals from the existing one:
\begin{itemize}
    \item \textbf{Mutation}: this operator is inspired by the genetic mutation happening in nature. 
    It works by randomly changing some of the values of the set of parameters composing the genotype;
    \item \textbf{Crossover}: inspired by sexual reproduction, it combines the genotypes of two or more individuals to generate a new one.
\end{itemize}
These two operators are blind to the reward, meaning that the operations the perform do not depend on the reward.
Nonetheless, they help the algorithm to continuously generate new individuals, properly exploring the genotype space in the search for solutions.
At the same time, given that the crossover works with multiple individuals at once, it is not always straightforward to apply, depending on the structure of the genome and the way it is expressed in the phenotype.
For this reason, many works in recent years tend to only use the mutation operator when generating new individuals \cite{cully2015creative}.

The mutation and crossover operators are fundamentally stochastic in the selection of both individual and parameters on which to operate.
While this renders the search less efficient compared to methods like \gls{rl}, it also removes the need for the calculation of a gradient.
Moreover, contrary to \gls{rl} methods requiring the problem to be structured as an \gls{mdp}, \glspl{ea} can work with optimization problems structured as black-box functions \cite{eiben2015evolutionary}.
The algorithm does not require any information about the internal structure of the problem or of the function it is optimizing, it just needs to provide an individual as possible solution and observe its final performance.
This allows \glspl{ea} to be applied to a wide variety of problems, from the design of buildings \cite{schwehr2011evolutionary} to the generation of music \cite{dostal2013evolutionary} and images \cite{secretan2008picbreeder} and the creation of virtual creatures \cite{lehman2011evolving, medvet2021biodiversity}.
\glspl{ea} have also be used for the evolution of the topologies of \glspl{nn} through neuroevolution \cite{stanley2002evolving, stanley2019designing}, or for the generation of policies used to control robots and embodied agents in general \cite{zykov2004evolving, doncieux2010behavioral, CUlly2015MAPElites}.
The only requirement in this regard is that the policy $\pi(\cdot)$ has to be parametrized by a set of parameters $\theta \in \Theta$.
These parameters correspond to the genotype of the controller policy, while the way the embodied agent acts in its environment is its phenotype.

\subsection{Multi-objective optimization}
\label{sec:related_moo}
\glspl{ea} can be applied also to \glsfirst{moo} problems, in which the method optimize multiple objective functions at the same time \cite{hwang2012multiple}.
An example of this can be seen in automated factories, in which the engineers have to optimize both the speed and the accuracy of the robots working on the assembly line.
The solution of these kind of problems requires to find a good \emph{trade-off} between the different objectives to optimize.
This is not an easy task, and the discovery of the right trade-off can require multiple evaluations of the problem.
The advantage \glspl{ea} have in these situations is that being population based they allow the discovery of multiple possible trade-offs at once.

\begin{figure}[!h]
    \centering
    \includegraphics[width=\textwidth]{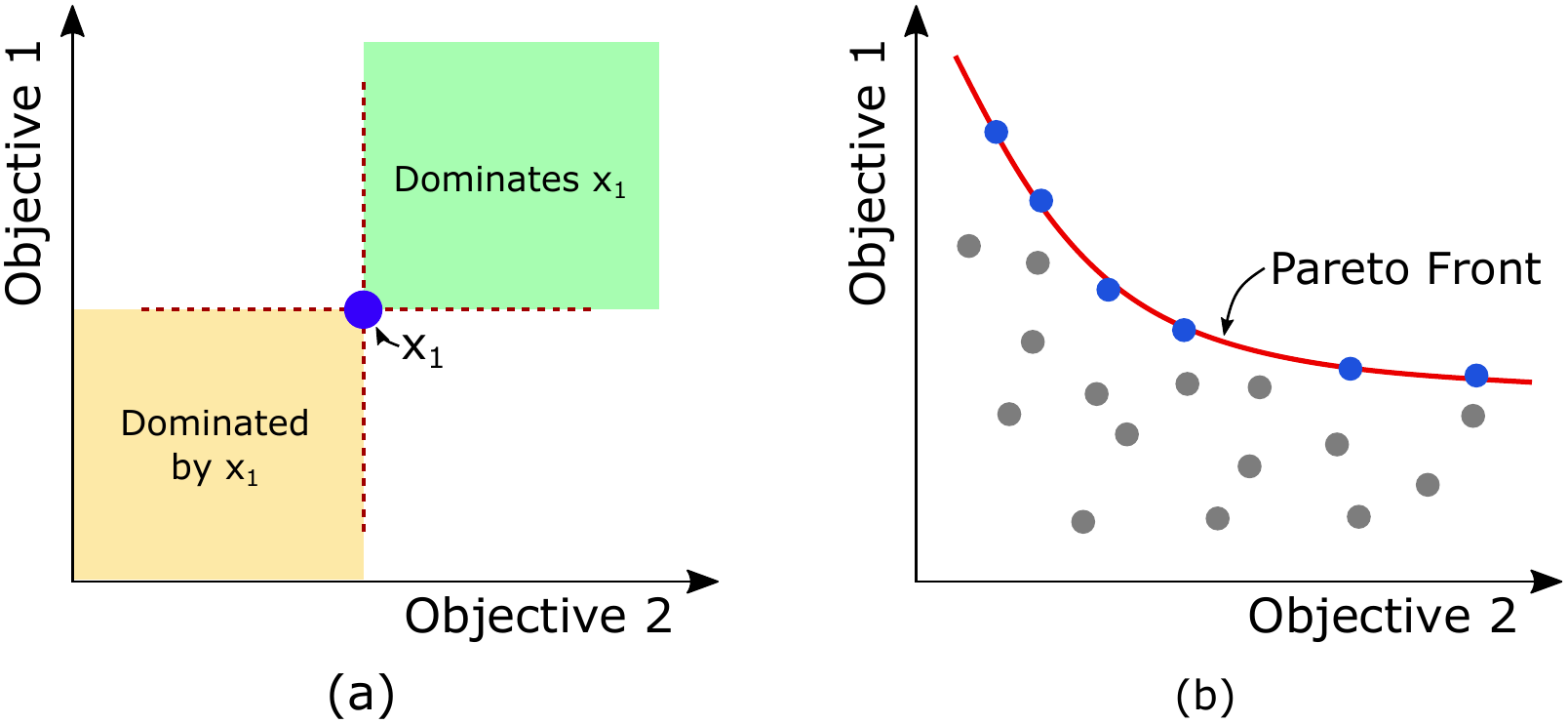}
    \caption[Pareto front]{An example of a \gls{moo} with two objectives to maximize. (a) A possible solution $x_1$ divides the plane defined by the two objectives into multiple areas of domination and non-domination. (b) A possible Pareto front, in red, for a \gls{moo} problem.}
    \label{fig:related_pareto}
\end{figure}

An important concept when dealing with \gls{moo} problems is the one of \emph{Pareto dominance} \cite{hwang2012multiple}.
To explain it, consider two candidate solutions for a \gls{moo} problem, $x_1$ and $x_2$.
The solution $x_1$ is said to \emph{dominate} $x_2$ if and only if the two following conditions apply:
\begin{enumerate}
    \item $x_1$ is not worse than $x_2$ on any of the problem's objectives;
    \item $x_1$ is better than $x_2$ on at least one of the objective functions.
\end{enumerate}
An example of how a solution $x_1$ divides the plane between dominated and non-dominated regions is shown in Fig. \ref{fig:related_pareto}.(a).
Among all the possible solutions to a \gls{moo} problem, the set of \emph{non dominated} solutions contains all the best trade-offs that can be found for the problem.
This set is referred as \emph{Pareto front}.
A representation of a possible Pareto front for a \gls{moo} with two objectives to maximise is shown in red in Fig. \ref{fig:related_pareto}.(b).
In blue are represented the solutions belonging to the front, while all the other dominated solutions are shown in grey.

\begin{figure}
    \centering
    \includegraphics[width=\textwidth]{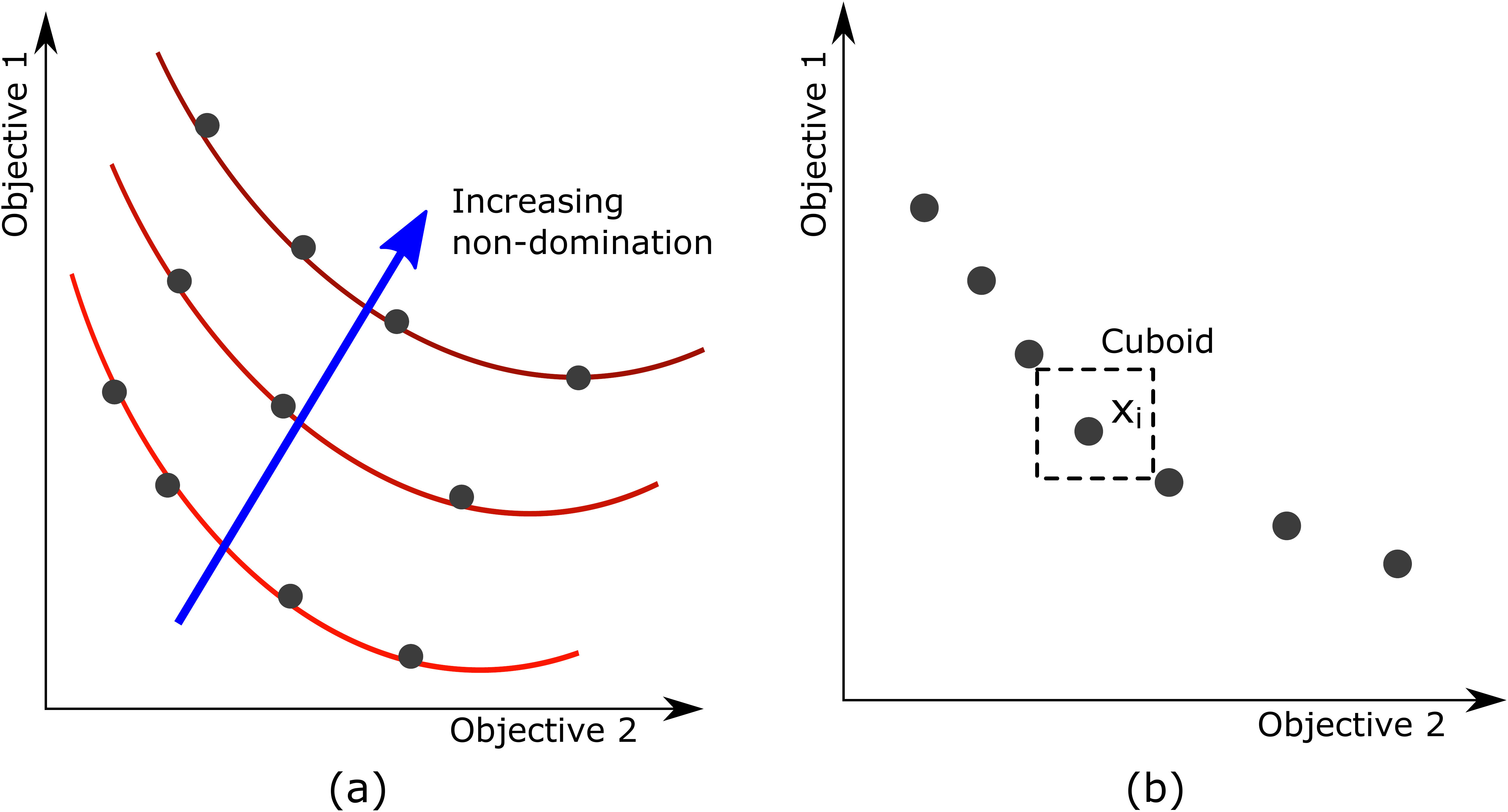}
    \caption[NSGA-II policy ordering]{(a) Representation of non-dominated fronts sorting. Each grey circle represents a solution with respect to the two optimization objectives. The blue arrow represents the direction of increasing non-domination. (b) Representation of crowding distance cuboid calculation around solution $x_i$.}
    \label{fig:nsga_fronts}
\end{figure}

A well known \gls{ea} designed to deal with \gls{moo} problems is NSGA-II \cite{deb2002fast}.
This method works by sorting all the possible solutions into non-dominated fronts on an ascending level of non-domination, as shown in Fig. \ref{fig:nsga_fronts}.(a).
At each generation $g$, the new population $\Gamma_{g+1}$ is then filled according to front ranking, by first adding to it the elements from the most non-dominated front, then the ones from the second-most non-dominated front, etc., until the population is complete.
If a front is only partially selected, that is, if the solutions on the front are more than the remaining positions in $\Gamma_{g+1}$, only the ones with the highest \emph{crowding distance} are selected \cite{deb2002fast, raquel2005effective}.
This distance is a metric providing an estimate of the density of individuals around any single solution $x_i$.
To have meaning, it needs to be calculated only between solutions on the same non-dominated front.
The calculation is performed by measuring the area of the largest cuboid surrounding an individual $x_i$, without including any other solution on the same front, as shown in Fig. \ref{fig:nsga_fronts}.(b) \cite{raquel2005effective}.

\begin{figure}[!h]
    \centering
    \includegraphics[width=\textwidth]{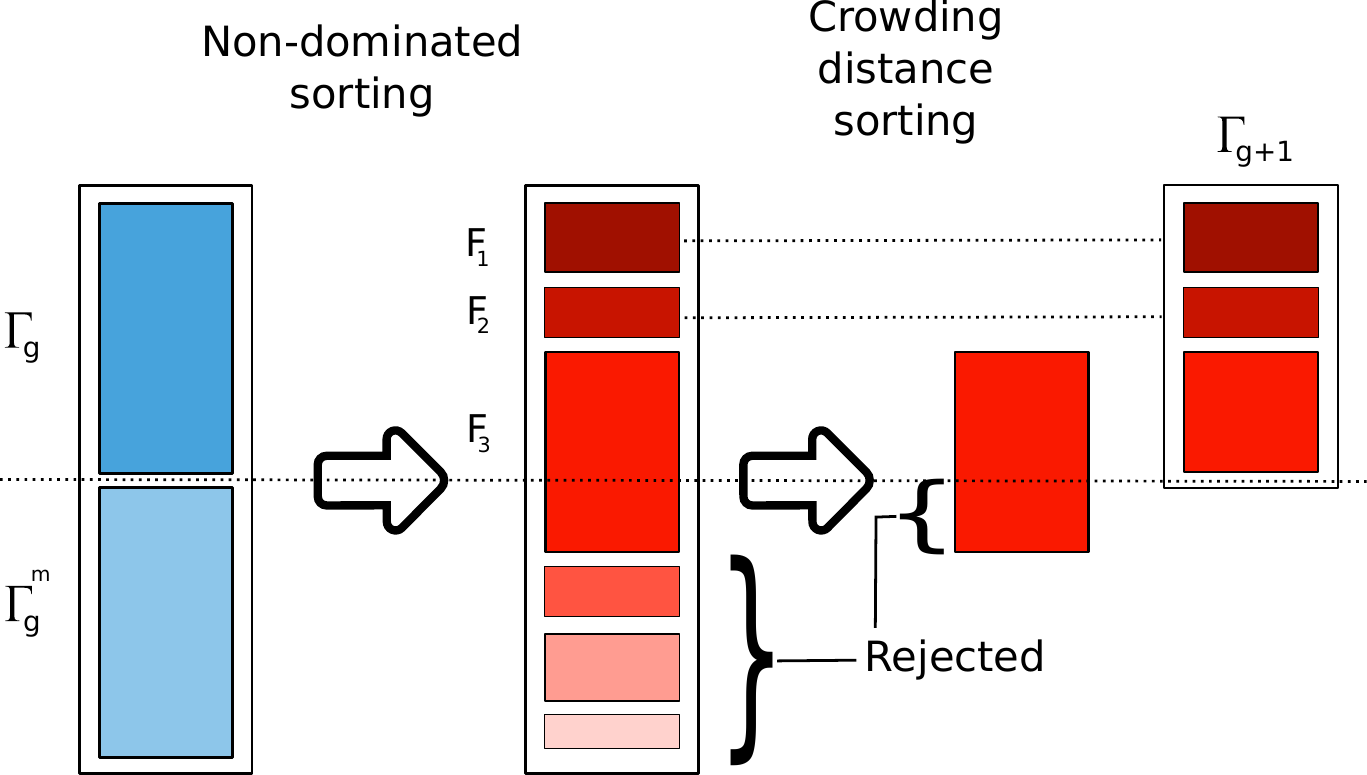}
    \caption[NSGA-II policy selection]{Schematic of the NSGA-II procedure as shown by \cite{deb2002fast}. $\Gamma_g$ is the population at generation $g$, while $\Gamma_g^m$ are the corresponding offsprings. $F_i$ are the non-dominated fronts in which the solutions are divided before selection. Finally $\Gamma_{g+1}$ is the new population at generation $g+1$.}
    \label{fig:stax_nsga}
\end{figure}

An overview of the whole NSGA-II selection procedure is illustrated in Fig. \ref{fig:stax_nsga}.

\subsection{Searching for Diversity}
\label{sec:related_qd}
\glspl{ea} require the definition of a fitness function measuring the quality of the solutions with respect to a given goal because they are \emph{objective-based}.
The designer has to know in advance the goal to reach and the way the progress towards this goal can be measured.
Designing such a fitness function is not an easy task.
Moreover, if the function is not properly defined, it can lead the \gls{ea} to fall victim of deception and converge to local optima \cite{lehman2011abandoning}.
This can lead to a phenomenon known as \emph{population collapse}, in which the whole population is composed by similar individuals, preventing any possible advancement out of the local optimum \cite{de2003multi, badran2007roles}.
There are multiple ways to overcome this kind of problem.
An approach is to use additional objectives through a \gls{moo} approach \cite{jensen2004helper, doncieux2014beyond}.
By using the additional objectives, the population can escape local optima with respect to the main optimization objective.
Another promising strategy is to force the algorithm to preserve the diversity of the population.
In order to do so, many approaches tried to preserve the diversity of the population from the genotype point of view \cite{mahfoud1995niching}.
This can be done through multiple mechanisms as speciation and clustering \cite{stanley2002evolving, yin1993fast} or tournament selections \cite{harik1995finding}.
However, preserving diversity in the genotype space does not imply that the individuals will behave in different ways. 
Multiple genotypes can express the same kind of phenotype.
For this reason, a more interesting approach is to push the algorithm to preserve diversity in the phenotype space.
Maximizing the diversity in the phenotype space also allows for the discovery of many different ways in which a problem can be solved.
This means that rather than returning a single solution, the algorithm should return a whole set of diverse solutions.
A very well known algorithm using this strategy is \glsfirst{ns} \cite{Lehman2008NS}. 
Introduced by Lehman and Stanley, the method performs the search not by targeting a fitness objective, but by searching for a set of individual whose phenotypes are as different as possible.
Algorithms performing the search through this strategy can be defined as \emph{divergent search algorithms} \cite{lehman2012benefits}.
Moreover, this strategy allows the algorithm to not get stuck in any local optima that could possibly be present in the fitness landscape.

Note that, since \gls{ns} was introduced in the context of evolutionary robotics and artificial life, the phenotype usually consisted in the way the agent behaved. 
For this reason \gls{ns}, and similar methods, are said to maximize the \emph{behavioral diversity} of the individuals \cite{mouret2012encouraging} and the phenotype itself is referred to as \emph{behavior descriptor}.
Considering this, and the fact that the focus of this manuscript is the learning of policies in situations of sparse rewards, from now on we will consider the individuals generated by the presented algorithms as policies $\pi(\cdot)$ parametrized by a set of parameters $\theta \in \Theta$.
As stated in Sec. \ref{sec:related_ea}, this parameters correspond to the genotype of the individuals, while their phenotype corresponds to the behavior these policies have in the environment.
Thanks to the flexibility of \glspl{ea}, this point of view comes with little loss of generality but it will help in keeping the discussion aligned to the way these methods are presented in the literature.

\subsubsection{Novelty Search}
\label{sec:related_ns}
\glsfirst{ns} is an \gls{ea} that works by replacing the fitness metric used by standard \glspl{ea} with a \emph{novelty} metric.
This metric pushes the search towards novel areas of the search space.
The novelty is calculated in a space $\mathcal{B}$ called \glsfirst{bs} in which the behavior of each policy $\pi(\cdot)$ parametrized by $\theta_i \in \Theta$ is represented.
This space is usually hand-designed by taking into account the system and the task to be fulfilled.
The policies generated by the algorithm are run on the system for a given number of steps $T$, traversing a trajectory of states $\tau_S = [s_0, \cdots, s_T]$, where each $s_t$ corresponds to the state of the system at time step $t$.
These traversed states are observed through some sensors, such that they produce a corresponding trajectory of observations $\tau_{\mathcal{O}} = [o_0, \cdots, o_T]$, with $o_t \in \mathcal{O}$, where $o_t\in\mathcal{O}$ is a potentially under-complete observation of the state of the system at time $t$.
An observer function $O_B: \mathcal{O}^T \rightarrow \mathcal{B}$ then maps this trajectory of observations to a hand-designed \emph{behavior descriptor} $b_i$ for the policy $\theta_i$.
The overall process can be summarized by introducing a behavior function $\phi$ mapping each policy $\theta_i$ to its corresponding behavior descriptor:
\begin{equation}
   \phi(\theta_i) = b_i
   \label{eq:ns_phi}
\end{equation}
The descriptor usually consists in a vector of real numbers.
For example, in the case of a robotic arm, it can be the final position and orientation of its end effector.

Once computed, the descriptors are used to calculate the policies' novelty.
This novelty represents how different the behavior of each policy is with respect to the behavior of the other agents.
In practice it is calculated as the average distance between the behavior descriptor of a given policy $\theta_i$ and the descriptors of the $k$ closest policies in the behavior space $\mathcal{B}$.
It is calculated as: 
\begin{equation}
\label{eq:ns_novelty}
\eta(\theta_i) = \frac{1}{|J|}\sum_{j \in J}\text{dist}(b_i, b_j)  = \frac{1}{|J|}\sum_{j \in J}\text{dist}\big(\phi(\theta_i), \phi(\theta_j)\big),
\end{equation}
where $J$ is the set of indexes of the $k$ policies closest to $\theta_i$ in $\mathcal{B}$. 

The novelty of the policies is calculated at each generation $g$ and used to choose the policies for the next generation.
Moreover, $N_Q$ policies are sampled to be stored into an \emph{archive} $\mathcal{A}_{\text{Nov}}$, or \emph{repertoire}, returned as outcome of the algorithm.
These policies can be chosen either randomly or by selecting the most novel ones from the current generation of offsprings \cite{Gomes2015NSHyperpar}.
The archive is also used to keep track of the already explored areas of the space $\mathcal{B}$.
This is done by choosing the $|J|$ closest neighbors used in equation \eqref{eq:ns_novelty} not only from the current population and offspring but also from the archive.
The whole \gls{ns} approach is shown in Alg. \ref{alg:rel_ns}, in which the evaluation budget $Bud$ is equal to the total number of policy evaluations.

\begin{algorithm}[!h]
\caption{Novelty Search}\label{alg:rel_ns}
\textbf{INPUT:} evaluation budget $Bud$, parameter space $\Theta$, behavior space $\mathcal{B}$, variation function $\mathbb{V}(\cdot)$, population size $M$, number of offspring per parent $m$, number of policies to add to archive $N_Q$\;
\textbf{RESULT:} archive $\mathcal{A}_{\text{Nov}}$\;
Initialize $\mathcal{A}_{\text{Nov}} = \emptyset$\;
Initialize generation counter $g=0$\;
$\Gamma_0 \leftarrow M$ policies from $\Theta$\;
Evaluate $\theta_i, ~~ \forall \theta_i \in \Gamma_0$\;
Calculate behavior descriptor $b_i= \phi(\theta_i) \in \mathcal{B} ~~ \forall\theta_i \in \Gamma_0$\;

\While{$Bud$ not depleted}{
    Generate offsprings $\Gamma^m_g \leftarrow \mathbb{V}(\Gamma_g)$\;
    Evaluate $\theta_i, ~~ \forall \theta_i \in \Gamma^m_g$\;
    Calculate behavior descriptor $b_i= \phi(\theta_i)  \in \mathcal{B} ~~ \forall\theta_i \in \Gamma^m_g$\;
    Calculate novelty $\eta(\theta_i) = \frac{1}{|J|}\sum_{j \in J}\text{dist}(b_i, b_j), ~~ \forall\theta_i \in \Gamma_g \bigcup \Gamma^m_g$\;
    $\mathcal{A}_{\text{Nov}} \leftarrow N_Q \text{ samples from } \Gamma^m_g$\; 
    Generate $\Gamma_{g+1}$ with most novel $\theta_i \in \Gamma_{g} \bigcup \Gamma^m_{g}$\;
    $g = g+1$\;
}
\end{algorithm}

By choosing the most novel policies from the current generation to compose the next population, the search is always pushed towards less explored areas of $\mathcal{B}$. 
Doncieux et al. have shown that the repertoire will tend to uniformly cover the \gls{bs} \cite{doncieux2019ns_theory}.
At the same time, measuring and comparing behaviors is still an open question, requiring the definition of a good behavior representation. 
Nonetheless, as long as the \gls{bs} is properly designed, \gls{ns} is a very good exploration algorithm in situations in which the reward is either sparse or contains local minima. 

Since the introduction of \gls{ns}, many divergent search algorithms have been developed, using different mechanisms to drive the search: curiosity \cite{stanton2016curiosity}, empowerment \cite{campos2020explore}, surprise \cite{Gravina2016Surprise}, diversity \cite{CUlly2015MAPElites, Eysenbach2018DIAYN, cully2017quality, pugh2016qdfontier}, and novelty \cite{lehman2011evolving}.

\subsubsection{Quality-Diversity algorithms}
Notwithstanding the capacity for exploration of \gls{ns}, completely ignoring the fitness function prevents the algorithm to exploit any of the rewards potentially found during the search.
This leads the set of solutions returned by the method to have arbitrary performances.
Trying to overcome this problem has led to the development of the \gls{qd} family of algorithms \cite{cully2017quality, pugh2016qdfontier}.
The methods of this family generate a set of diverse solutions that at the same time have high performances with respect to the given objective.

Among the first \gls{qd} algorithms developed is \gls{nslc}, an algorithm that combines \gls{ns} with the mechanism of \emph{niching} to foster the evolution of high performances solutions \cite{lehman2011evolving}.
The niching mechanism consists in considering policies that are close enough in the \gls{bs} as belonging to the same niche.
The competition with respect to performances then happens only between members of the same niche.
The rationale for this approach is that agents with high performance in an area of the search space cannot dominate the ones in a different area that have lower performances, thus preventing the collapse of diversity during the search process.
This is done by calculating a \emph{local competition objective} for each agent $\theta_i$ by counting how many agents among the $k$ nearest neighbors used to calculate the novelty in Eq. \eqref{eq:ns_novelty} have lower fitness than $\theta_i$.
The objective encodes the performance of an agent with respect only to its nearest neighbors.
The algorithm then uses a \gls{moo} strategy to optimize both the novelty of the agents and their local competition objective.
This strategy allowed the authors to generate a wide set of robot morphologies capable of efficient locomotion.

\gls{ns} based methods are characterized by two main factors: they can work with continuous \gls{bs}, and the population of agents performing the search is kept separated from the archive of stored solutions. 
A different approach is the one introduced by Mouret and Clune with \glsfirst{me} \cite{mouret2015illuminating}.
This method works by discretizing its \gls{bs} through a multidimensional grid in which the agents are organized.
Moreover, the algorithm does not keep the population separated from the archive: the agents in the grid act both as population and as archive.
The method works by randomly initializing a set of policies $\theta$ that are evaluated in order to calculate their behavior descriptor.
The policies are then placed in the cells of the \gls{bs} grid corresponding to their descriptor.
At this point, the main cycle of \gls{me} starts.
The algorithm samples one of the policies from the grid and generates a new agent by mutating the sampled individual.
The new policy is then evaluated and its behavior descriptor $b_i$ calculated.
If the cell of the grid in which the policy belongs thanks to $b_i$ is empty, the agent is placed in it and another element from the grid is sampled.
On the contrary, if the cell is already occupied by another policy, the algorithm performs a tournament selection between the two policies, only storing the one with the highest fitness.
The cycle is repeated until the whole evaluation budget is depleted.
With time, this allows to improve on the quality of the discovered solutions.
The whole process is shown in Alg. \ref{alg:mapelites}.

\begin{algorithm}[!h]
\caption{MAP-Elites}\label{alg:mapelites}
\textbf{INPUT:} evaluation budget $Bud$, parameter space $\Theta$, discretized behavior space $\mathcal{B}$, variation function $\mathbb{V}(\cdot)$, number of initial policies $M$\;
\textbf{RESULT:} discretized archive $\mathcal{A}_{\text{ME}}$\;
Initialize $\mathcal{A}_{\text{ME}} = \emptyset$\;
$\Gamma_0 \leftarrow M$ policies from $\Theta$\;
Evaluate $\theta_i, ~~ \forall \theta_i \in \Gamma_0$\;
Calculate behavior descriptor $b_i= \phi(\theta_i) \in \mathcal{B} ~~ \forall\theta_i \in \Gamma_0$\;
$\mathcal{A}_{\text{ME}} \leftarrow \theta_i,  ~~ \forall\theta_i \in \Gamma_0$\; 
\While{$Bud$ not depleted}{
    Sample $\theta_i$ from $\mathcal{A}_{\text{ME}}$\;
    Generate offspring $\Tilde{\theta_i} \leftarrow \mathbb{V}(\theta_i)$\;
    Evaluate $\Tilde{\theta_i}$\;
    Calculate behavior descriptor $b_i= \phi(\Tilde{\theta_i})  \in \mathcal{B}$\;
    \eIf{$\mathcal{A}_{\text{ME}}(b_i) == \emptyset$}{
        $\mathcal{A}_{\text{ME}}(b_i) \leftarrow \Tilde{\theta_i}$;
    }{
        $\theta_j \leftarrow \mathcal{A}_{\text{ME}}(b_i)$\;
        \If{$fitness(\theta_j) < fitness(\Tilde{\theta_i})$}{
            $\mathcal{A}_{\text{ME}}(b_i) = \emptyset$\;
            $\mathcal{A}_{\text{ME}}(b_i) \leftarrow \Tilde{\theta_i}$;
        }
    }
}
\end{algorithm}

The discretization of the search space allows \gls{me} to reduce its memory and computational footprints with respect to the continuously growing archive of \gls{ns}.
The cell look-up operation in \gls{me} has a computation cost of just $O(1)$.
At the same time, the computational cost of the nearest neighbors calculation in \gls{ns} is $O(n \log n)$ \cite{friedman1975algorithm}, increasing at each generation with the size of the archive.
Thanks to its performances and its simplicity, \gls{me} has become one of the most well known and widely used \gls{qd} approaches.

Discovering multiple policies and collecting them into an archive, as done by \gls{qd} methods, allows for great generalization, even to settings not seen at training time.
An example of this is the work done by Cully et al. in which a six legged robot learned how to recover from damage received to one of its legs and still be able to walk \cite{CUlly2015MAPElites}.
Initially, the method generates a collection of policies that allowed the undamaged robot to walk.
After one of the robot's legs is disabled, the algorithm explores its collection of policies in order to find the most promising solution that still allows the robot to advance.
This solution is then evaluated and its performance on the damaged robot calculated. 
The algorithm then updates the performances not only of the selected policy, but also of its neighbors, in order to quickly reduce its uncertainty about the quality of all solutions with respect to the new situation.
The rationale of this is that similar behaviors have similar performances even in the new setting.
This reduced uncertainty allows the algorithm to better select which policy to test next.
The cycle repeats until performance on the damaged robot is 90\% or greater of the maximum expected performance for any behavior \cite{CUlly2015MAPElites}.

\gls{me} has been extended in CVT-\gls{me} to address the explosion in the number of cells of the grid while working in high-dimensional \gls{bs} \cite{vassiliades2016scaling}.
In situations of high-dimensional \gls{bs}, the number of cells increases exponentially with the number of dimensions.
With 50 cells per dimension, in the case of a 2-dimensional \gls{bs}, the total number of cells will be $50 \times 50 = 2500$.
This numbers increases to $50 \times 50 \times 50 = 125000$ in a space with just 3 dimensions.
The exponential increase in dimensions reduces the performances of \gls{me} due to the fact that the occupancy of the cells will be very sparse, so very little performances optimization will be performed.
This explosion can be reduced by autonomously changing the size and number of cells along each dimensions \cite{vassiliades2016scaling}.

Other extensions address the issue of working in noisy domains. 
In order to properly compare the behaviors of the policies, \gls{qd} algorithms need to work in deterministic settings.
This greatly limits the range of applicability of these methods.
The problem can be addressed in multiple ways.
Justensen et al. used adaptive sampling and drifting elites \cite{justesen2019map} to strengthen the resistance of \gls{me} to noise.
The method consists in the multiple evaluation of the policies in order to estimate their behaviors and performances.
An agent is evaluated multiple times until it either: it ends up in an empty cell or, if it remains in an already filled cell, it has been evaluated as many times as the best performing policy in that cell, the \emph{elite}.
If the mean performance is higher than the one of the elite, the agent is added to the grid, otherwise it is discarded.
Moreover, the number of evaluations performed on each individuals is increased with time, in order to refine their performance estimate.
To prevent cells of the grid to become empty once the reevaluation of a policy makes it move from one cell to the other, multiple solutions are stored in the same cell.
From all the solutions in a single cell, only the best performing ones will be considered for reproduction when the cell is selected to generate a new policy.
Another approach storing multiple agents in the same \gls{me} cell to deal with the problem of noisy domains is \emph{deep-grids} \cite{flageat2020fast}.
Contrary to what has been done by Justensen et al., deep grids does not use only the elite from a selected cell to generate new offsprings, but rather samples the agents according to their fitness.
This allows to generate a collection of policies that is more stable to noise.

\gls{qd} algorithms have also been combined with other kind of \glspl{ea}, namely \glsfirst{es}, to increase their efficiency and speed of convergence.
\gls{es} is a family of fitness based \glspl{ea} that work by estimating a distribution from which the population to be evaluated is sampled \cite{beyer2002evolution}. 
After each evaluation, the distribution is updated according to the performances of the sampled agents and a new population is sampled.
\glspl{es} are particularly efficient \glspl{ea} and have been recently shown to perform comparably to \gls{rl} methods in certain situations \cite{salimans2017evolution}.

Conti et al. introduced an \gls{es} that used \gls{ns}'s novelty objective to look for novel solutions while improving their performances.
This allows to improve the \gls{es} exploration in setting of sparse rewards while retaining scalability.
The novelty metric allowed the method to overcome the local optima in which standard \gls{es} would easily get stuck.
A different approach is the one of \emph{emitters} \cite{fontaine2020covariance, cully2020multi}.
These works use \gls{me} as a scheduler to launch instances of \gls{es} that perform local search in the \gls{bs} around their point of initialization.
The search is then performed while optimizing the quality of the discovered solutions.
These instances of \gls{es} performing search in a neighborhood of the search space are called emitters.
The concept of emitters is very flexible, allowing any reward-based method to be used as an emitter in combination with a \gls{qd} algorithm.

\gls{qd} algorithms have also been combined with \gls{rl} methods \cite{pourchot2018cem, cideron2020qd}, by using the \gls{ea} to collect the data on which the \gls{rl} algorithm is trained.
This allows for better exploration by overcoming any possible deceptive gradients that could undermine the performances of gradient-based approaches.
At the same time, the data efficiency of the \gls{rl} approach allows to quickly optimize the policy with respect to the goal.
The higher-performing gradient-optimized policy is then injected into the evolving population to steer the search towards more profitable areas.

To conclude, the thesis focuses on the study and development of algorithms capable of dealing with sparse reward settings.
In these situations, a good strategy for the agent is to focus on exploration until a reward is discovered, at which point the agent has to be able to quickly optimize the rewarding solution.
While \gls{rl} methods can efficiently optimize policies thanks to gradient descent, they struggle with hard to explore problems.
On the contrary, \gls{qd} methods have proven to be very powerful in solving problems in which exploration is difficult.
Nonetheless, they present one major limitation: diversity is measured in low-dimensional \gls{bs} that are usually hand-designed.
This requires more involvement from the system’s designer, leading to the possible introduction of bias and to increased costs of deployment.
The next chapter will present an approach designed to deal with this limitation.
\clearpage\null\thispagestyle{empty}

\chapter{TAXONS}
{\hypersetup{linkcolor=black}\minitoc}
\label{chap:taxons}
\vspace{0.4pt}
\par\noindent\rule{\textwidth}{0.4pt}
This chapter is adapted from the following publication:\\

\noindent\textit{Paolo, G., Laflaquiere, A., Coninx, A., \& Doncieux, S.} \textbf{Unsupervised learning and exploration of reachable outcome space.} In 2020 IEEE International Conference on Robotics and Automation (ICRA 2020).
\par\noindent\rule{\textwidth}{0.4pt}

\section{Introduction}
\label{sec:tax_intro}
Chapter \ref{chap:related} discussed how, in order to deal with sparse rewards, a good strategy is to completely focus on exploration.
Divergent search \glspl{ea} like \gls{qd} methods are good candidates for this.
This is due to multiple reasons.
First, these methods generate simple policies, each one specialized in reaching a sub-part of the search space, rather than looking for a single complex policy able to cover the whole space as is the case for \gls{rl} algorithms.
This can seem limiting from the point of view of generalization: being so simple, these policies cannot generalize to new problems.
However, the whole collection of policies returned by \gls{qd} methods can be used on different problems, as long as the collected policies are reevaluated to asses their performances.
This strategy has been useful, for instance, in making a robot resilient to damage \cite{CUlly2015MAPElites} or to generate complex behaviors by combining these simple policies in the context of hierarchical \gls{rl} \cite{Eysenbach2018DIAYN}.
Moreover, these methods do not need any reward to drive the exploration and find new policies, the search is therefore not misled by deceiving reward gradients.
At the same time, an outcome space can be shared by different tasks and different domains, meaning that the same repertoire of policies can thus be applied to multiple tasks a posteriori \cite{gomes2018approach, chatzilygeroudis2018reset, duarte2017evolution, cully2016evolving}. 

This chapter will present one of these divergent search algorithms: the \glsreset{taxons}\gls{taxons} method \cite{paolo2020unsupervised}.
It takes advantage of the exploration strength of \gls{ns} while overcoming one of its main limitations: the need to hand-design the outcome space, also called behavior space, in which the novelty of each policy is evaluated.
\gls{ns} has recently been shown to tend towards a uniform exploration of the outcome space, which is an unbiased strategy in the absence of reward \cite{doncieux2019ns_theory}. 
Removing the need to hand-design the \gls{bs} helps in reducing the amount of prior information needed at design time, rendering the algorithm more widely applicable.
The outcome space, in fact, needs to be adapted for any new agent and/or environment. 
Apart from being costly in terms of human resources, defining by hand the appropriate features of the outcome space requires for the experiment designer to know the features of the robot, environment, and tasks. 
The search will also be constrained by the biases of the designer's choices. 
Being able to autonomously learn the outcome space in which policies are discriminated can remove said limitations while improving the applicability of these approaches.
\gls{taxons} does so by learning the outcome space while performing the search through the use of an \glsfirst{ae} \cite{Hinton1993Autoencoder}. 
\glspl{ae} are a family of \glspl{nn} architectures commonly used for dimensionality reduction. 
\newline

In the following, other approaches from the literature that learn the \gls{bs} will be presented in section \ref{sec:tax_related}, then \gls{taxons} will be introduced in section \ref{sec:tax_method}.
Section \ref{sec:tax_experiments} will discuss the experiments performed to test the method and its related results.
The chapter will conclude with section \ref{sec:tax_conclusion} with a recap of what has been done and a discussion on the next steps and ideas explored in the rest of the thesis.
\section{Related work}
\label{sec:tax_related}
Given the important role played by the \gls{bs} in divergent search algorithms, its design is fundamental.
Hand-designing it can be problematic, requiring prior knowledge on the system and on the problem to be solved.
Moreover, the designer can introduce some unexpected bias in situations in which it is not clear what features would benefit the search.
For these reasons, methods combining representation learning algorithms with \glspl{ea} have recently been proposed.
Representation learning algorithms are a family of methods that can learn abstract features characterizing data.
They can be used to autonomously represent high-dimensional data into a lower dimensional space \cite{lesort2018state}.
Combining them with \glspl{ea} reduces the amount of task related prior knowledge required at design time, moving the engineering efforts from the design of the task specific representation to the design of the task agnostic representation learning algorithm.
This makes the approaches more flexible with respect to the kind of problem they are applied on: the same representation learning method can learn good representations for different situations without many algorithmic changes.

Meyerson et al. \cite{meyerson2016learning} proposed to learn a domain-specific behavior descriptor by starting from an underlying, generic descriptor requiring minimal domain-specific knowledge. 
The domain-specific descriptor then is in the form of a weighted vector over all the possible state-action pairs, thus reducing the need to take into account every situation at design time.
At the same time, having to weight every state-action pair limits the approach to discrete and low-dimensional domains.
In DeLeNoX\cite{Liapis2013Delenox}, the authors developed a \gls{ns}-based algorithm to evolve 2-dimensional space-ships shapes in which the novelty of each shape was calculated in the low-dimensional feature space of an \gls{ae}.
The fact that the \gls{ae} is trained from scratch on each new generation of shapes is the limiting factor of the algorithm, removing any memory of previous generations. 

Autonomous learning has not been limited only to algorithms relying on continuous \gls{bs} like \gls{ns}, but has also been applied to approaches with discrete ones like MAP-Elites\cite{CUlly2015MAPElites}.
Innovation Engines \cite{nguyen2015Innovation} generate and classify novel images thanks to the learned feature space of an \gls{ae}.
However, the \gls{ae} is trained beforehand on a dataset of images and not during the search process.
More recently, the approach introduced by Cully and Demiris \cite{cully2018hierarchical} uses a hierarchical strategy to combine hand-designed features with learned ones.
Having a hierarchy of \glspl{bs} has the advantage of requiring the update of only a part of the \glspl{bs} stack when changing the task, rendering the algorithm more flexible. 
Another approach that uses an \gls{ae} to learn the \gls{bs} online while searching for policies is AURORA \cite{cully2019aurora}.
In this work the \gls{ae} learns the space from the trajectories of the raw observations of the internal states of the system.
The learned space is then used to differentiate the policies by calculating the distances among the generated trajectories projected in the learned space.
Notwithstanding the power of the algorithm to discover a host of different trajectories, using observations of the internal states can be limiting in situations in which the states do not carry enough information.
For example if a robot is being trained to move a box from one position to another, the box needs to be explicitly tracked in order to distinguish between the different outcomes of the policies, adding another layer of complexity to the algorithm.
It is to notice that while AURORA is similar in spirit to \gls{taxons}, there are some important differences.
Namely, \gls{taxons} learns the search space directly from high-dimensional RBG images of the environment, extracting the features in an unsupervised way. 
Moreover, the diversity of each policy is assessed not only by using the self-learned outcome space, but also by taking advantage of the information provided by the reconstruction error of the \gls{ae}.
This added measure helps improve both the quality of the search space and the diversity of the generated policies.
At the same time, AURORA can work on whole trajectories while \gls{taxons} can only work on the last observation.
These features will be discussed in more detail in the next section.
\section{Methodology}
\label{sec:tax_method}
As already mentioned, hand-designing the \gls{bs} requires a lot of involvement from the system's designer.
Having to know exactly what he wants to measure, what the task to be solved is, and how to measure the progress towards finding a solution.
All of this requires prior knowledge that has to be collected by accurately studying the system.

This problem is approached with \gls{taxons}\cite{paolo2019unsupervised}, an algorithm based on \gls{ns}.
Instead of relying on a hand-designed \gls{bs}, \gls{taxons} builds it in the form of a low-dimensional representation of the trajectory of observations $[o_0, \dots, o_T]$ collected during the evaluation of each policy.
Here each observation $o_t \in \mathcal{O}$ corresponds to the observations defined in section \ref{sec:related_ns}.
\begin{figure}
  \begin{center}
    \includegraphics[width=\textwidth]{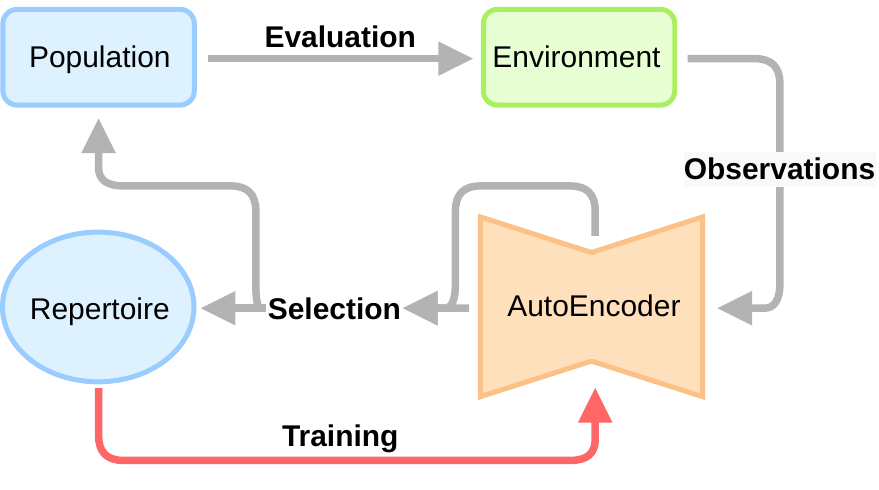}
  \end{center}
  \caption[TAXONS]{High level schematic of \gls{taxons}. It consists of two processes operating in parallel. The first one, the search process (gray arrows), generates a set of new policies, evaluates them in the environment and stores the best ones in the repertoire. The second process, highlighted by the red arrow, is the training of the AE on the observations collected during the evaluation of the policies.}
  \label{fig:taxons}
\end{figure}
To simplify the approach, the last observation $o_T$ of the trajectory is considered to be informative enough to characterize the behaviour of the system during the execution of the related policy.
Consequently, only this last observation will be used to build the \gls{bs} representing the outcome of the policy, for this reason also called \emph{outcome space}.

\begin{figure}
    \centering
    \includegraphics[width=\textwidth]{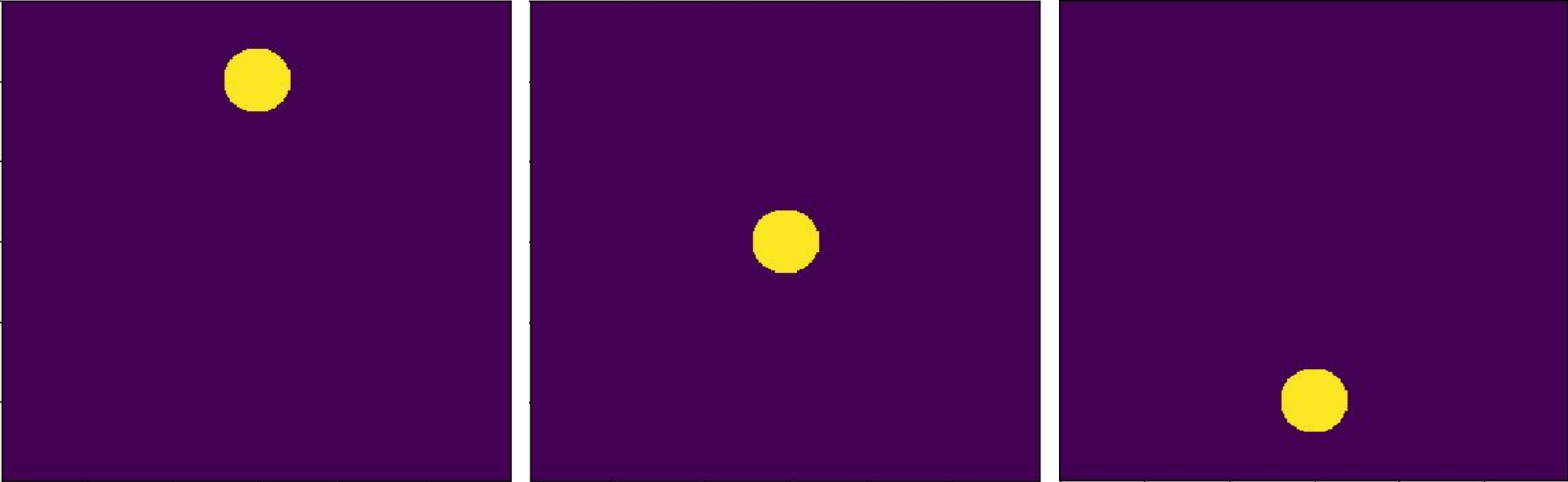}
    \caption[Pixel-wise distance example]{The pixel-wise distance between the second and third images with respect to the first one is the same, notwithstanding the real distance of the circle in the euclidean space being different.}
    \label{fig:ball_distance}
\end{figure}

The observations are considered to be high-dimensional RGB images.
Working directly on such high-dimensional observations can be complicated, if not infeasible, due to two main reasons:
\begin{itemize}
    \item The euclidean distance metric, used to calculate the novelty of the behaviors, does not work properly in high-dimensional spaces \cite{Aggarwal2002distance};
    \item In the case of RGB images, calculating the pixel-wise distance would return a distance between the images themselves rather than a distance between the objects in the images. 
    This makes it difficult to distinguish between different situations. 
    An example of this can be seen in figure \ref{fig:ball_distance}. 
    The pixel-wise distance between the second and third images with respect to the first one is the same, notwithstanding the real distance between the disks being different.
\end{itemize}
For these reasons a dimensionality reduction algorithm is needed to move to a low-dimensional space where a more meaningful distance between observations can be calculated.
Among the many dimensionality reduction algorithms \cite{sorzano2014survey} that can be used to build the low-dimensional space, this work uses an \gls{ae} \cite{Hinton1993Autoencoder}, given the power and flexibility these methods have.

\glspl{ae} are a class of neural network composed by an encoder $E$ and a decoder $D$.
The encoder projects its input $x \in \mathcal{O}$ in a  space called \emph{feature space} $\mathcal{F}$, usually of lower dimension than the input space.
The decoder then returns the projection from the feature space to the output space.
The goal of the \gls{ae} is to learn to reconstruct its input $x$ on its output:
\begin{equation}
\label{eq:tax_outcome_space}
    \begin{split}
        E: \mathcal{O} \rightarrow \mathcal{F}, \\
        D: \mathcal{F} \rightarrow \mathcal{O}.
    \end{split}
\end{equation}
An important point of the process is the fact that the \gls{ae} has to project $x$ in the lower-dimensional feature space.
This forces it to extract only the important features from the input to properly reconstruct it on its output. 
This whole process happens in an unsupervised way, with the \gls{ae} being trained by minimizing the error function
\begin{equation}
\label{eq:ae_error}
    L(x) = ||x - D(E(x))||^2.
\end{equation}
The whole \gls{ae} is trained in an online fashion on the observations generated at each iteration when evaluating the new policies. 

\gls{taxons} uses the encoder $E$ as observation function, to extract the features from the image, and its lower dimensional feature space $\mathcal{F}$ as outcome space.
Note that in this process no task or reward is required, thus the algorithm needs minimal interventions from the designer, apart from choosing which observation to feed to the \gls{ae}.

A high-level schematic of \gls{taxons} is shown in figure \ref{fig:taxons}, while a more detailed view is given in Alg. \ref{alg:tax_training}.
The method consists of two processes running in parallel:
\begin{itemize}
    \item \textbf{Search process}: highlighted in gray in figure \ref{fig:taxons}, it generates and evaluates the policies, storing the ones with higher diversity in the repertoire. 
    It does so by firstly generating a population of policies, evaluating them in the environment, and then selecting the most novel ones to add to the repertoire. 
    During the evaluation, the last observation $o_T$ from each of the policies is collected and fed to the \gls{ae} that will be used to extract the features for the selection step. 
    This last step is fundamental to the correct operation of \gls{taxons} and is described in detail in section \ref{sec:tax_policy_selection};
    
    \item \textbf{AE training}: highlighted by the red arrow in figure \ref{fig:taxons}, is the training of the \gls{ae} on the observations collected during the evaluation of the policies.
    This training process happens in parallel with the search process and is described in more detail in section \ref{sec:tax_training}.
\end{itemize}

\subsection{Policy selection}
\label{sec:tax_policy_selection}
While in \gls{ns} only the distance in the \gls{bs} is used as a metric, in our approach the best policies are selected according to two metrics, ensuring both their novelty and the representativity of the behaviour space.
The first one, referred to as \emph{novelty}, corresponds to the novelty metric of \gls{ns} already defined in Eq. \eqref{eq:ns_novelty}, reported here for clarity:
\begin{equation}
\eta(\theta_i) = \frac{1}{|J|}\sum_{j \in J}\text{dist}(b_i, b_j)  = \frac{1}{|J|}\sum_{j \in J}\text{dist}\left(\phi\left(\theta_i\right), \phi\left(\theta_j\right)\right),
\end{equation}
More precisely, the mapping $\phi$ in \eqref{eq:ns_phi} is replaced by the mapping:
\begin{equation}
\label{eq:tax_phi}
f(\theta_i) = E\big(o_T^{(\theta_i)}\big), 
\end{equation}
where $o_T^{(\theta_i)}$ is the last observation generated by the policy~$\theta_i$.
This allows us to rewrite Eq. \eqref{eq:ns_novelty} as:
\begin{equation}
    \eta(\theta_i) = \frac{1}{|J|} \sum_{j \in J} \text{dist}\left(f(\theta_i), f(\theta_j)\right) = \frac{1}{|J|} \sum_{j \in J} \text{dist}\left(E\big(o_T^{(\theta_i)}\big), E\big(o_T^{(\theta_j)}\big)\right)
    \label{eq:tax_ae_novelty}
\end{equation}

The second metric, referred to as \emph{surprise}, corresponds to the reconstruction error of the \gls{ae}; it is expressed as:
\begin{equation}
\label{eq:tax_surprise_ae}
    s(\theta_i) = \big|\big|o_T^{(\theta_i)} - D\left(E(o_T^{(\theta_i)})\right)\big|\big|^2.
\end{equation}
This reconstruction error tends to be large when the \gls{ae} processes observations which have not been frequently encountered yet.
This idea is similar to the one introduced with the \gls{rnd} approach \cite{burda2018exploration}.
Maximising this metric during the search pushes new policies to explore novel parts of the state (observation) space. 
This ensures that the observations are representative of the states the system can reach.
In practice, one of the two metrics is picked with a probability of $0.5$ to evaluate every new iteration of policies. 
This strategy is similar to the one used by Doncieux and Mouret \cite{doncieux2010behavioral} to mix different behaviour descriptors.

Combining these two metrics drives the search towards learning an outcome space that is representative of the reachable states of the system and towards policies that are diverse in this space.

\begin{algorithm}
\caption{\gls{taxons}}
 \label{alg:tax_training}
\textbf{INPUT:} evaluation budget $Bud$, parameter space $\Theta$, training interval $I$, variation function $\mathbb{V}(\cdot)$, population size $M$, number of offsprings per parent $m$, number of policies to add to archive $N_Q$\;
\textbf{RESULT:} archive $\mathcal{A}_{\text{Nov}}$\;
Initialize \gls{ae} $D(E(\cdot))$ with random parameters\; 
Initialize population $\Gamma_0 \leftarrow M$ policies from $\Theta$\;
Initialize $\mathcal{A}_{\text{Nov}} = \emptyset$\;
Initialize dataset buffer $DS = \emptyset$\;
\While{$Bud$ not depleted}{
    Generate offsprings $\Gamma^m_{g} \leftarrow \mathbb{V}(\Gamma_{g})$\;

    \For{$\theta_i \in \Gamma_{g} \bigcup  \Gamma^m_{g}$}{ 
        Evaluate policy $\pi(\theta_i) \rightarrow o_T$\;
        Calculate outcome descriptor $E(o_T^{(\theta_i)}) = b_i$\;
        Calculate surprise $s(\theta_i) = \big|\big|o_T^{(\theta_i)} - D\big(E(o_T^{(\theta_i)})\big)\big|\big|^2.$\;
        Store outcome observation $o_T \rightarrow DS$\;
    }
    Calculate novelty $\eta(\theta_i) = \frac{1}{|J|}\sum_{j \in J}\text{dist}(b_i, b_j), ~~ \forall\theta_i \in \Gamma_g \bigcup \Gamma^m_g$\;
    $\mathcal{A}_{\text{Nov}} \leftarrow N_Q \text{ most novel from } \Gamma^m_g$\; 
    Substitute $N_Q$ less novel $\theta_i \in \Gamma_g$ with $N_Q$ most novel in $\theta_i \in \Gamma^m_g$\;
    \If{$g$ multiple of $I$}{
        Train $D(E(\cdot)) \leftarrow DS$\;
        $DS \leftarrow \emptyset$\;
        Update stored policies' descriptors $b_i, ~ \forall\theta_i \in \Gamma_g \bigcup \mathcal{A}_{\text{Nov}}$\;
   }
    $g = g +1$\;
 }
\end{algorithm}

\subsection{Search and Training}
\label{sec:tax_training}
Similarly to \gls{ns}, the repertoire of diverse policies is built iteratively. 
At each iteration, a set of $M$ new policies, parametrized by $\theta \in \Theta$, is generated by modifying the ones from the previous iteration.
More precisely, the $Q$ best policies, according to the metric (novelty or surprise), are duplicated to replace the $Q$ worst ones. Then the parameters $\theta$ of all $M$ policies are perturbed by adding gaussian noise with probability $p_d$.
Moreover, in the process, the $Q$ best policies $\theta_i$ are also stored in the repertoire, along with their final observation $o_T^{(\theta_i)}$.

The \gls{ae} is trained to minimize the reconstruction error by feeding it the observations generated during the policies evaluation. 
In particular, the final observations $o_T$ are stored for $I$ iterations (for a total of $M\times I$ observations) before the \gls{ae} is trained for $J$ epochs.
This buffering step helps in collecting enough data for the training process of the \gls{ae}.\\
Note that, because the outcome space changes during the training of the \gls{ae}, the policies in the repertoire are reassigned an updated outcome descriptor after each training episode of the \gls{ae}, obtained by feeding the associated final observation to the current version of the \gls{ae}'s encoder.

The \gls{taxons} search process is described in detail in algorithm \ref{alg:tax_training}.
\section{Experiments}
\label{sec:tax_experiments}
\subsection{Experimental setup}
\begin{figure}[h]
\centering
\includegraphics[width=\textwidth]{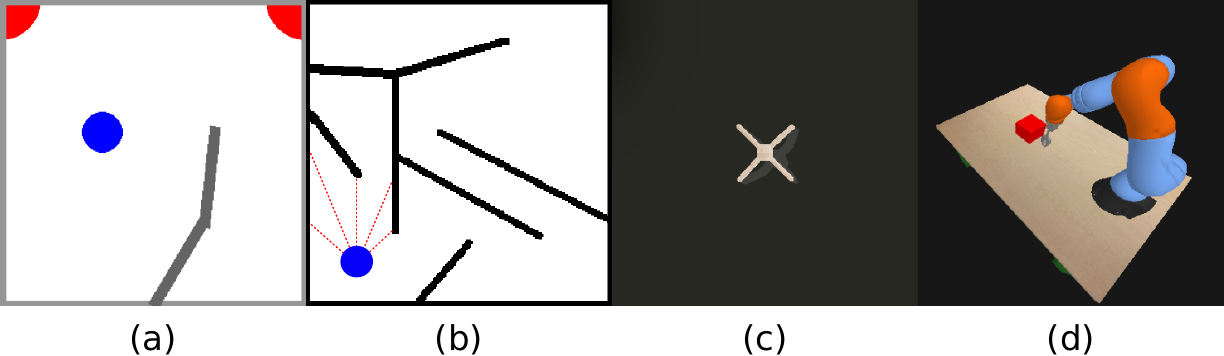}
\caption[TAXONS test environments]{The four different experimental environments.}
\label{fig:tax_envs}
\end{figure}

To test if \gls{taxons} can drive exploration by using only high-dimensional RGB images, it has been tested on four different environments:
\begin{itemize}
    \item \textbf{Billiard}: a 2-jointed arm pushing a ball in a 2D room, shown in figure \ref{fig:tax_envs}.(a);
    \item \textbf{Maze}: a two wheeled robot navigating a 2D maze \cite{Lehman2008NS}, shown in figure \ref{fig:tax_envs}.(b);
    \item \textbf{Ant}: a four legged robot ant moving on the floor \cite{Todorov2012mujoco}, represented in figure \ref{fig:tax_envs}.(c);
    \item \textbf{Kuka}: a 7-jointed Kuka robotic arm, simulated in Pybullet \cite{coumans2019PyBullet}, that has to learn how to push a box on a table. The setup is shown in figure \ref{fig:tax_envs}.(d).
\end{itemize}
In all of these situations, driving exploration through images removes the need to track the objects in order to extract the internal states used by \gls{ns} to drive the search, e.g. the ball or the robot position.

The scenarios are observed through an RGB-camera looking at the scene from the top, with the exception of the Kuka environment that is observed from the side. 

\gls{taxons} is compared against five different baselines:
\begin{itemize}
    \item \textbf{NS}: a vanilla novelty search algorithm \cite{Lehman2008NS}. This algorithm works with features that are hand-crafted using a priori knowledge about the agent and environment;
    \item \textbf{PNS}: a novelty search algorithm, where the outcome space directly corresponds to the parameter space $\Theta$ of the policies. 
    The outcome descriptor characterizes the policy but not the final observation. 
    The idea behind this baseline is to verify if diversity in the policy space reflects in diversity in the behavior space;
    \item \textbf{RNS}: a novelty search algorithm where the outcome description of each policy is randomly sampled in a $10$D space. The outcome descriptor does not characterize the observation nor the policy. The reason for using this baseline is to verify how important an accurate representation of the behavior is when performing exploration thorugh \gls{ns};
    \item \textbf{RS}: a random search in which all policies are randomly generated and randomly selected to be added to the repertoire. The idea behind the selection of this baseline is to analyze if a random search algorithm can compete with the exploration abilities of \gls{ns} based algorithms;
    \item \textbf{NT}: a novelty search algorithm in which the outcome space corresponds to the feature space of a \gls{ae} whose weights are randomly generated at the beginning of the search and left unmodified during the whole search process. As with \gls{taxons}, the \gls{ae} is fed only with the last observation $o_T$ of the environment.
    This baseline has been chosen to test the importance of the training process of the \gls{ae}.
\end{itemize}

The vanilla version of \gls{taxons} is also compared against two ablated versions:
\begin{itemize}
    \item \textbf{TAXO-N}: in which only the novelty calculated in the learned feature space is used as selection metric;
    \item \textbf{TAXO-S}: in which only the reconstruction error of the \gls{ae} is used as selection metric.
\end{itemize}

\begin{figure}[h]
\centering
\includegraphics[width=\textwidth]{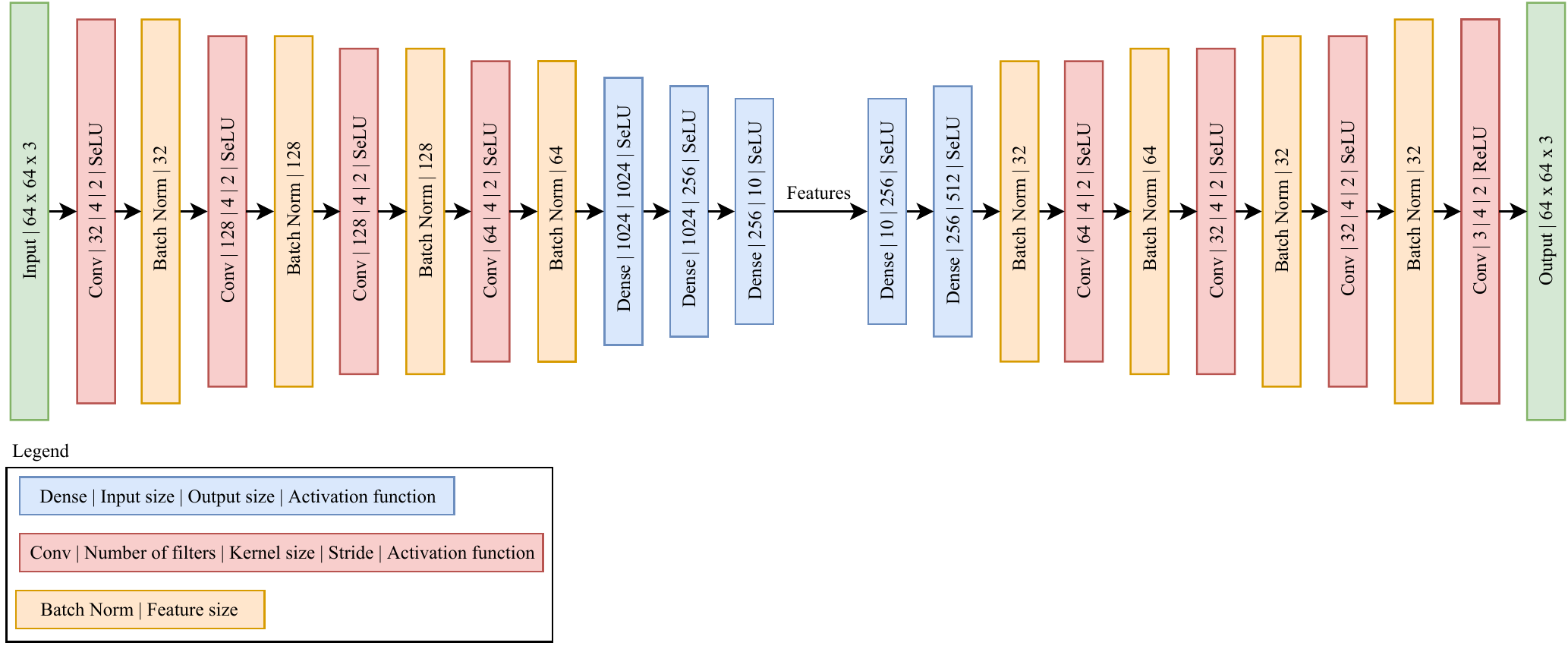}
\caption[TAXONS AE structure]{\gls{ae} structure. The input and output of the network are represented in green; in red are the convolutional layers, with the associated batch normalization operation highlighted in orange. In blue are the fully connected layers.}
\label{fig:tax_ae}
\end{figure}

All experiments have been conducted on a population of $M=100$ policies for each iteration. 
The novelty of each policy is calculated by using a value of $k=15$ neighbours in Eq.\eqref{eq:ns_novelty}, as proposed by Gomes et al. \cite{Gomes2015NSHyperpar}, with the $N_Q=5$ best policies added to the repertoire. 
At each iteration, the parameters $\theta_i$ of each policy are independently perturbed, with probability $p_d = 0.2$, by adding noise sampled from $\mathcal{N}(0, 0.05)$.
The observations $o_T$ consist of RGB images of size $64 \times 64 \times 3$.
The \gls{ae} consists in an encoder $E$ with $4$ convolutional layers, of sizes [$32$, $128$, $128$, $64$] and $3$ fully connected layers, of sizes [$1024$, $256$, $10$]; followed by a decoder $D$ with $2$ fully connected layers, of sizes [$256$, $512$], and $4$ convolutional layers of sizes [$64$, $32$, $32$, $3$]. 
For the convolutional operations, the kernel is of size $4$ and a the stride of size $2$ with padding of $1$.
Moreover, in order to improve the network performances a batch normalization operation \cite{ioffe2015batch} is applied after each convolutional layer.
The activation functions used are SeLU \cite{Klambauer2017selu} for every layer, except for the last layer of the decoder, in which a ReLU activation is used to force the non-negativity of the output values. 
The structure of the \gls{ae} is shown in figure \ref{fig:tax_ae}.
Note that the decoder is smaller than the encoder in order to prevent it from generalizing too well.
By having a decoder that is too powerful, after few training iterations, the reconstruction error will be small also for images that have not been seen yet.
This would reduce the effect of the surprise metric.
\\
The training is done every $I=30$ search iterations for $J=5$ epochs, with a learning rate of $0.001$. 
The optimizer used is Adam\cite{kingma2014adam}.

\subsection{Evaluation}
The goal of \gls{taxons} is to produce diverse policies.
In light of this, the algorithms are compared based on how well they cover the ground-truth outcome space of the system.
By design, this ground-truth outcome space corresponds to the $(x,y)$ position of the ball for the Billiard environment, of the center of the robot for the Maze and Ant environment, and of the box on the table for the Kuka arm environment. 
The \emph{coverage} metric is thus defined as the percentage of this $(x, y)$ space reached by the final repertoire of policies. 
This is done by dividing this space in a $50 \times 50$ grid and then calculating the ratio of number of cells reached at least once over the total number of cells.
\\
Note that the ground-truth $(x, y)$ space is unknown to the methods (except for \gls{ns}) and is only used a posteriori to compare them.

Moreover, to evaluate the statistical significance of the results, each experiment was run 20 times on different random seeds, and the results compared by performing a Mann-Whitney test \cite{mann1947test}, with Holm-Bonferroni correction \cite{holm1979simple}.

The evolution of the coverage over the training for the different methods is displayed in Fig. \ref{fig:tax_results}.(a) and the final coverage comparison is displayed in Fig. \ref{fig:tax_results}.(b).

\subsubsection{Billiard environment}
As illustrated in Fig. \ref{fig:tax_envs}.(a), the  agent  consists  in  a  two-jointed arm, depicted in gray, that can push a blue ball inside a squared room. 
Two additional corners are depicted in red; in the absence of a task they have no specific function in the simulation.
The policy controlling the speed of each joint of the agent is defined by a fifth-degree polynomial \gls{dmp} \cite{Schaal2006DynamicMP}. 
The policy is run for a time horizon of $500$ steps.
In order to remove any possible distractor \cite{stone2021distracting}, the arm is brought out of the experimental table at the end of the evaluation of each policy.
As shown in Fig. \ref{fig:tax_exp}.(b), the final observation $o_T$ consists in a top-view of the experimental table. 
Note that, for the \gls{ns} baseline the $(x,y)$ ground-truth position of the ball is used as outcome descriptor.
\\
The search methods are run for $2000$ evaluations.

\subsubsection{Maze environment}
As illustrated in Fig. \ref{fig:tax_envs}.(b), the agent consists in a two-wheeled robot, depicted in blue, navigating in a maze \cite{Lehman2008NS}. 
The agent is equipped with $5$ distance sensors in the front, represented by the dotted lines.
The policy controlling the speed of each wheel of the agent is defined by a $2$-layers, fully connected, neural network that takes as input the robot sensors readings. 
The policy is run for a time horizon of $2000$ steps. 
As shown in Fig. \ref{fig:tax_exp}.(b), the final observation $o_T$ consists in a top-view of the maze and the agent. 
Note that, for the \gls{ns} baseline the $(x,y)$ ground-truth position of the robot is used as outcome descriptor.
\\
The search methods are run for $1000$ evaluations.

\subsubsection{Ant environment}
As illustrated in Fig. \ref{fig:tax_envs}.(c), the agent consists in a four-legged ant robot \cite{Schulman2015AntEnv}, moving in a $2$D plane of size $3m\times 3m$.
The policy controlling the torque of each of the $8$ agent joints is defined by a sinusoidal \gls{dmp}.
The experiment is run for a time horizon of $500$ steps or until the robot reaches the borders of the plane. 
As shown in Fig. \ref{fig:tax_exp}.(b), the final observation $o_T$ consists in a top-view of environment. 
Note that, for the \gls{ns} baseline the $(x,y)$ ground-truth position of the robot is used as outcome descriptor.
\\
The search methods are run for $500$ evaluations.

\subsubsection{Kuka environment}
\begin{figure}[!p]
\centering
\includegraphics[height=\dimexpr
  \textheight-5\baselineskip-\parskip-.2em-
  \abovecaptionskip-\belowcaptionskip\relax]{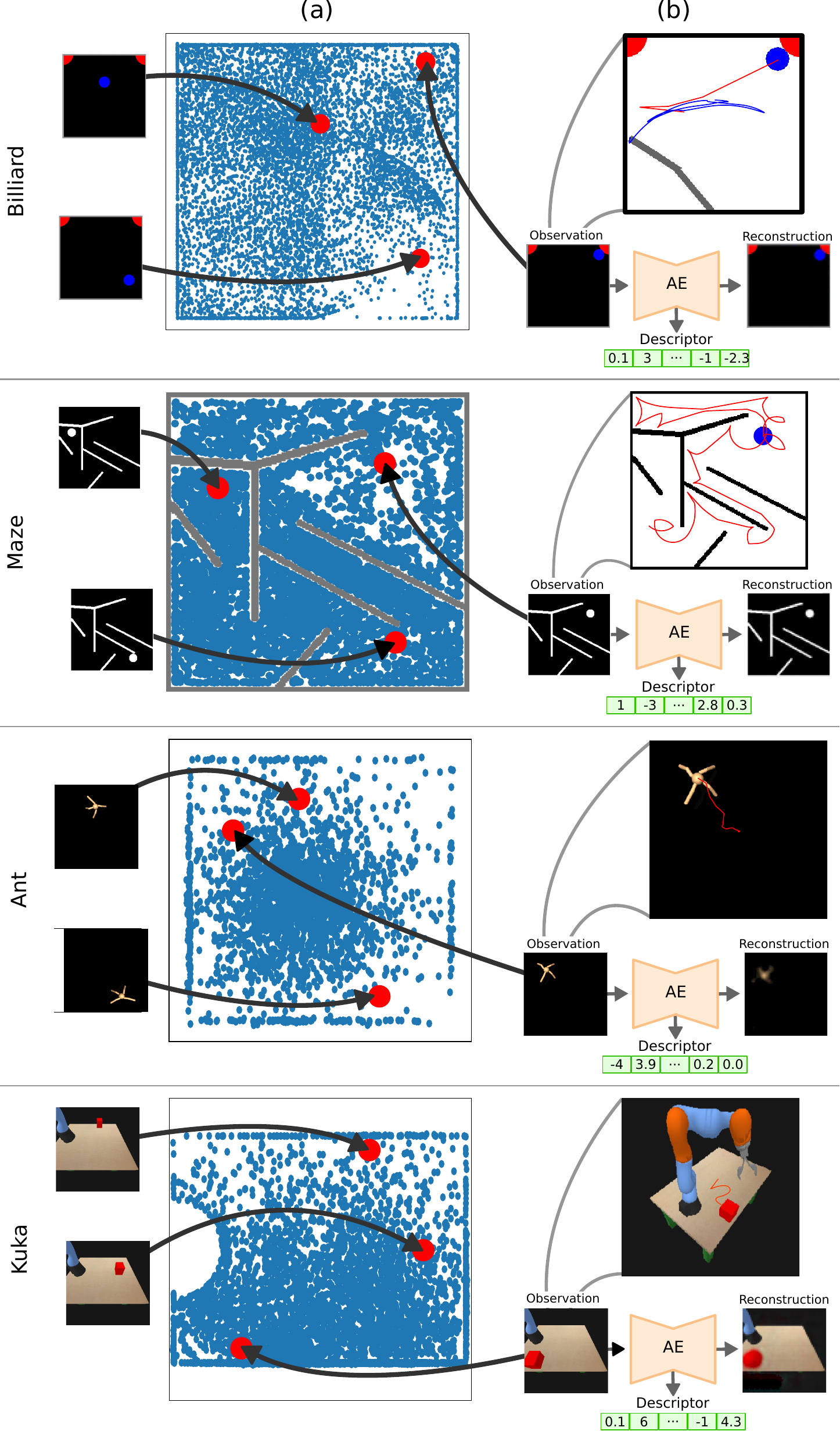}
\caption[TAXONS final archives]{(a) Coverage of the repertoire of policies found by \gls{taxons} in the ground-truth $(x,y)$ space. Highlighted in red are 3 policies for which the related final observation $o_T$ are shown. (b) Sample policy from the repertoire generated by \gls{taxons}. The trajectories followed by the policies are highlighted in red; for the Billiard, the trajectory of the point of the arm is also highlighted in blue. The final observation of the trajectory, with the respective reconstruction and descriptor generated by the \gls{ae}, are also shown.}
\label{fig:tax_exp}
\end{figure}

As illustrated in Fig. \ref{fig:tax_envs}.(d), the agent consists in a 7-jointed Kuka robotic arm simulated in PyBullet \cite{coumans2019PyBullet}. 
The arm can move thanks to a joint position controller and can push a red cube on a table. 
The policy consists of a sequence of $5$ points in the 7-dimensional joint space that need to be reached in a time interval of $2000$ timesteps.
Moreover, at the end of the policy execution the arm is reset to the initial position to prevent any obstruction of the view of the cube and to reduce the factors of variations in the observation.
As shown in Fig. \ref{fig:tax_exp}.(b), the final observation $o_T$ consists on a lateral view of the table on which the cube rests. 
Note that, for the \gls{ns} baseline the $(x,y)$ ground-truth position of the cube on the table is used as outcome descriptor.
\\
The search methods are run for $1000$ evaluations.

\subsection{Results}
\label{sec:tax_results}
\begin{figure}[!p]
    \centering
    \includegraphics[width=\textwidth]{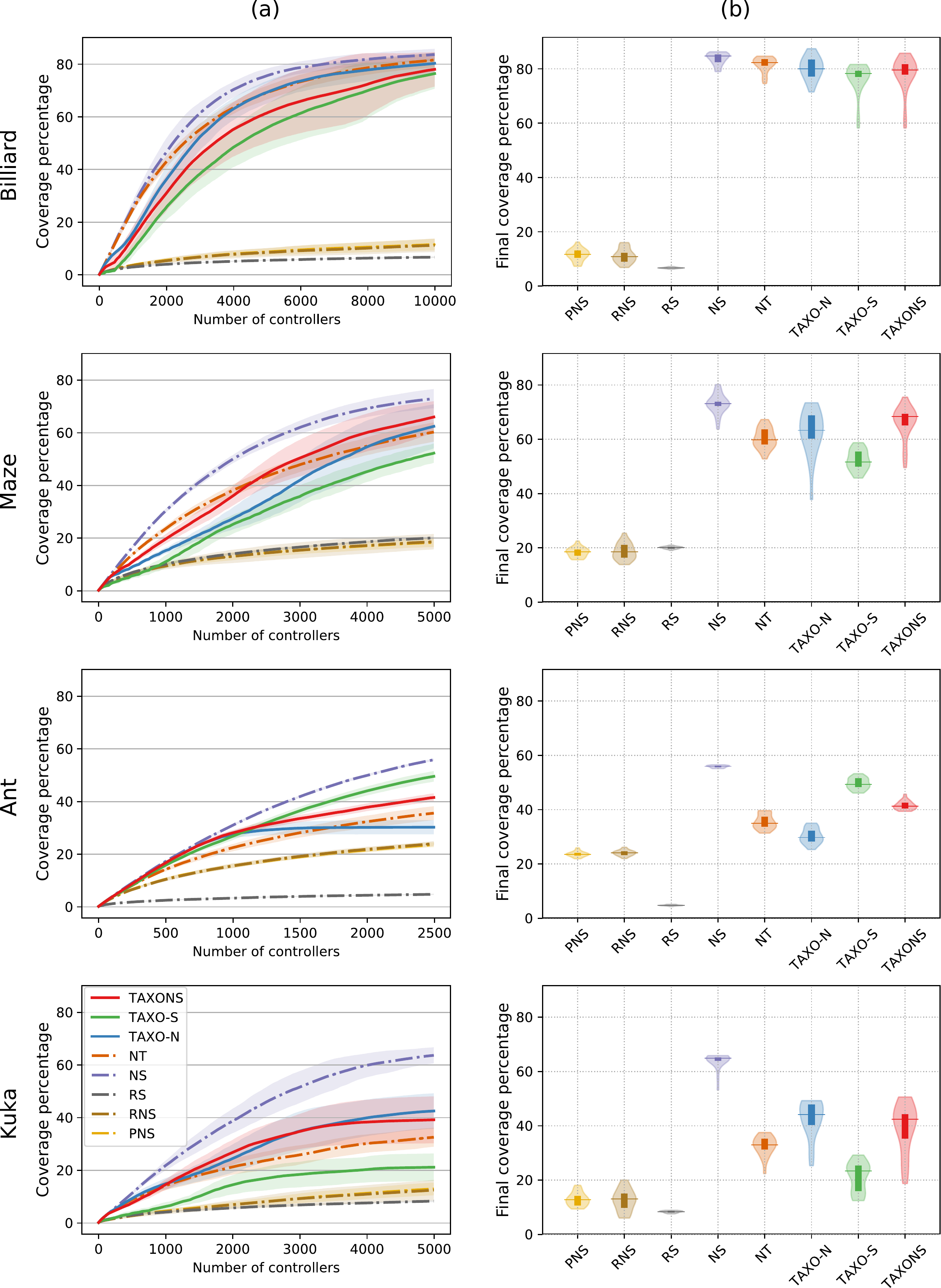}
    \caption[TAXONS coverage results]{(a) Evolution of coverage metric over the number of policies in the repertoire. Note that the PSN and RNS curves are overlapping.
    (b) Final coverage measures for the tested methods.}
    \label{fig:tax_results}
\end{figure}

The results displayed in Fig.\ref{fig:tax_results} show that \gls{taxons} leads to a good coverage of the ground-truth $(x,y)$ outcome space. 
Its performance is lower than the upper-bound performance of \gls{ns}, which has direct access to the ground-truth outcome space, but significantly higher than the other baselines, which use as outcome descriptor the policy parameters (PNS), a random vector (RNS), or no outcome descriptor at all (RS). 
This shows that i) performing \gls{ns} in a low-dimensional outcome space capturing informations about the final state of the system (through the last observation) is beneficial, and ii) that \gls{taxons} can successfully build this space. 
Indeed when the generation and selection of policies is purely random (RS) the coverage is very low. 
Similarly, when low-dimensional outcome descriptors are randomly assigned to the policies (RNS) the coverage is only slightly better than purely random (Ant), or as bad (Billiard, Maze and Kuka). 
Finally, performing the \gls{ns} directly in the high-dimensional policy parameters space $\Theta$ (PNS), leads to a coverage that is similar to the RNS case. 
This suggests that, in these different experiments, performing the search in the high-dimensional policy parameter space is equivalent to assigning random descriptors to the policy, meaning that optimizing for diversity in the policy space does not help in exploring the behavior space.
\\
In contrast, the performance of \gls{taxons} is significantly higher and more consistent in the four experiments (Billiard: $p=3.38\times 10^{-8}$, Maze: $p=3.38\times 10^{-8}$, Ant: $p=1.6\times 10^{-7}$, Kuka: $p=3.94 \times 10^{-8}$), showing that by learning the outcome space through an \gls{ae} it is possible to capture relevant information about each system.
The final performance of \gls{taxons} 
(Billiard: $78.03$, Maze: $66.02$, Ant: $41.55$, Kuka: $39.2$) 
even approaches that of \gls{ns} 
(Billiard: $83.66$, Maze: $72.99$, Ant: $55.88$, Kuka: $63.9$) 
although it remains inferior 
(Maze:  $p=3.69\times 10^{-5}$, Ant:  $p=7.65\times 10^{-8}$, Kuka: $p=3.39 \times 10^{-8}$), especially for the Kuka environment, in which the skewed perspective requires the \gls{ae} to encode also informations about the changing size of the cube rather than only its positions, rendering the task more complex, as the size covaries with the position.
It must be highlighted that NS has direct access to the ground-truth $(x,y)$ space, thus guaranteeing a very good performance.

Surprisingly, the NT baseline, in which the \gls{ae} is never trained, performs similarly to \gls{taxons} in the Billiard environment and only slightly worse in the other settings (Maze: $p=2.5 \times 10^{-4}$, Ant: $1.44 \times 10^{-7}$, Kuka: $p=1.5 \times 10^{-3}$).
This is probably due to the intrinsic power of convolutional neural networks in extracting relevant features in their inputs even when no training has been done, as shown in \cite{saxe2011random}, and can be an interesting direction of investigation.

The performance of the two ablated versions (TAXO-N and TAXO-S) is similar to the one of \gls{taxons}, as they lay between the NS upper-bound and the PNS, RNS and RS baselines. 
Nonetheless, their efficiency varies between experiments. 
TAXO-S performs similarly to \gls{taxons} in the Billiard environment ($p=0.036$), worse in the Maze ($p=1.53\times 10^{-6}$) and Kuka environments ($p=1.34 \times 10^{-6}$) and better in the Ant one ($p=7.69\times 10^{-8}$). 
On the other hand, TAXO-N performs similarly to \gls{taxons} in the Billiard, Kuka and Maze environments, while being significantly worse in the Ant one ($p=7.67\times 10^{-8}$). 
After investigation, we hypothesize that the low performance of TAXO-N in the Ant environment is due to the specific dynamics of the \gls{ae}. 
In the first phase of the training, the \gls{ae} learns to reconstruct the large body of the agent while disregarding its legs. 
This leads to the outcome space temporarily capturing informations about the position of the agent in the $(x,y)$ space, and thus allowing novelty search to better cover the ground-truth space. 
In a second phase, the \gls{ae} learns how to reconstruct the legs. 
This leads the algorithm on exploring also different legs arrangements, rather than fully focusing on only covering the $(x,y)$ space.
TAXO-S performs significantly better in the same environment, as the impact of the body position on the reconstruction error is greater than the one of the legs. 
Thus maximizing the surprise also leads to maximizing the coverage.

From the results, it is possible to see that while the performances of the novelty and surprise are dependent of the dynamics of the environment, combining them renders \gls{taxons} more robust than its two ablated versions.
This allows performances almost as good as the ones of \gls{ns}. 
\section{Conclusion}
\label{sec:tax_conclusion}
Studying the problem of sparse rewards, alongside the study of \gls{qd} and divergent algorithms led to focusing the research effort towards the development of an algorithm that could learn and explore a search space with minimal supervision.
\gls{taxons} can generate a repertoire of diverse policies, without any external reward and with minimal prior knowledge about the system by exploring the space of reachable outcomes available. 
It does so by taking advantage of the exploration power of \gls{ns} in a low-dimensional outcome space learned online by an \gls{ae} trained directly on RGB images observations collected during the search.
Using these observations to autonomously learn the search space renders \gls{taxons} easier to apply to different situations and tasks for which is not clear what the important features are.

The approach was tested on four different simulated environments and compared against different baselines.
The results show that, by maximizing both novelty in the learned outcome space and surprise, derived from the \gls{ae}'s reconstruction error, \gls{taxons} finds a set of policies that covers the ground-truth outcome space, while being robust to different environments. 
At the same time, autonomously learning a \gls{bs} means that the designer has no control on which information will be embedded in the representation.
This can be a problem in more complex environments in which multiple aspects need to be represented in order to perform proper exploration.
Given its nature, the algorithm will mainly focus on the most obvious aspects, while ignoring the smaller, but possible important ones.
Moreover, the presence of \emph{distractors} - elements not related to the ground-truth space to explore - can lead the algorithm to learn a poor representation.
These and other limitations of similar methods will be discussed in depth in Chapter \ref{chap:discussion}, together with the proposition of strategies to deal with them.

As said, the main objective of \gls{taxons} is to explore as much as possible, but this is only a first step towards a more complete approach to sparse rewards problems.
Once exploration is over, and rewards have been discovered, the algorithm should be able to take advantage of those rewards.
At the same time, to address this problem, good exploration algorithms requiring minimal designer prior knowledge are fundamental.
While \gls{taxons} can efficiently explore and learn a low-dimensional representation of an unknown outcome space, it is still strongly limited by the restriction of working only on the last observation of the trajectory.
This requirements limits the application of the algorithm only to environments whose last observation is informative enough with respect to the task.
There are many ways in which this problem can be approached, some more naive than others.
The next chapter will analyse one of these possible approaches, the \emph{signature transform}, comparing its effectiveness to some simpler baselines. 
\clearpage\null\thispagestyle{empty}

\chapter{Signatures}
{\hypersetup{linkcolor=black}\minitoc}
\label{chap:signatures}
\section{Introduction}
\label{sec:sign_intro}
Chapter \ref{chap:taxons} discussed a big limitation that \gls{qd} algorithms, and \gls{ns} in particular have: the \gls{bs}, that is the space in which the behavior of the policies is represented and the diversity is measured, needs to be hand-designed.
To address this problem, \gls{taxons} was introduced.
This is an algorithm that can learn, online and in an unsupervised way, a \gls{bs} representative of the outcome of each policy, allowing to reduce the amount of prior information and the design effort needed to solve a task.
Notwithstanding the promising results of \gls{taxons}, the method still relies on a strong assumption: the last observation of a trajectory should contain enough information to describe the behavior of a policy.
In situations in which the most informational moment of the policy behavior happens at an unspecified time during the evaluation, the exploration process can be less efficient, preventing the discovery of good solutions.
An example of this can be a robot trying to launch a ball against a wall, from which it can bounce back.
The final position reached by the ball is not informative of the position in which it hit the wall.
To properly explore, the method should take into account the moment in which the ball hits the wall, but this is difficult to know.
This makes the maximization of the diversity of these positions complicated.
For this reason, in these situations the algorithm should be able to push for diversity along the whole trajectory of traversed states, not only on the last one.

Pushing for diversity over the space of trajectories requires a way to calculate a distance between different trajectories.
There are two strategies that can be followed.
The first one consists in calculating the distance between the whole trajectories, either state-by-state, or through the \emph{dynamic time warping} method \cite{muller2007dynamic}.
The calculation of the distance state-by-state is not always feasible because the trajectories can be of different lengths.
At the same time the dynamic time warping can be extremely slow when comparing distances between thousands of trajectories, as it happens for \gls{ns} based methods.
The other possible approach consists in first obtaining a compressed representation, or encoding, of the trajectories and then calculating the distance on said compressed representations.
This has the advantage of reducing the dimensionality of the space in which the distance is calculated, preventing many of the issues related to distance metrics in high-dimensional spaces \cite{Aggarwal2002distance}.
However, these approaches add the additional complexity of calculating the compressed representation of the trajectory.
There are multiple way to do so, ranging from a naive sub-sampling of the trajectory, to more sophisticated but also more computationally intensive machine learning approaches like \gls{lstm} encoders \cite{hochreiter1997long}.

In this chapter, another method to perform the trajectory encoding will be analyzed: the \emph{signature transform}.
This transform, introduced by Chen \cite{chen1954iterated, chen1957integration, chen1958integration}, is a mathematical object defined in the scope of rough field theory.
It allows to describe a stream of data as a single vector, the signature, obtained through an infinite series of integrals.
In practice the signature is calculated up to a certain order of the infinite series of integrals, proving a quick and easy way to compress a sequence of data to any desired precision.
This, and the other useful properties that the signature transform enjoys, have attracted the attention of the machine learning community over this method with the goal of encoding sequences of data in a fast and easy way \cite{bonnier2019deep, fermanian2021embedding}.

The signature transform can be used to describe the policies behaviors.
This is done by compressing the whole trajectory of states traversed by the system during the policy evaluations into a smaller behavior descriptor vector.
The goal is to obtain more diverse solutions over the whole space of trajectories, rather than only over the final outcome.
Moreover, this approach will reduce even more the amount of prior information needed at design time, allowing the application of \gls{qd} methods to new and more complex domains.

The chapter is structured as follows.
Sec. \ref{sec:sign_method} will present the signature transform, while Sec. \ref{sec:sign_bd} will describe how it can be applied to \gls{ns}.
The testing environment will be presented in Sec. \ref{sec:sign_exp} and the results discussed in Sec. \ref{sec:sign_results}.
The chapter will conclude with Sec. \ref{sec:sign_concl} in which the lessons learned during this analysis will be discussed.

\section{The signature transform}
\label{sec:sign_method}
This section will describe in a formal way what the signature transform of a sequence of data is and what its properties are.

Let's start by noting that a sequence of data $\mathrm{x} = (x_0, \dots, x_T)$ of length $T$, where $x_i \in \mathbb{R}^d$, can be described by a continuous function:
\begin{equation}
\label{eq:sign_path_func}
    f: [0,1] \rightarrow \mathbb{R}^d, ~~ \text{such that}~~ f\Big(\frac{i}{T}\Big) = x_i \in \mathbb{R}^d.
\end{equation}
Moreover, the function $f$ can also be represented as the list of its components along each dimension:
\begin{equation}
    f(t_i) = \left\{f^1(t_i), \dots, f^d(t_i)\right\},
\end{equation}
with $t_i = \nicefrac{i}{T}$.
Thanks to this function, the study of a sequence of data $\mathrm{x}$ can be viewed as the geometrical study of the path of $f(\cdot)$ in the space $ \mathbb{R}^d$.
For example the changing temperature of a room during the day can be seen as a path in $\mathbb{R}$; the motion of an ant going back to its nest as a path in $\mathbb{R}^2$; the changes within $d$ indices of financial markets as a path in $\mathbb{R}^d$; etc. \cite{bonnier2019deep}.

These functions are called \emph{paths} in the field of rough field theory.
The goal of the signature transform is to study the geometrical characteristic of these paths.
This is done by the iterated application of integrals on the function, in order to obtain a vector encoding a collection of geometrical statistics of $f(\cdot)$.
This vector is called \emph{signature}.
The method allows to extract an accurate summary of the path and to obtain arbitrarily good linear approximations of continuous functions of paths \cite{hairer2013solving}.

The main assumption needed in order to calculate the signature of these functions is that these paths are of \emph{bounded variation}, i.e. the first derivative of the function exists almost everywhere.
More formally, a function $f$ is of bounded variation if its total variation is finite.
The total variation is expressed as:
\begin{equation}
    ||f||_{1-var} = \sup_D \sum_{t_i \in D}|f(t_i) - f(t_{i-1})|,
\end{equation}
where the supremum is calculated over all the finite partitions:
\begin{equation}
    D = \{(t_0, \dots, t_T) | T \geq 1, 0=t_0 < t_1< \dots < t_{T-1} < t_T =1\},
\end{equation}
with $t_i = \nicefrac{i}{T}$.

To define the signature, let's start by considering the simplest case of a path for which $d=1$.
In this case the path would be $f(t_i) = \left\{f^1(t_i)\right\}$, for which its integral would be:
\begin{equation}
\label{eq:sign_1_fold}
    \mathcal{S}(f)^1_{0, T} = \int\limits_{0\leq t \leq T}df^1(t) = f^1(T) - f^1(0). 
\end{equation}
This integral is what it is called the \emph{1-fold iterated integral} of the path and is a real valued integral representing the increment, or the displacement, of this one-dimensional path over the whole time interval \cite{yang2017developing}.

The \emph{2-fold iterated integral} of the path can be obtained by reapplying the same integral operator:
\begin{equation}
\label{eq:sign_2_fold}
     \mathcal{S}(f)^{1,1}_{0, T} = \int\limits_{0 < t_2 \leq T} \mathcal{S}(f)^1_{0, t_2} df^1(t_2) = \frac{1}{2} \left(f^1(T) - f^1(0)\right)^2,
\end{equation}
which represents the square of the increment of the path over the time interval.
The recursive application of this operator leads to the definition of the \emph{k-fold iterated integral},
which is proportional to the increment of the path to the power of $K$:

\begin{dmath}
    \mathcal{S}(f)^{I_K}_{0, T} = \int\limits_{0<t_K\leq T} \cdots \int\limits_{0<t_2 \leq t_3} \int\limits_{0<t_1 \leq t_2} df^1(t_1) ~ df^1(t_2)\cdots df^1(t_K) = \frac{1}{K!}\left(f^1(T) - f^1(0)\right)^K, 
\end{dmath}
where $I_K = (1, 1, \dots, 1)$ is the set of indexes of size $K$. 

\begin{figure*}[!h]
    \centering
    \includegraphics[width=\textwidth]{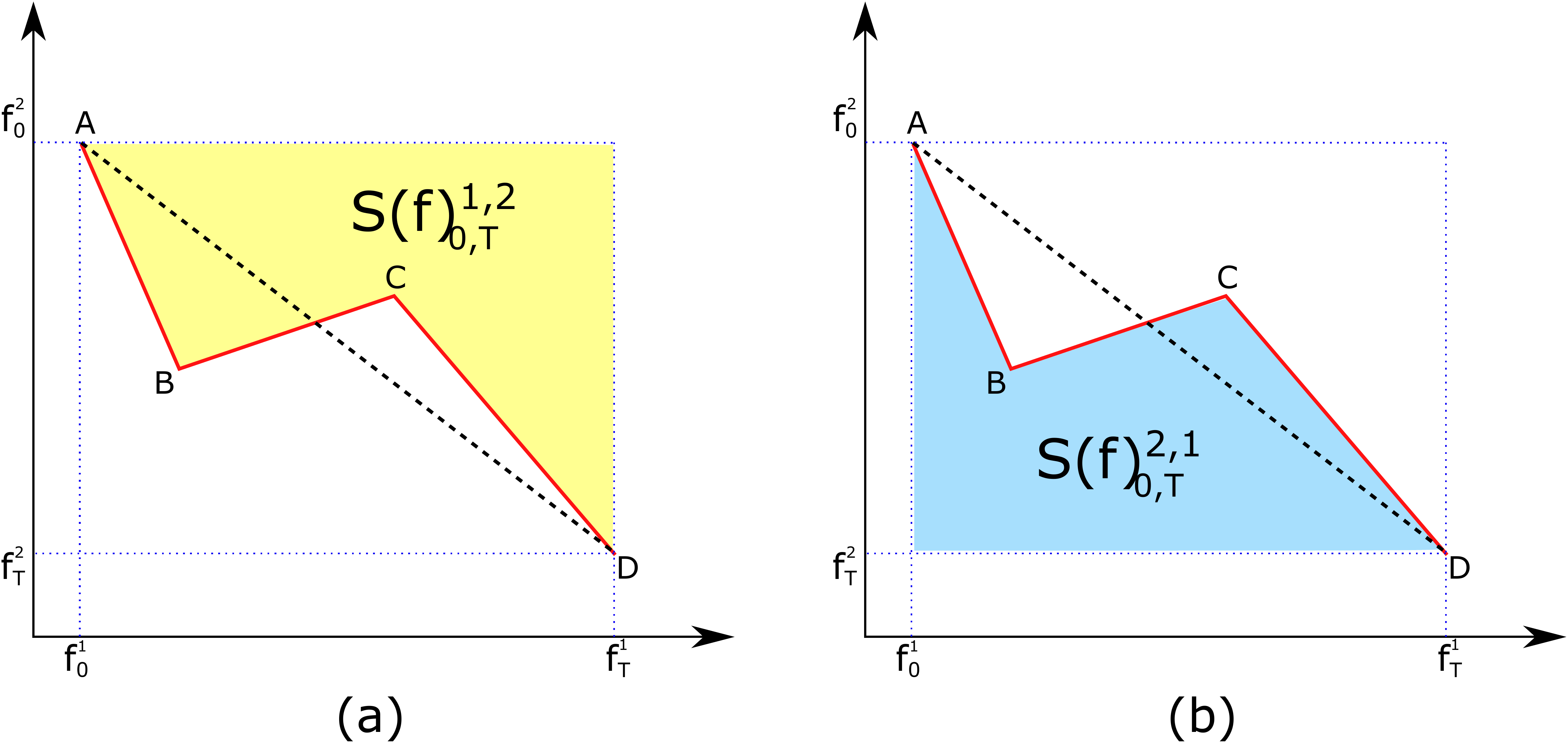}
    \caption[Geometric representation of signature]{Geometric intuition of the 2-fold iterated integral of a 2D path as shown by Yang et al. \cite{yang2017developing}. The path in red moves from point A to D during a time interval of $[0, T]$. The chord connecting the endpoints is shown as a dashed line. Figure (a) and (b) show the last two elements of the 2-fold integral of the path.
    }
    \label{fig:sign_levy}
\end{figure*}
In the case of a path of $d=2$, the path would be $f(t_i) = \left\{f^1(t_i), f^2(t_i)\right\}$.
This leads to a 1-fold iterated integral containing two elements, the same as in Eq. \ref{eq:sign_1_fold} but for each single dimension:
\begin{align}
    \mathcal{S}(f)^1_{0, T} = \int\limits_{0\leq t \leq T}df^1(t) = f^1(T) - f^1(0), \\
    \mathcal{S}(f)^2_{0, T} = \int\limits_{0\leq t \leq T}df^2(t) = f^2(T) - f^2(0). 
\end{align}
At the same time, the 2-fold iterated integral will contain $d^2 = 4$ elements, due to the cross-terms between the dimensions:
\begin{gather}
     \mathcal{S}(f)^{1,1}_{0, T} = \int\limits_{0 < t_2 \leq T} \mathcal{S}(f)^1_{0, t_2} df^1(t_2) = \frac{1}{2} \left(f^1(T) - f^1(0)\right)^2, \\
      \mathcal{S}(f)^{2,2}_{0, T} = \int\limits_{0 < t_2 \leq T} \mathcal{S}(f)^2_{0, t_2} df^2(t_2) = \frac{1}{2} \left(f^2(T) - f^2(0)\right)^2,\\
      \mathcal{S}(f)^{1,2}_{0, T} = \int\limits_{0 < t_2 \leq T} \int\limits_{0<t_1 \leq t_2} df^1(t_1) ~ df^2(t_2),\\
      \mathcal{S}(f)^{2,1}_{0, T} = \int\limits_{0 < t_2 \leq T} \int\limits_{0<t_1 \leq t_2} df^2(t_1) ~ df^1(t_2).
\end{gather}
The first 2 correspond to the square of the displacement caused by the path for each dimension, as in Eq. \ref{eq:sign_2_fold} for the 1 dimensional path.
At the same time, $\mathcal{S}(f)^{1,2}_{0, T}$ and $\mathcal{S}(f)^{2,1}_{0, T}$, shown respectively in Fig. \ref{fig:sign_levy}.(a) and Fig. \ref{fig:sign_levy}.(b), encode information about the area covered by the path.

This example can be generalized to the \emph{k-fold integral} of a path in $\mathbb{R}^d$.
Such an integral will have $d^K$ components that can be expressed as:
\begin{equation}
    \mathcal{S}(f)_{0, T}^{I_K} = \int\limits_{0<t_K\leq T} \cdots \int\limits_{0 < t_2 \leq t_3} \int\limits_{0<t_1 \leq t_2} df^{i_1}(t_1) ~ df^{i_2}(t_2) \cdots df^{i_K}(t_K),
\end{equation}
where $I_K = (i_1, \dots, i_K)$ is the set of indexes of size $K$, with $i_j \in \{1, \dots, d\}$.

Finally, the signature of a path $f(\cdot)$ can be defined.
This will be the collection of all the iterated integrals over the path and can be expressed as:
\begin{dmath}
    \label{eq:signature}
    \text{Sig}(f)_{0,T} = \left(1, \mathcal{S}(f)_{0, T}^{1}, \dots, \mathcal{S}(f)_{0, T}^{d},\\
    \mathcal{S}(f)_{0, T}^{1,1}, \dots, \mathcal{S}(f)_{0, T}^{1, d}, \mathcal{S}(f)_{0, T}^{2, d}, \dots, \mathcal{S}(f)_{0, T}^{d, d},\\
     \dots,\mathcal{S}(f)_{0, T}^{1,1,\dots,1}, \dots, \mathcal{S}(f)_{0, T}^{d,d,\dots,d}, \dots \right),
\end{dmath}
where the first element is set to 1 as a convention.
Being calculated over all possible combination of indices of finite length, the signature is a vector of infinite size.
In practice, it has to be truncated up to a certain level, or \emph{order}, $K$:
\begin{equation}
    \label{eq:sign_trunc_signature}
    \text{Sig}^K(f)_{0,T} = \left(1, \mathcal{S}(f)_{0, T}^{1}, \dots, \mathcal{S}(f)_{0, T}^{d}, \dots \mathcal{S}(f)_{0, T}^{i_1,i_2,\dots,i_K}, \dots, \mathcal{S}(f)_{0, T}^{d,d,\dots,d}\right).
\end{equation}
In this case, the dimensionality of $\text{Sig}^K(f)_{0,T}$ is: $(d^{K+1}-d)(d-1)^{-1}$.
From this two main observations can be made:
\begin{enumerate}
    \item The size of the signature is fixed with respect to the length of the path. This property is extremely useful in situations in which paths of different lengths might have to be compared one against the other, as is the case of \gls{ns};
    \item The size of the signature increases exponentially with $K$ and polynomially with $d$. 
    This means that the right balance between the size of the signature and the amount of information preserved when truncating it needs to be found.
\end{enumerate}

\subsection{Signature of a discrete path}
\label{sec:sign_emb}
As discussed, the path signature is defined for continuous paths with bounded variation.
This means that, in case of a sequence of sampled points, the path function defined in Eq. \eqref{eq:sign_path_func} needs to perform interpolation between successive points of the sequence.
This operation is called \emph{embedding} and it can be done in multiple ways \cite{fermanian2021embedding}.
This work will consider two among them:
\begin{itemize}
    \item \textbf{Linear}: consisting of connecting each consecutive points by a straight line. This is one of the most used interpolation in the literature thanks to its simplicity \cite{graham2013sparse, fermanian2021embedding, yang2017developing, bonnier2019deep};
    \item \textbf{Time Path}: building over the linear path, it adds another monotone coordinate encoding the time \cite{yang2017developing, fermanian2021embedding}. Having a monotone coordinate allows to have an unique signature for each path.
    This is because the path will contain information also about its speed rather than only its geometry.
    However, this comes at the cost of having an additional coordinate to the path, leading to higher dimensional signatures. 
\end{itemize}

\begin{figure*}[!p]
    \centering
    \includegraphics[height=.9\textheight]{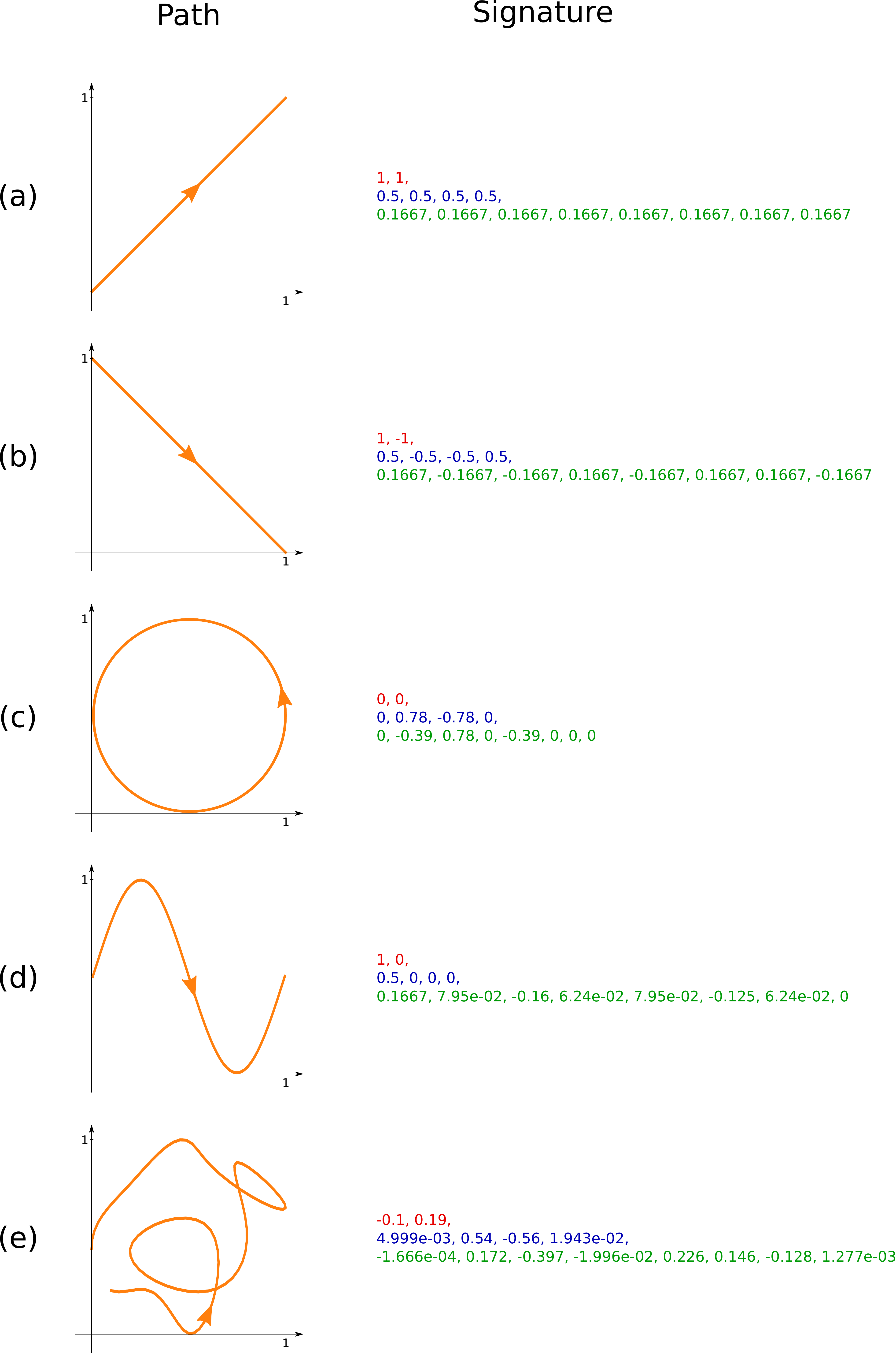}
    \caption[Signature examples]{Signature of degree 3 of different oriented paths. The paths are shown in orange. Each line of the signature represents the elements of a different order: in red the elements of the first order, in blue of the second and in green of the third.
    }
    \label{fig:sign_example}
\end{figure*}

An example of the signature of order 3 for different oriented paths with linear embedding is shown in Fig. \ref{fig:sign_example}.
The distances calculated between the signatures of these different paths are shown in Tab. \ref{tab:sign_dist}.
It is possible to see how, in the signature transform space, the two straight lines Fig. \ref{fig:sign_example}.(a) and \ref{fig:sign_example}.(b) are the farthest ones.
At the same time, they are both very close to the sinusoidal in Fig. \ref{fig:sign_example}.(d), thanks to this having a trend piecewise similar to either two of the straight lines.
The other closest curves in the signature space are the circle (Fig. \ref{fig:sign_example}.(c)) and the more complex path (Fig. \ref{fig:sign_example}.(e)).
This is likely due to the complex path having starting and finishing positions very close one to the other.
From this it is possible to observe that the signature transform, and the distance in the corresponding space, can encode multiple - and not immediately evident - aspects of a path.

\begin{table}[!h]
\centering
\begin{tabular}{|c||c|c|c|c|c|}
\hline
    &     \includegraphics[width=0.12\textwidth]{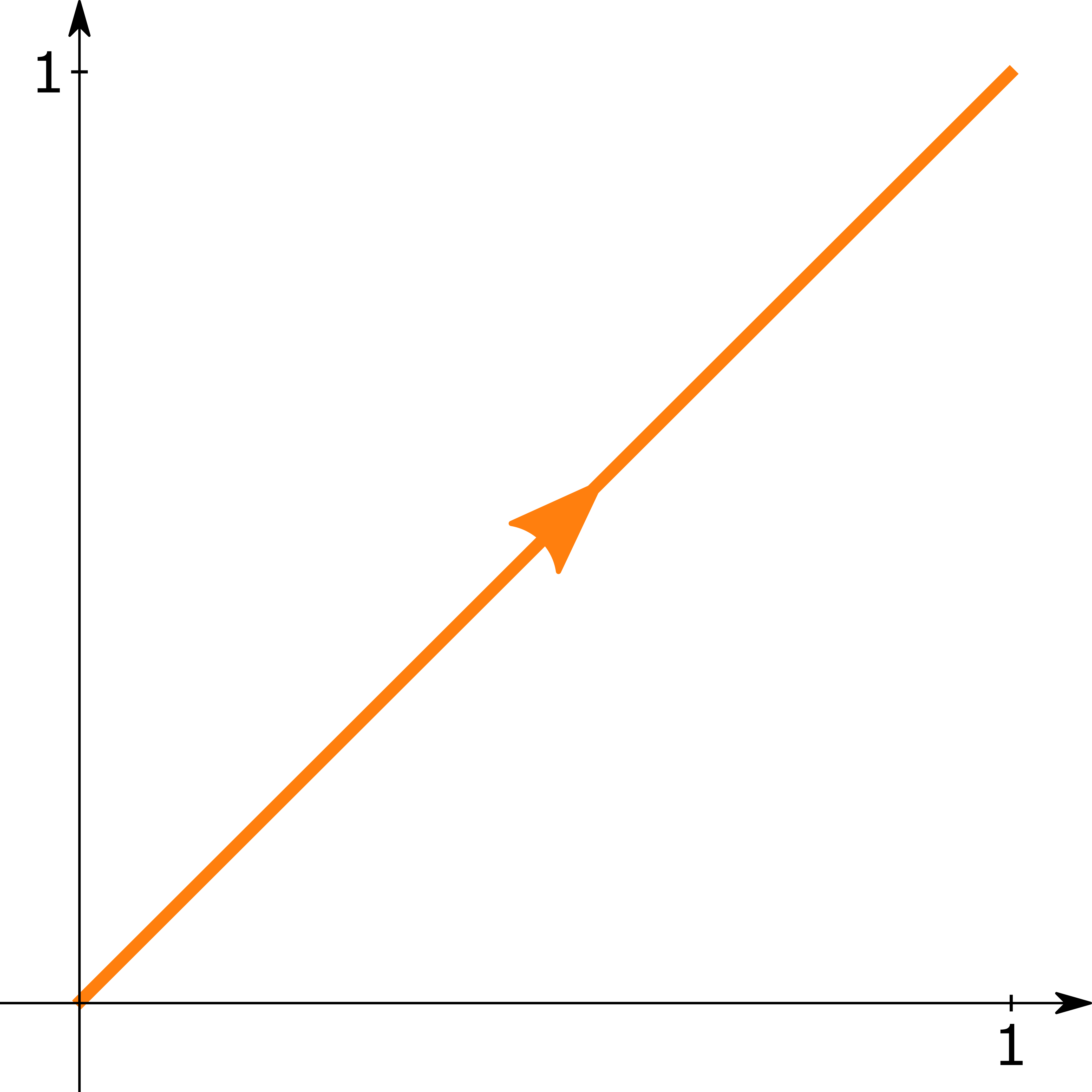}
 & \includegraphics[width=0.12\textwidth]{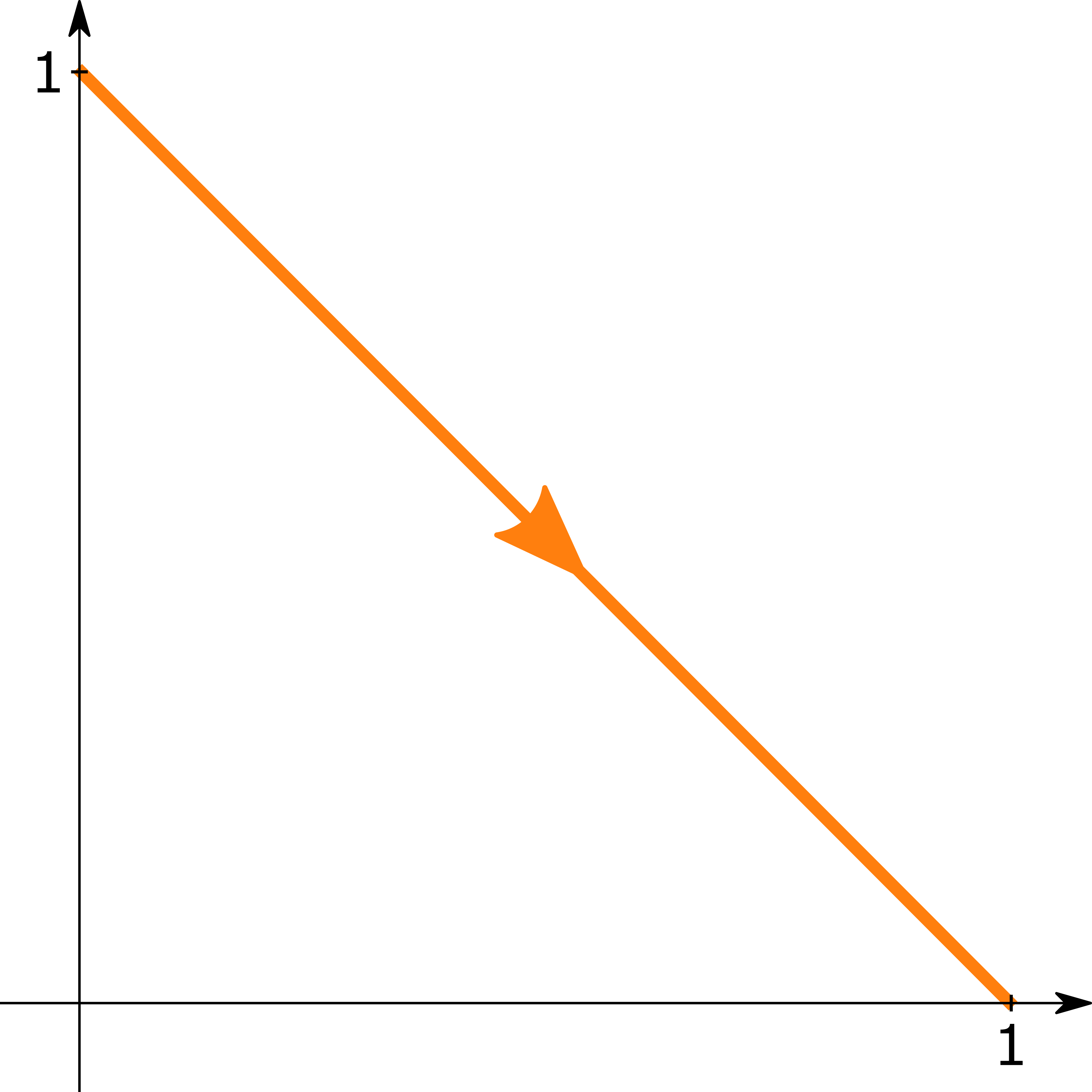}   
 & \includegraphics[width=0.12\textwidth]{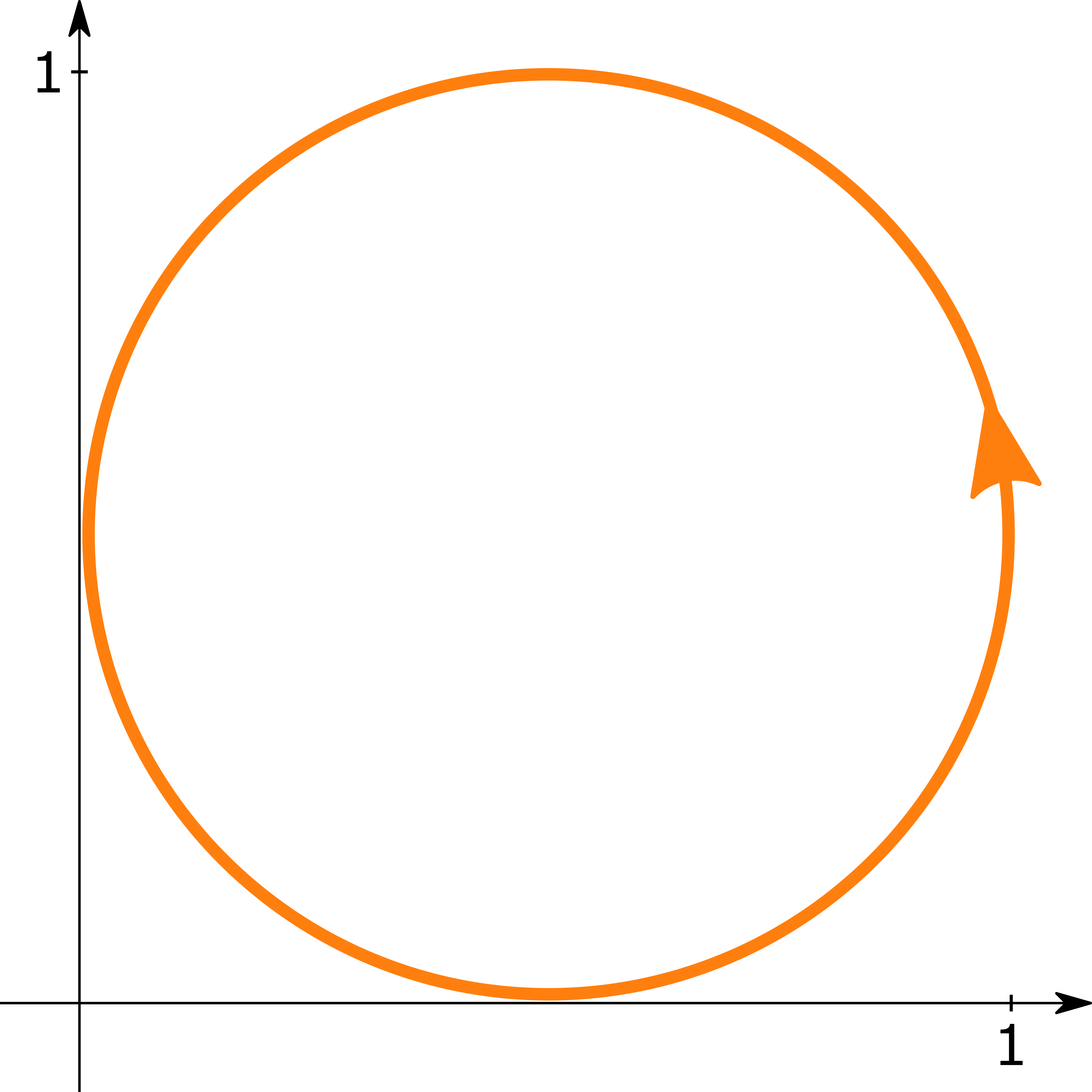}
 & \includegraphics[width=0.12\textwidth]{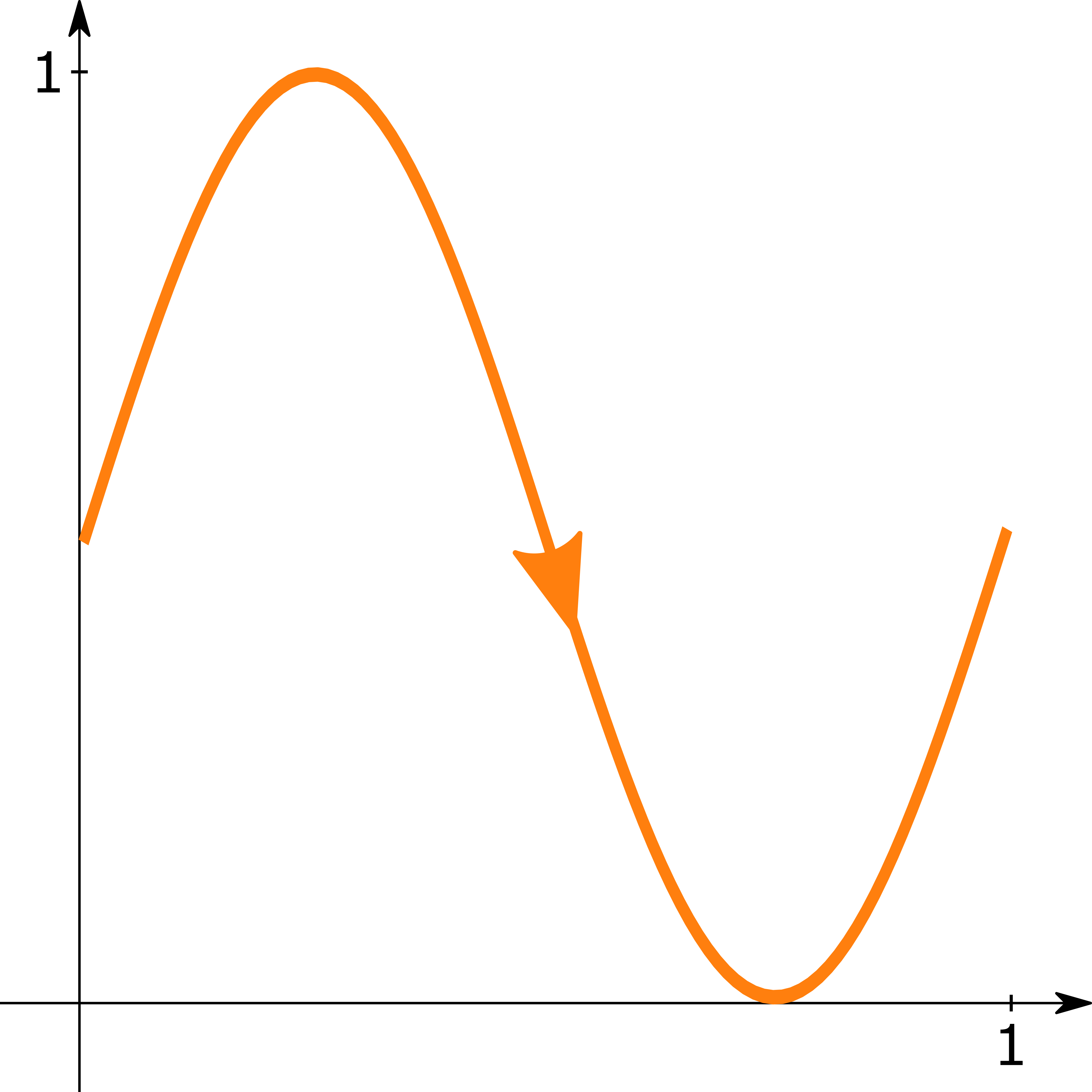}
 & \includegraphics[width=0.12\textwidth]{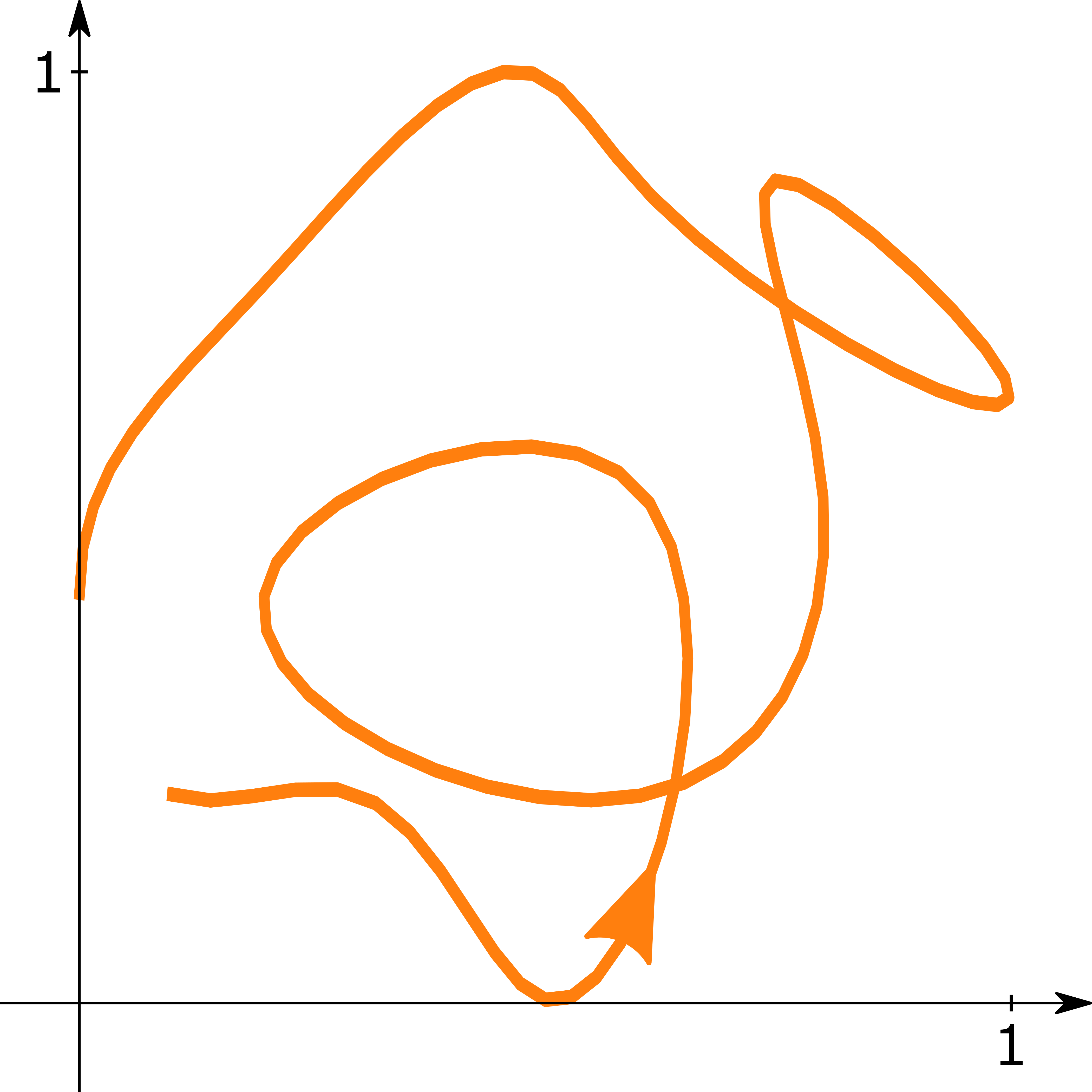}   \\ \hhline{|=||=|=|=|=|=|}
 
\centered{\includegraphics[width=0.12\textwidth]{signatures/figures/path1.png}} & 
\centered{0} & \centered{2.538} & \centered{2.313} & \centered{1.417} & \centered{1.991} \\ \hline

\centered{\includegraphics[width=0.12\textwidth]{signatures/figures/path2.png}} &
& \centered{0} & \centered{2.313} & \centered{1.416} & \centered{2.165} \\ \hline

\centered{\includegraphics[width=0.12\textwidth]{signatures/figures/path3.png}} &
&       & \centered{0}     & \centered{1.96}  & \centered{1.507} \\ \hline

\centered{\includegraphics[width=0.12\textwidth]{signatures/figures/path4.png}} &
&       &       & \centered{0}     & \centered{1.526} \\ \hline
\end{tabular}
\caption{Distances between the signatures of the curves shown in Fig. \ref{fig:sign_example}.}
\label{tab:sign_dist}
\end{table}

\section{Signed Behavior Descriptor}
\label{sec:sign_bd}
Extracting the behaviour descriptor from the trajectory of observations requires the knowledge of the time-step at which the interesting behaviour has happened.
This can be limiting in situations in which the time-step is not always the same or when multiple events can happen at different time-steps.

Using the whole trajectory of observations as behaviour descriptor can help in overcoming this problem.
At the same time, this requires to encode this trajectory in a lower dimensional space to prevent the degeneration of the distance metrics that happens in high-dimensional spaces \cite{Aggarwal2002distance}.
The signature transform provides exactly this while being fast to compute and having strong mathematical guarantees.

This means that the signature transform can be used to extract the behavior descriptor $b_i$ used by \gls{ns} to evaluate the novelty of a policy $\theta_i \in \Theta$.
In section \ref{sec:related_ns}, it was shown how a policy $\theta_i$ traverses a trajectory of states $\tau = [s_0, \dots, s_T]$ when evaluated.
The states are observed through some sensors to generate a corresponding trajectory of observations $\tau_{\mathcal{O}} = [o_0, \dots, o_T]$.
\gls{ns} then uses an observer function $O_B : \mathcal{O}^{T} \rightarrow \mathcal{B}$ to extract the behavior descriptor from the trajectory of observations. 
Usually the observer function only works on some selected observations from the whole trajectory, forcing the designer to know at which time-step the most interesting observations can be found.
Using the truncated signature transform $\text{Sig}^K(\cdot)_{0, T}$ as observer function to encode the whole trajectory into $b_i$ can remove this limitation.
This means that Eq. \eqref{eq:ns_phi} can be rewritten as:
\begin{equation}
    \label{eq:sign_bd}
    \phi(\theta_i) = \text{Sig}^K(\tau^i_{\mathcal{O}})_{0, T} = b_i,
\end{equation}
where $\tau^i_{\mathcal{O}}$ is the trajectory of observations generated by the evaluation of the policy $\theta_i$.

To be able to use the signature in this way, the constraint defined at the beginning of Sec. \ref{sec:sign_method} has to be respected.
This requires the function defined in Eq.\eqref{eq:sign_path_func}, representing the trajectory of observations, to be of bounded variation.
In practice, this constraint is not too restraining: as long as $o_t \in \mathbb{R}^d$, the simple linear embeddings defined in Sec. \ref{sec:sign_emb} will always limit $f(\cdot)$ according to:
\begin{equation}
   -\infty < \min(\tau_{\mathcal{O}}) \leq f(\cdot) \leq \max(\tau_{\mathcal{O}}) < \infty.
\end{equation}
Note that the most external inequalities are always true thanks to the closeness of $\mathbb{R}^d$ with respect to addition and multiplication, meaning that neither $-\infty$ nor $\infty$ are included in the set of the real numbers.

This method will be tested in the next sections, to study if a signature-based behavior descriptor can help the exploration process of \gls{ns} in discovering a more diverse set of solutions with respect to classical descriptors.
Note that, while the final goal is to apply the discussed ideas to \gls{taxons}, the methods in this Chapter will be tested on the ground-truth low-dimensional observations of the environment.
This allows to better focus on the properties of the methods themselves, without any possible interference from the dimensionality reduction component of \gls{taxons}.
\section{Experiments}
\label{sec:sign_exp}
The \emph{signature based behavior descriptor} for \gls{ns} is tested in this section.
The question this study wants to answer is:\\

\begin{center}
\noindent\emph{Is the signature encoding of the whole trajectory of traversed states beneficial to the exploration performed by \gls{ns}?}\\
\end{center}

\begin{figure}
    \centering
    \includegraphics[width=.5\textwidth]{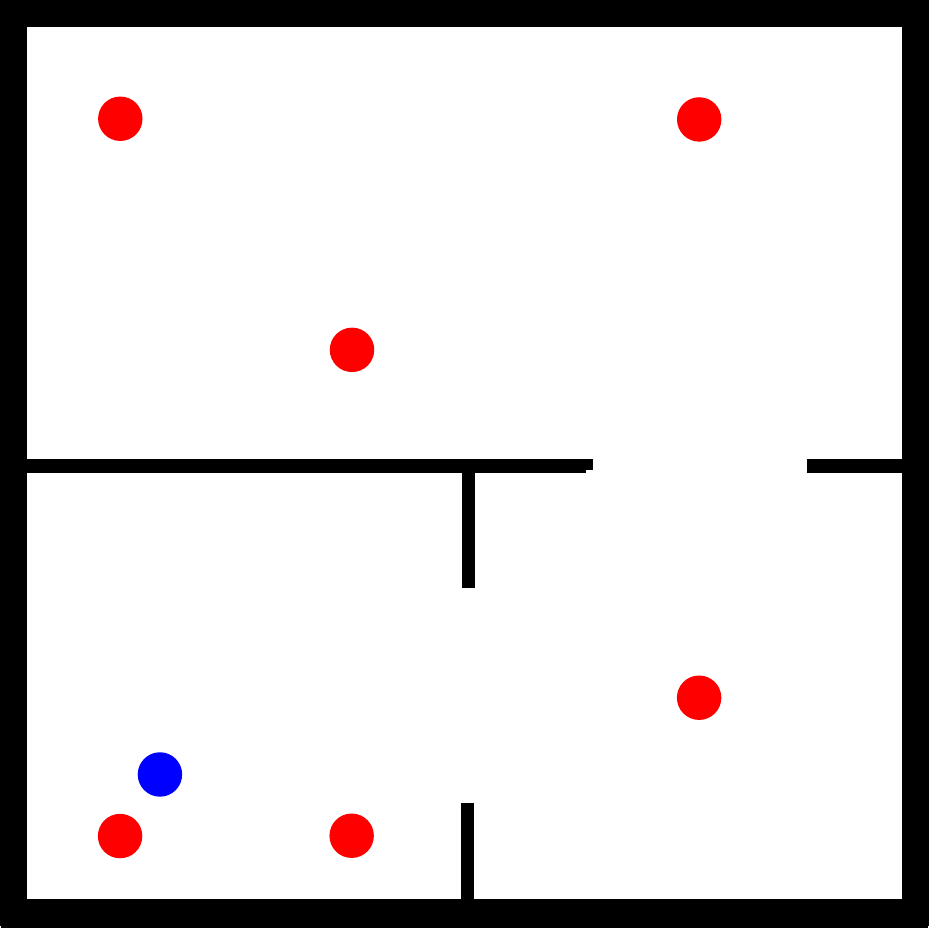}
    \caption[CollectBall environment]{The CollectBall environment. In blue is shown the 2 wheeled robot, while in red are the 6 balls to be collected. The collection point corresponds to the same position of where the robot starts the episodes.}
    \label{fig:sign_collectball}
\end{figure}
In light of this, the algorithm is tested in the CollectBall environment \cite{doncieux2010behavioral}, shown in Fig. \ref{fig:sign_collectball}.
This consists of a maze in which a 2-wheeled robot has the goal of collecting 6 red balls.
The robot, in blue in the figure, is controlled by a 2 layers \gls{nn}, with each layer of size 5.
The controller receives a 10 dimensional state vector as input and outputs a 3 dimensional control vector.
The input state vector consists of the reading of the sensors with which the robot is equipped: 3 distance sensors, 2 bumpers, 2 ball detection sensors, 2 goal detection sensors and a "collected ball" sensor.
At the same time, the control vector returned by the \gls{nn} consists of the speed of the 2 wheels and a "collect ball" signal. 
If the robot is close to a ball and the value of this signal is over 0.5 the ball is collected; if the robot is carrying a ball and the value of this signal is lower than 0.5, the ball is released.
The robot can carry a single ball at a time. 
A ball is considered collected if it is released close to the basket located at the initial position of the robot.
On the contrary, if a ball is released away from the basket, it appears in the maze at the position it was released.
Each policy is run in the environment for 2000 timesteps.
The final reward given to the agent is equal to the number of collected balls at the end of an episode.

The following variants and baselines are compared in the experiments:
\begin{itemize}
    \item \textbf{NS}: vanilla \gls{ns}, whose behavior descriptor is calculated as the final $(x,y)$ position of the robot in the map;
    \item \textbf{NS_multi}: \gls{ns} whose behavior descriptor is calculated by concatenating 5 vectors, sampled at regular intervals during each trajectory, containing the $(x,y)$ position of the robots. The descriptor has size 10;
    \item \textbf{SIGN}: \gls{ns} in which the behavior descriptor is the signature of order $K=5$ of the trajectory of the robot. This baseline uses the \emph{linear} embedding: each datapoint along the trajectory consists of the $(x,y)$ position of the robot at the corresponding time-instant. This leads to a descriptor of size 62;
    \item \textbf{TIME-SIGN}: \gls{ns} in which the behavior descriptor is the signature of order $K=5$ of the trajectory of the robot. This baseline uses the \emph{time path} embedding: each datapoint of the trajectory is composed by concatenating the $(x,y)$ position of the robot and the time coordinate $t\in [0,1]$. This gives a descriptor of size 363.
\end{itemize}
Finally, the statistical results are computed over 15 runs for each experiment.

\section{Results}
\label{sec:sign_results}
This section discusses the results obtained during the experiments.

\subsection{Exploration}
\label{sec:sign_exploration}
\begin{figure}[h]
    \includegraphics[width=\linewidth]{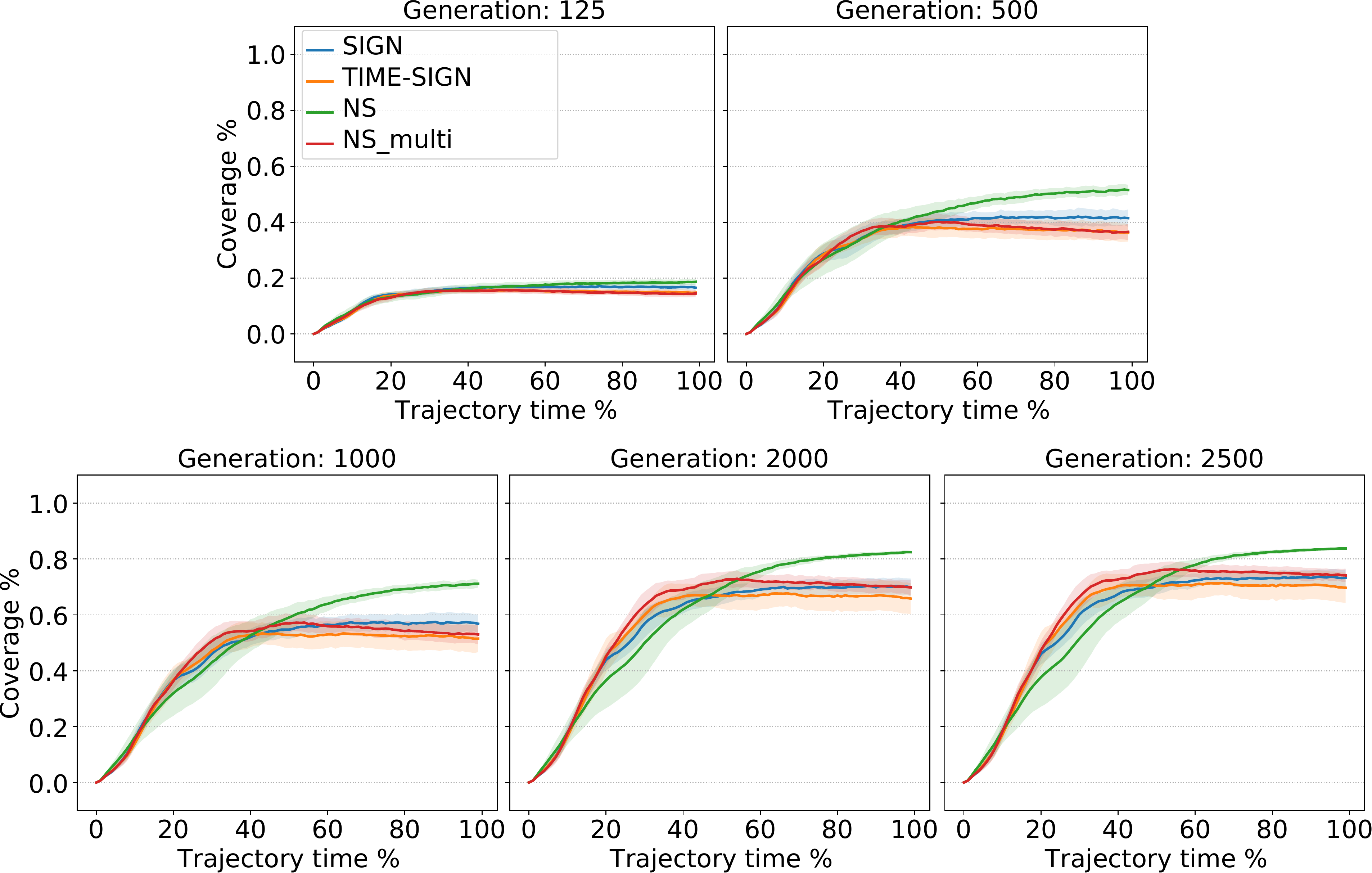}
    \caption[Signature coverage results]{Average coverage with respect to the time of the trajectory for 5 different generations. The shaded areas represent one standard deviation.}
    \label{fig:sign_time_cvg}
\end{figure}

The first thing to verify is the ability of the signature transform to drive exploration for \gls{ns}.
To do so, this section studies how well the population can cover the \gls{bs} at any given moment while performing the trajectory.
This means that the trajectories are discretized in 100 equidistant timesteps and the coverage is calculated at each timestep.
The coverage itself is measured by calculating the percentage of occupied cells of a $50 \times 50$ grid discretizing the $(x,y)$ space in which the robot moves \cite{mouret2015illuminating, paolo2019unsupervised}.
The results for different generations during the experiments are shown in Fig. \ref{fig:sign_time_cvg}.
For the first generations all the methods perform similarly, as can be see for generation 125 in the Figure.
With the passing of the generations a clear trend starts to emerge, with the vanilla \gls{ns} variant reaching the highest coverage at the end of the trajectories ($p = 3.93\times 10{-11}$).
This is not surprising, given that this version optimizes for the higher diversity at the end of the trajectories themselves.
The other methods optimizing for diversity along the whole trajectories reach higher coverages earlier in the trajectory than \gls{ns}.
\gls{ns}\_multi can cover almost 80\% of the whole space just after only 30\% of the total timesteps.
\gls{ns} to reaches the same values only after more than 50\% of the trajectories have been executed.
The signature based variants can also explore in a similar fashion to \gls{ns}\_multi, reaching high coverage values early on in the trajectories.
This is also expected: contrary to vanilla \gls{ns}, these variants are designed to diverge along the whole time dimension.
At the same time, the signature based variants perform worse or similarly than the simpler \gls{ns}\_multi.
While early on in the exploration, at generation 500, SIGN reaches higher coverage than both \gls{ns}\_multi and TIME-SIGN $(p = 1.81 \time 10^{-4}$), by the end of the run, at generation 2500, both SIGN and TIME-SIGN perform worse or similar than \gls{ns}\_multi ($p=0.389$ for SIGN and $p=9.58 \times 10^{-3}$ for TIME-SIGN).
This phenomenon happens along the whole trajectory.

\begin{figure}[h]
    \includegraphics[width=\linewidth]{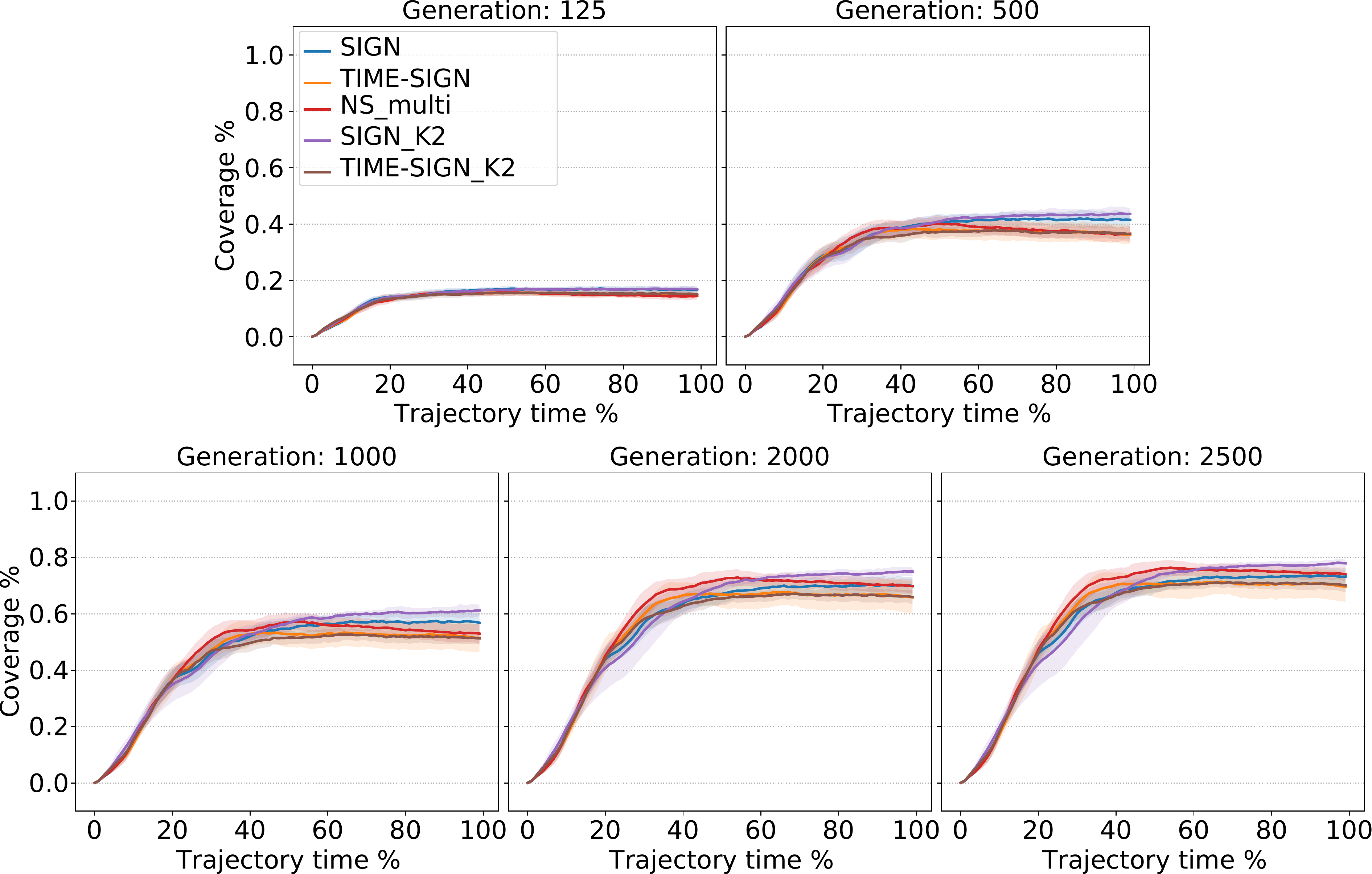}
    \caption[Signature dimensionality experiments - coverage results]{Average coverage with respect to the time of the trajectory for 5 different generations. The shaded areas represent one standard deviation.}
    \label{fig:sign_order_cvg}
\end{figure}

A possible explanation of this effect is the higher dimensionality of the descriptor generated by the signature methods.
Aggarwal et al. showed how the euclidean distance, used to calculate the novelty of the policies' behaviors in Eq. \eqref{eq:ns_novelty}, loses meaning in high-dimensional spaces.
To test this hypothesis experiments with variants of both SIGN and TIME-SIGN with a lower order of the transform: $K=2$ rather than $K=5$ have been performed.
While this lowers the amount of information about the structure of the trajectory included in the descriptor, it also reduces the dimension of the descriptor.
The two variants are called \textbf{SIGN\_K2} and \textbf{TIME-SIGN\_K2}, with a descriptor dimensionality of 6 and 12, respectively.
The coverage results are shown in Fig. \ref{fig:sign_order_cvg}.
It is possible to see that while the TIME-SIGN variants perform in a very similar fashion, the SIGN\_K2 tends to reach higher coverages towards the end of the trajectory ($p = 1.74\times 10^{-05}$). 
This is due to the fact that the signature of this order takes into account only the displacement caused by the trajectory and the area covered, thus not being able to encode all the variations in the path that can happen between different policies.
A possible reason for this is that the loss of information due to a lower order of the signature balances out any improvements obtained by the reduction in dimensionality of the descriptor.
At the same time, the \gls{ns}\_multi still reaches the highest coverage from early on in the trajectory, notwithstanding having similar final coverage with SIGN.
This shows how such a simple method can outperform the more complex signature transform when using it in combination with \gls{ns} to drive exploration.
This can be due to the signature fostering exploration in a space not relevant to the task, thus different than the one in which the coverage is calculated and \gls{ns}\_multi operates.

\subsection{Rewards}
\label{sec:sign_reward}

This section studies the reward obtained by the tested variants in the environment.
Note that none of the methods used in this and in the previous chapter have an explicit way of optimizing the reward.
This means that any obtained reward is only a by-product of the exploration process.
However, an exploration process capable of obtaining higher rewards is extremely useful in sparse reward settings.
This is because the method can easily be used in conjunction with a reward exploitation approach to optimize the rewarding solutions, as discussed in Chapter \ref{chap:serene}.

\begin{figure}[!h]
    \centering
    \includegraphics[width=.65\textwidth]{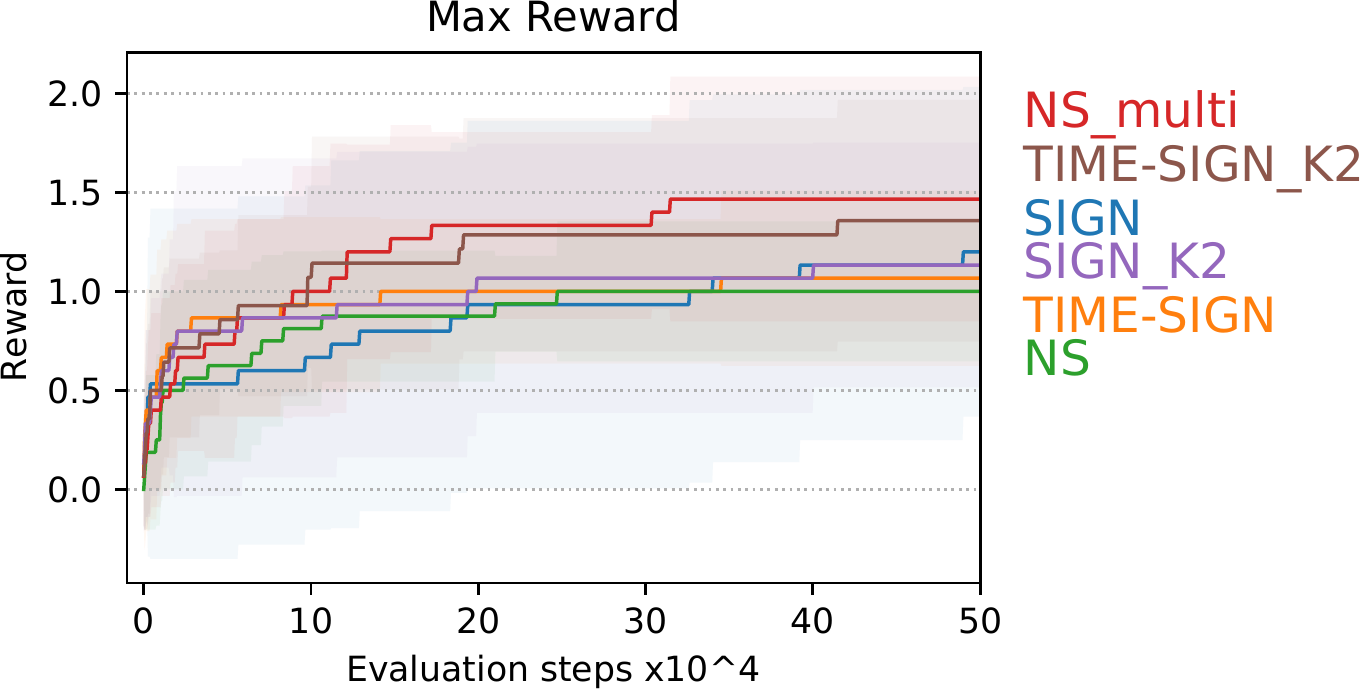}
    \caption[Signature reward results]{Average maximum reward. The shaded areas represent one standard deviation}
    \label{fig:sign_reward}
\end{figure}

In this setting, the reward corresponds to the amount of balls collected by the robot.
The idea behind this is that a diverse enough set of trajectories should be able to collect more balls than policies diversified only with respect to the final position of the robot.
The average over 15 runs of the maximum reward collected with respect to the policies evaluations of each variant is shown in Fig. \ref{fig:sign_reward}.
It is possible to see how \gls{ns} on average tends to collect less than 1 ball, while the variants optimizing for diversity over the whole trajectories reach higher rewards.
The best performing one is \gls{ns}\_multi, proving a more effective method than the signature based ones in performing exploration geared towards sparse rewards systems.
\section{Conclusion}
\label{sec:sign_concl}
This chapter analyzed a way to reduce the amount of prior information needed by the \gls{taxons} algorithm.
Removing the reliance of the method - and \gls{ns} based methods in general - on a specific observation to calculate the behavior descriptor of a policy can greatly extend the generalizability of these algorithms.
In order to do so, the signature transform was tested as a way to encode information about the whole trajectory of traversed states into the behavior descriptors used to calculate the policies novelties.
This novelty would be calculated then not only on a single state, but with respect to the whole trajectory.
The idea has been inspired by the success of the application of the signature transform in other domains of machine learning to encode a stream of data into a single vector \cite{fermanian2021embedding, bonnier2019deep}.
Before applying the transform, the stream of data can be \emph{embedded} in multiple ways.
Two of the most commonly used embeddings were considered: the linear and the time path.
This was compared against two other approaches of calculating the policy behavior descriptor.
The first one uses only the final observation.
The other variant concatenates multiple states sampled at regular intervals along the trajectory.
Notwithstanding its simplicity, this last variant has shown to perform better than all the signatures based variants tested.
A possible reason for this could be the high dimensionality of the extracted features by the signature transform.
This was tested by running experiments with a lower order signature.
The obtained results moved in the direction of confirming our hypothesis for the variant with the linear encoding, but they also showed that the amount of information discarded by the lower order tends to be too high to properly diversify the behaviors along the whole trajectories.
Finally, a test on which of the variants could push for an exploration able to discover as much reward as possible was conducted.
This is a fundamental aspect in sparse rewards settings.
Once again the simpler method of stacking multiple observations proved to be the best performing one.
At the same time, these methods have only been tested on a single setup.
More experiments should be performed on a bigger variety of environments in order to confirm these results.

Nonetheless, in Chapter \ref{chap:stax} the simpler \gls{ns}\_multi strategy was used in order to remove the limitation due to \gls{taxons} working only on the last observation of the trajectory.
Another limiting aspect of the signature approach together with \gls{taxons} is that in order to keep the dimension of the signature vector acceptable, the high-dimensional observations need to be encoded fist by the \gls{ae}.
This would greatly increase the computational cost of the method, having to use the encoder on all the observations of the trajectory rather than just the few sampled ones for the \gls{ns}\_multi approach.

At the same time, this does not disqualify the use of the signature transform with divergent search algorithms.
More research should be done in this regard, also in light of the usefulness that this method has shown in different domains of machine learning research for the embedding of streams of data.

In the last two chapters, methods for reducing the amount of task-specific prior information needed at design time for \gls{ns} have been proposed.
The next chapter will focus on how to take advantage of the rewards discovered during the search.
This will be done by introducing a method augmenting \gls{ns} with the capability of focusing on these rewards to optimize the policies with respect to them.
\clearpage\null\thispagestyle{empty}

\chapter{SERENE}
{\hypersetup{linkcolor=black}\minitoc}
\label{chap:serene}
\vspace{0.4pt}
\par\noindent\rule{\textwidth}{0.4pt}
This chapter is adapted from the following publication:\\

\noindent \textit{Paolo, G., Coninx, A., Doncieux, S., \& Laflaquière, A.} \textbf{Sparse Reward Exploration via Novelty Search and Emitters.} In The Genetic and Evolutionary Computation Conference 2021 (GECCO 2021).
\par\noindent\rule{\textwidth}{0.4pt}

\section{Introduction}
\label{sec:ser_intro}
When \gls{ns} was discussed in chapter \ref{chap:related}, it was highlighted how well the algorithm can explore and cover the search area by ignoring any potential reward and tending towards a uniform exploration of the search space \cite{doncieux2019ns_theory}.
At the same time, its strength is also its limitation: considering all the non-rewarding areas of the search space as valuable as the rewarding ones prevents the algorithm from finding the best possible solutions for solving the task.
Augmenting \gls{ns} with the ability to shift its focus from pure exploration to reward exploitation could help address this issue. 
One possible way of doing so is by using multi-objective optimization methods like NSGA-II \cite{deb2002fast} that can focus both on the diversity and on the reward at the same time.
However, as it will be shown in the following, merging exploration and exploitation through a Pareto front can degrade the exploring power of the algorithm.
A different approach is taken by \gls{qd} algorithms, a family of methods that build a set of both diverse and high-quality solutions \cite{pugh2016qdfontier}.

In this chapter, the \glsreset{serene}\gls{serene} method \cite{paolo2021sparse} is introduced.
This is a \gls{qd} algorithm explicitly designed to address sparse reward problems.
\gls{serene} augments \gls{ns} with \emph{emitters} \cite{fontaine2020covariance} to perform rewards maximization while keeping its exploration ability, thanks to a clear separation between exploration and exploitation.
Introduced as a way to improve the efficiency of \gls{me} \cite{mouret2015illuminating} in the CMA-ME method \cite{fontaine2020covariance}, \emph{emitters} are instances of reward-based algorithms scheduled to perform a local optimization in the search space.
In the original formulation, \gls{me} acts as a scheduler by initializing emitters in different areas of the search space.
The emitters then perform both local exploration and exploitation of the reward, leading to degraded performances in settings with very sparse rewards, where not all policies can obtain a reward.
Conversely, \gls{serene} decouples exploration from exploitation to better deal with these situations.
The former is performed through \gls{ns}, completely ignoring the reward.
Once a reward area is found, \gls{serene} spawns emitters focusing solely on its maximization.
This allows minimal interference between the two processes of exploration and exploitation while letting our algorithm shift its focus between the two processes at any moment.
Persisting in exploring even after some reward areas have been found is essential, since other reward areas could be present in the search space.

In the following, emitters and how they work will be presented in detail in Section \ref{sec:ser_emitters}.
\gls{serene} itself will be introduced in Section \ref{sec:ser_method}, tested in Section \ref{sec:ser_exps}, and the results discussed in Section \ref{sec:ser_discussion}.
The Chapter will conclude with Section \ref{sec:ser_conclusion}, pointing at possible extensions and improvements.
\section{Emitters}
\label{sec:ser_emitters}
An emitter \cite{fontaine2020covariance, cully2020multi} is an instance of an reward optimization algorithm. 
Its objective is to rapidly examine a small area of the search space while optimizing on the reward.
In the work from Fontaine et al. \cite{fontaine2020covariance} and Cully \cite{cully2020multi} the CMA-ME algorithm combines CMA-ES-based emitters  \cite{hansen2016cma} with \gls{me} \cite{mouret2015illuminating}, by using the latter as a scheduler for the emitters evaluation.
It works by initializing a population of policies $\theta$
by sampling their parameters from a distribution $\mathcal{N}(\mu, \Sigma)$ and adding them to the \gls{me} archive.
The algorithm then samples one of these policies and uses it to initialize the population of the emitter $\mathcal{E}_i$.
At this point, $\mathcal{E}_i$ is evaluated until a termination criterion is met; e.g. a lack of increase of the reward found.
Moreover, the policies found during the evaluation of the emitter are added to the \gls{me} archive according to \gls{me} addition strategy.
After the termination of $\mathcal{E}_i$, a new emitter is initialized by sampling another policy from the archive.
This is repeated until the whole evaluation budget is depleted. 

Different types of algorithms can be used as emitters, changing how the search is performed and how the policies are selected.
This shows the flexibility of the approach.
At the same time, existing methods perform exploration through reward-following emitters \cite{fontaine2020covariance, cully2020multi}.
This reduces performances in situations where the reward is very sparse and many of the policies do not get any reward.
Decoupling the exploitation of the reward from the exploration allows to more efficiently deal with sparse rewards settings \cite{colas2018gep}.
\section{Method}
\label{sec:ser_method}
\gls{serene} disentangles the exploration of the behavior space $\mathcal{B}$ from the exploitation of the reward through a two-steps process.
In the first phase, called \emph{exploration phase}, $\mathcal{B}$ is explored by performing \gls{ns}.
As per equation \eqref{eq:ns_phi}, the policies $\theta_i$ found during exploration are assigned a behavior descriptor $\phi(\theta_i)$.
A policy obtaining a reward means that its $\phi(\theta_i)$ belongs to the subspace of rewarding behaviors $\mathcal{B}_{\text{Rew}} \subseteq \mathcal{B}$.
It is in this subspace that the exploitation of the reward happens.
This is done in the second phase, called \emph{exploitation phase}, in which emitters are initialized using the rewarding policies found in $\mathcal{B}_{\text{Rew}}$ during exploration.
During the exploitation phase the most rewarding policies are stored to be returned as result of the algorithm.
Moreover, the particularly novel policies found by the emitters are stored together with the policies found during the exploration process.

By launching emitters only in the neighborhoods of the reward areas, \gls{serene} keeps the exploitation of the reward separated from the exploration of the search space.
This results in taking the best of both worlds: the exploration power of \gls{ns} and the focused exploitation of reward-based algorithms.
\begin{figure}[!ht]
    \centering
    \includegraphics[width=\linewidth]{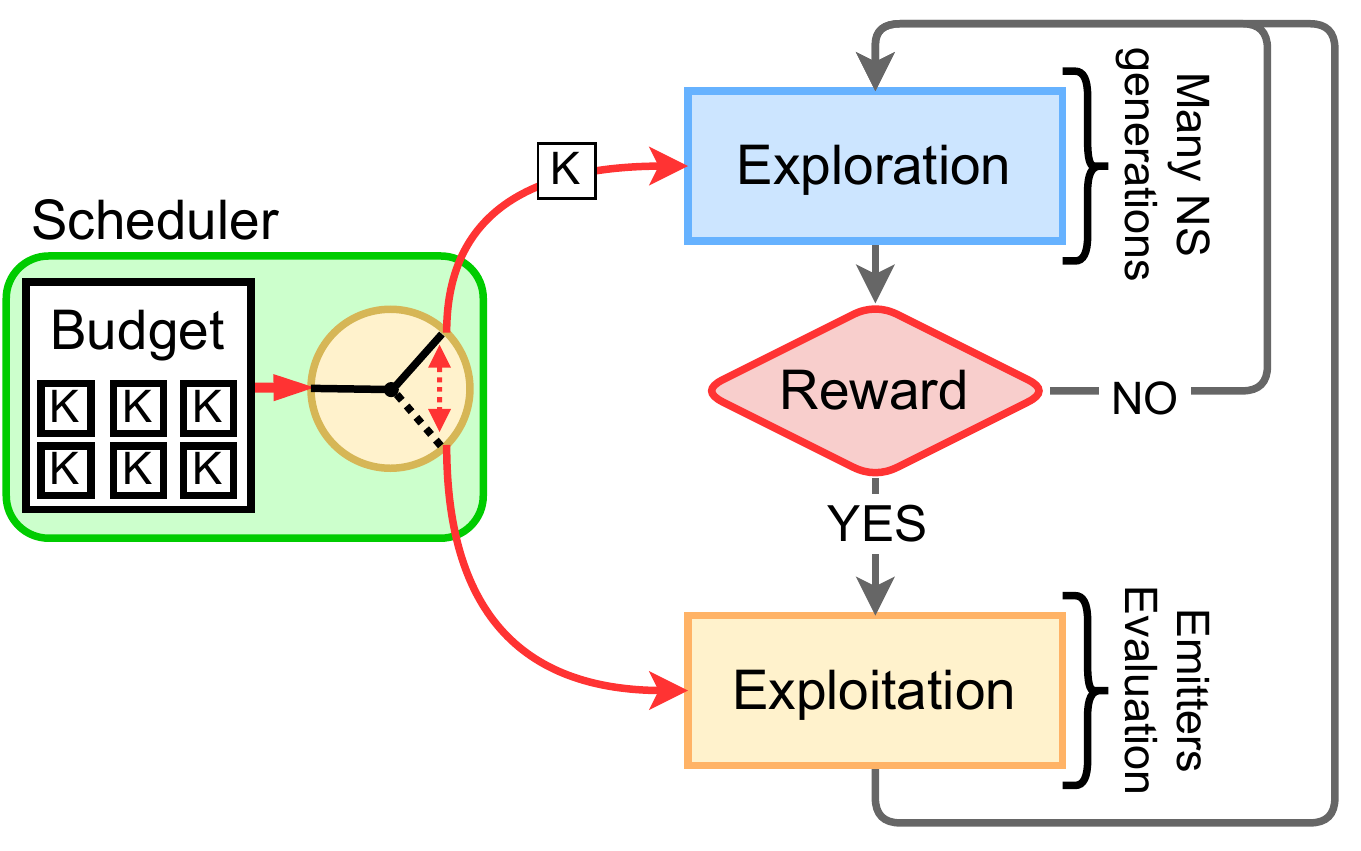}
    \caption[SERENE]{SERENE consists of two exploration and exploitation processes, controlled by a scheduler.
    The exploration process searches for novel solutions through Novelty Search.
    The exploitation process uses emitters to optimize the rewards discovered during exploration.
    The scheduler alternates between the two processes by splitting the total evaluation budget into chunks of size K to assign to either of them.} 
    \label{fig:ser_algo}
\end{figure}

The exploitation and exploration phases are alternated repeatedly through a meta-scheduler.
This scheduler divides a total evaluation budget $Bud$ in smaller chunks of size $K_{Bud}$ 
and assigns them to either one of the two phases.
The whole process is illustrated in Figure \ref{fig:ser_algo} and described in Algorithm \ref{alg:serene}.
\begin{algorithm}[!h]
\caption{\gls{serene}}\label{alg:serene}
\textbf{INPUT:} evaluation budget $Bud$, budget chunk size $K_{Bud}$, population size $M$, emitter population size $M_{\mathcal{E}}$, offspring per policy $m$, mutation parameter $\sigma$, number of policies added to novelty archive $N_Q$\;
\textbf{RESULT:} Novelty archive $\mathcal{A}_{\text{Nov}}$, rewarding archive $\mathcal{A}_{\text{Rew}}$\;
$\mathcal{A}_{\text{Nov}} = \emptyset$;
$\mathcal{A}_{\text{Rew}} = \emptyset$\;
$\mathcal{Q}_{\text{Em}} = \emptyset$;
$\mathcal{Q}_{\text{Cand\_Nov}} = \emptyset$;
$\mathcal{Q}_{\text{Cand\_Em}} = \emptyset$\;
$\Gamma_0 \leftarrow M$ policies from $\Theta$\;
Split $Bud$ in chunks of size $K_{Bud}$\;
\While{$Bud$ not depleted}{
    \If{$\Gamma_0$}{
        Evaluate $\theta_i, ~~ \forall \theta_i \in \Gamma_0$\;
    Calculate $b_i = \phi(\theta_i) \in \mathcal{B}, ~~ \forall\theta_i \in \Gamma_0$\;
    }
    \emph{ExplorationPhase} ($K_{Bud}$, $m$, $\sigma$, $\mathcal{A}_{\text{Nov}}$, $\mathcal{Q}_{\text{Cand\_Em}}$, $\Gamma_g$,~$N_Q$)\;
    
    \If{\textbf{not} $\mathcal{Q}_{\text{Cand\_Em}} == \emptyset$ \textbf{or} \textbf{not} $\mathcal{Q}_{\text{Em}} == \emptyset$}{
    \emph{ExploitationPhase} ($K_{Bud}$, $\mathcal{Q}_{\text{Cand\_Em}}$, $\lambda$, $m$, $\mathcal{Q}_{\text{Em}}$, $\mathcal{A}_{\text{Nov}}$, $\mathcal{A}_{\text{Rew}}$, $M_{\mathcal{E}}$)\;
    }
}
\end{algorithm}
\\

\begin{figure*}
    \centering
    \includegraphics[width=\linewidth]{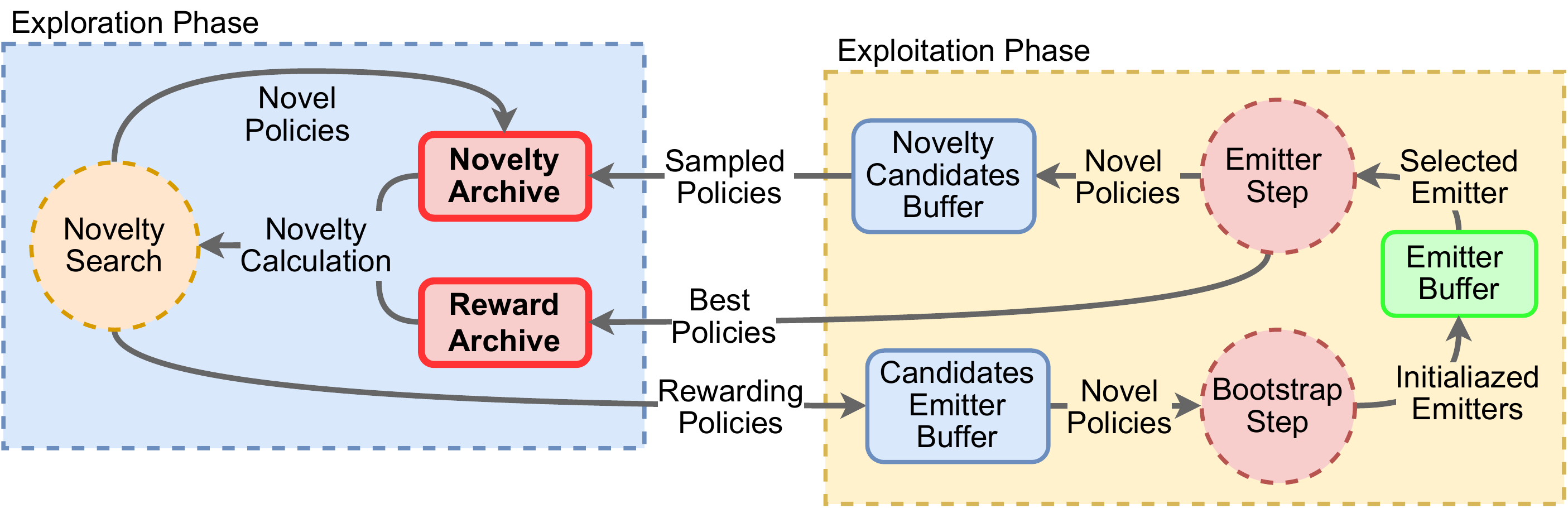}
    \caption[SERENE sets]{Overview of the sets used by \gls{serene} to keep track of the explored areas and the initialized emitters. Highlighted in red are the two archives returned as final result of the algorithm execution.
    }
    \label{fig:ser_sets}
\end{figure*}

To keep track of policies generated during the different phases, \gls{serene} uses the following buffers and containers:
\begin{itemize}
    \item \emph{novelty archive} $\mathcal{A}_{\text{Nov}}$: a repertoire of the novel policies found during the \emph{exploration phase}, and returned as first output of \gls{serene};
    \item \emph{reward archive} $\mathcal{A}_{\text{Rew}}$: a repertoire of rewarding policies found during the \emph{exploitation phase}, returned as second output of \gls{serene};
    \item \emph{candidates emitter buffer} $\mathcal{Q}_{\text{Cand\_Em}}$: a buffer containing the rewarding policies $\phi(\theta_i) \in \mathcal{B}_{\text{Rew}}$ found during the \emph{exploration phase} and used in the \emph{exploitation phase} to initialize emitters;
    \item \emph{emitter buffer} $\mathcal{Q}_{\text{Em}}$: a buffer containing all the initialized emitters to be evaluated during the \emph{exploitation phase};
    \item \emph{novelty candidates buffer} $\mathcal{Q}_{\text{Cand\_Nov}}$: a buffer containing the most novel policies found by the emitter. Each emitter has its own instance of this buffer and the policies in it are sampled for addition to the novelty archive $\mathcal{A}_{\text{Nov}}$ once the emitter is terminated. 
\end{itemize}
A high-level overview of how these sets interact during the two phases is given in Figure \ref{fig:ser_sets}, and a more detailed description is proposed in the two following subsections.

\noindent
\begin{minipage}{\textwidth}
\gls{serene} uses multiple populations during each process of the search.
They are listed here for clarity:
\begin{itemize}
    \item $\Gamma_g$: population of size $M$ at generation $g$. Used by \gls{ns} during the exploration phase;
    \item $\Gamma^m_g$: offspring population of size $m \times M$ at generation $g$. Used by \gls{ns} during the exploration phase, it is generated by spawning $m$ agents from each policy $\theta_i \in \Gamma_g$;
    \item $P_{\gamma}$: emitter population of size $M_{\mathcal{E}}$ at the emitter generation $\gamma$. Used by the emitter during the exploitation of the reward;
    \item $P^m_{\gamma}$: emitter offspring population of size $m \times M_{\mathcal{E}}$  at the emitter generation $\gamma$. 
    Used by the emitter during the exploitation of the reward, it is generated by spawning $m$ agents from each policy $\Tilde{\theta}_i \in P_{\gamma}$.
\end{itemize}
\end{minipage}

\subsection{Exploration phase}
\gls{serene} starts by generating an initial population $\Gamma_0$ of size $M$.
This is done by sampling the parameters of the population's policies $\theta_{j}$ from a normal distribution $\mathcal{N}(0, I)$.
The population is used to explore the behavior space $\mathcal{B}$ through \gls{ns}.
At each generation $g$, a mutation operator generates $m$ new policies $\theta_j^i$ (offspring) from each of the policies $\theta_j \in \Gamma_g$:
\begin{equation}
\label{eq:ser_mutation}
    \forall j, i \in \{1, \dots, M\} \times \{1, \dots, m\}, \theta^i_j = \theta_j + \epsilon, ~~~ \text{with} ~~~ \epsilon \sim \mathcal{N}(0, \sigma I).
\end{equation}
The resulting offspring population $\Gamma^m_g$, of size $m \times M$, is then evaluated to obtain the behavior descriptors $\phi(\theta_j^i) = b_j^i \in \mathcal{B}$, used to calculate the novelty of $\Gamma_g$ and $\Gamma^m_g$ according to Eq. \eqref{eq:ser_novelty}:
\begin{equation}
\label{eq:ser_novelty}
\eta(\theta_i) = \frac{1}{|J|}\sum_{j \in J}\text{dist}(b_i, b_j)  = \frac{1}{|J|}\sum_{j \in J}\text{dist}\big(\phi(\theta_i), \phi(\theta_j)\big),
\end{equation}
The novelty is then used to generate the next generation population $\Gamma_{g+1}$ by taking the most novel policies from the current population and the offsprings.
At the same time, $N_Q$ policies among the offsprings are uniformly sampled to be added to the \emph{novelty archive} $\mathcal{A}_{\text{Nov}}$. 
Finally, all the policies for which $r_i > 0$ are stored in the \emph{candidates emitters buffer} $\mathcal{Q}_{\text{Cand\_Em}}$.
The process just described is detailed in Algorithm~\ref{alg:ser_exploration}.

The exploration phase is executed for the $K_{Bud}$ evaluation steps in the given budget chunk, where each evaluation step corresponds to one policy evaluation.
Once the chunk is depleted, the scheduler assigns the next chunk to the \emph{exploitation phase} only if $\mathcal{Q}_{\text{Cand\_Em}} \neq \emptyset$.
On the contrary, if the buffer is empty, i.e. no new reward has been found during exploration, another \emph{exploration phase} is performed. 
This means that in the worst case scenario where no reward can be discovered, i.e. $\mathcal{B}_{\text{Rew}} = \emptyset$, \gls{serene} performs exactly like \gls{ns}.
\begin{algorithm}
\caption{\gls{serene} Exploration Phase}\label{alg:ser_exploration}
\textbf{INPUT:} budget chunk $K_{Bud}$, number of offspring per parent $m$, mutation parameter $\sigma$, novelty archive $\mathcal{A}_{\text{Nov}}$, candidate emitters buffer $\mathcal{Q}_{\text{Cand\_Em}}$, population $\Gamma_g$, number of policies $N_Q$\;
\While{$K_{Bud}$ not depleted}{
    Generate offspring $\Gamma^m_g$ from population $\Gamma_g$\;
    Evaluate $\theta_i, ~~ \forall \theta_i \in \Gamma^m_g$\;
    Calculate $b_i = \phi(\theta_i) \in \mathcal{B}, ~~ \forall\theta_i \in \Gamma^m_g$\;
    Calculate $\eta(\theta_i) = \frac{1}{|J|}\sum_{j \in J}\text{dist}(b_i, b_j), ~~ \forall\theta_i \in \Gamma^m_g \bigcup \Gamma_g$\;
    $\mathcal{A}_{\text{Nov}} \leftarrow N_Q \text{ samples from } \Gamma^m_g$\; 
    \If{$\phi(\theta_i) \in \mathcal{B}_{\text{Rew}}$}{
        $\mathcal{Q}_{\text{Cand\_Em}} \leftarrow \theta_i$
    }
    Generate $\Gamma_{g+1}$ from most novel $\theta_i \in \Gamma^m_g \bigcup \Gamma_g$\;
}
\end{algorithm}
\subsection{Exploitation phase}
The \emph{exploitation phase} consists of two sub-steps: the \emph{bootstrapping step}, in which the policies in the candidates emitter buffer $\mathcal{Q}_{\text{Cand\_Em}}$ are used to initialize and bootstrap emitters, 
and the \emph{emitter step}, in which the initialized emitters are evaluated.

\subsubsection*{Bootstrap step}
During this step, emitters are initialized from the rewarding policies $\theta_i$ in the candidates emitter buffer $\mathcal{Q}_{\text{Cand\_Em}}$, and their potential for reward improvement evaluated.
This insures that only emitters capable of improving the rewards are considered for full evaluation, reducing wasted evaluation budget.
The policies used to initialize the emitters are selected according to their novelty with respect to the reward archive $\mathcal{A}_{\text{Rew}}$.
This enables \gls{serene} to focus on less explored areas of the rewarding behavior space $\mathcal{B}_{\text{Rew}}$.
The whole bootstrapping phase lasts $\nicefrac{K_{Bud}}{3}$ evaluations.

As discussed in Section \ref{sec:ser_emitters}, an emitter is an instance of a \emph{reward-based algorithm}. 
Contrary to the previous work of Fontaine et al. \cite{fontaine2020covariance} and Cully \cite{cully2020multi}, this work does not use estimation-of-distribution algorithms like CMA-ES \cite{hansen2016cma} but rather an elitist reward-based \gls{ea}.
The reason behind this choice is that, as the name suggests, estimation-of-distribution algorithms work by estimating a probability distribution, usually a gaussian, from where the policies for the next generation are selected.
Having to estimate a distribution requires the estimation of a covariance matrix $\Sigma$ whose reliability strongly depends on the ratio between the dimension of the parameter space $\Theta$ and the size of the population.
If the size of population used to estimate the matrix is smaller than the dimension of the parameter space, the estimation of $\Sigma$ is not reliable.
CMA-ES circumvents the issue by using information from previous generations to calculate $\Sigma$ through the evolution path.
While stabilizing $\Sigma$, having to wait for multiple generations leads to a less efficient use of the evaluation budget.
Hence, the emitters used in this work are based on an \emph{elitist evolutionary algorithm} not requiring the estimation of any distribution.
Conversely, the population is composed with the most rewarding policies from the previous generation's population and offspring, while the offspring are generated by mutating the parents according to equation \ref{eq:ser_mutation}.

An emitter $\mathcal{E}_i$ based on this algorithm consists of:
\begin{itemize}
    \item a population $P$ containing $M_{\mathcal{E}}$ policies $\Tilde\theta \in \Theta$;
    \item a population of offspring $P^m$ of size $m \times M_{\mathcal{E}}$;
    \item a generation counter $\gamma$;
    \item a tracker for the maximum reward found so far $R_{\gamma}$;
    \item an improvement measure $I(\cdot)$;
    \item a novelty measure $\eta_i$ equal to the novelty of the policy used to initialize the emitter;
    \item a \emph{novelty candidate buffer} $\mathcal{Q}_{\text{Cand\_Nov}}$.
\end{itemize}
The emitter $\mathcal{E}_i$ is initialized from a policy $\theta_i$ in the candidates emitter buffer by sampling its initial population $P_0$ from the distribution $\mathcal{N}(\theta_i, \sigma_iI)$.
To keep the emitter's exploration local and prevent overlapping with the search space of possible nearby emitters, $\sigma_i$ is initialized as:
\begin{equation}
    \sigma_i = \frac{\min_{j}\big(\text{dist}(\theta_i, \theta_j)\big)}{3}, ~~\forall \theta_j \in \Gamma^m_g \cup \Gamma_g.
    \label{eq:sigma}
\end{equation}
This shapes $\mathcal{N}(\theta_i, \sigma_iI)$ such that all other $\theta_j$ are at least 3 standard deviation away from its center.
Once $\mathcal{E}_i$ has been initialized, its potential is evaluated by running it for $\lambda$ generations and calculating its \emph{emitter improvement} $I(\mathcal{E}_i)$.
This improvement is defined as the difference between the average rewards obtained during the most recent and the initial generations of the emitter:
\begin{equation}
\label{eq:ser_improv}
    I(\mathcal{E}_i) = \frac{1}{\lambda M_{\mathcal{E}}} \left( \sum_{\gamma=T-\nicefrac{\lambda}{2}}^{T} \sum_{j=0}^{M_{\mathcal{E}}}r_{(\gamma,j)} - \sum_{\gamma=\gamma_0}^{\nicefrac{\lambda}{2}} \sum_{j=0}^{M_{\mathcal{E}}}r_{(\gamma,j)} \right).
\end{equation}
Here $T$ is the last evaluated generation, $r_{(\gamma,j)}$ is the reward of policy $\Tilde{\theta}_j \in P_\gamma$, and $\gamma_0$ is the generation at which the emitter is at the beginning of the exploitation phase;
it is always $\gamma_0 = 0$ for an emitter in the bootstrap step.
If $I(\mathcal{E}_i) \leq 0$, the chances for the emitter to find better solutions than the initial ones are low, so it is not worth allotting more budget to its evaluation.
On the contrary, $I(\mathcal{E}_i) > 0$ means that the emitter has high potential for improvement.
Thus all the initialized emitters for which $I(\mathcal{E}_i) > 0$ are added to the \emph{emitter buffer} $\mathcal{Q}_{\text{Em}}$ for further evaluation.

\subsubsection*{Emitter step} 
The initialized emitters in the \emph{emitter buffer} $\mathcal{Q}_{\text{Em}}$ are run during this step.
It starts by calculating the pareto front between the improvement $I(\mathcal{E}_i)$ and the novelty $\eta(\mathcal{E}_i)$ of each of the emitters $\mathcal{E}_i$ in the emitter buffer.
The emitter to run is then randomly sampled from the front of the \emph{non-dominated} emitters.
Using both the novelty and the fitness to select which emitter to run allows \gls{serene} to focus both on the less explored and most promising areas of $\mathcal{B}_{\text{Rew}}$.

The policies $\Tilde{\theta}_j$ generated by an emitter can be stored either for the reward they achieve or for their novelty.
At every generation $\gamma$ all the policies $\Tilde{\theta}_j$ in the current population with a reward $r(\Tilde{\theta}_j) > R_{\gamma-1}$ are added to the reward archive $\mathcal{A}_{\text{Rew}}$.
Additionally, the policies $\Tilde{\theta}_j$ with a novelty higher than the emitter novelty $\eta_i$ are stored into the emitter's \emph{novelty candidates buffer} $\mathcal{Q}_{\text{Cand\_Nov}}$.

\begin{algorithm}
\caption{\gls{serene} Exploitation Phase}\label{alg:ser_emitter}
\textbf{INPUT:} budget chunk $K_{Bud}$, candidate emitters buffer $\mathcal{Q}_{\text{Cand\_Em}}$, number of bootstrap generations $\lambda$, emitter population size $M_{\mathcal{E}}$, number of offspring per policy $m$, emitters buffer $\mathcal{Q}_{\text{Em}}$, rewarding archive $\mathcal{A}_{\text{Rew}}$, novelty archive $\mathcal{A}_{\text{Nov}}$\;
\CommentSty{/* Bootstrap step */}\\
\While{$\nicefrac{K_{Bud}}{3}$ not depleted}{
Select most novel policy $\theta_i$ from $\mathcal{Q}_{\text{Cand\_Em}}$\;
Calculate $\sigma_i$\;
Initialize: $\mathcal{E}_i$, $\mathcal{Q}^i_{\text{Cand\_Nov}} = \emptyset$, and $P_0$\;
\For {$\gamma \in \{0, \dots, \lambda\}$}{
    \If{$P_0$}{
        Evaluate $\Tilde{\theta}_j$, $\forall \Tilde{\theta}_j \in P_0$\;
    }
    Generate offspring population $P^m_{\gamma}$ from $P_{\gamma}$\;
    Evaluate $\Tilde{\theta}_j$, $\forall \Tilde{\theta}_j \in P^m_{\gamma}$\;
    Generate $P_{\gamma+1}$ from best $\Tilde{\theta}_j \in P^m_{\gamma} \bigcup P_{\gamma}$\;
}
Calculate $I(\mathcal{E}_i)$\;
\If{$I(\mathcal{E}_i) > 0$}{
    $\mathcal{Q}_{\text{Em}} \leftarrow \mathcal{E}_i$\;
}
}
\CommentSty{/* Emitters step */}\\
Calculate pareto fronts in $\mathcal{Q}_{\text{Em}}$\;
\While{$\nicefrac{2}{3}K_{Bud}$ not depleted}{
Sample $\mathcal{E}_i$ from \emph{non-dominated emitters} in $\mathcal{Q}_{\text{Em}}$\;

\While{\textbf{not} $terminate(\mathcal{E}_i)$}{
Generate offspring population $P^m_{\gamma}$ from $P_{\gamma}$\;

Evaluate $\Tilde{\theta}_j$, $\forall \Tilde{\theta}_j \in P^m_{\gamma}$\;

$\mathcal{A}_{\text{Rew}} \leftarrow \Tilde{\theta}_j, ~~ \forall \Tilde{\theta}_j \in P^m_{\gamma} \mid r(\Tilde{\theta}_j) > R_{\gamma}$\; 
$\mathcal{Q}^i_{\text{Cand\_Nov}} \leftarrow \Tilde{\theta}_j, ~~ \forall \Tilde{\theta}_j \in P^m_g \mid \eta(\Tilde{\theta}_j) > \eta_i$\;

Generate $P_{\gamma+1}$ from best $\Tilde{\theta}_j \in P^m_{\gamma} \bigcup P_{\gamma}$\;
Update $I(\mathcal{E}_i)$ and $R_{\gamma}$\;

\If{$terminate(\mathcal{E}_i)$}{
$\mathcal{A}_{\text{Nov}} \leftarrow N_Q \text{ samples from } \mathcal{Q}^i_{\text{Cand\_Nov}}$\;
Discard emitter $\mathcal{E}_i$\;
}
}
}
\end{algorithm}

The emitter $\mathcal{E}_i$ is run until either the given budget chunk is depleted or a termination condition is met.
In the first case, \gls{serene} recalculates the improvement $I(\mathcal{E}_i)$ from the beginning of the \emph{emitter phase} and assigns the next budget chunk to the \emph{exploration phase}.
On the contrary, if a termination condition is met, $\mathcal{E}_i$ is discarded and another emitter to evaluate is sampled from the Pareto front.
There can be multiple termination conditions depending on the algorithm used as emitter.
A CMA-ES based emitter can use all the termination criteria listed by Hansen in Appendix B.3 in its tutorial on CMA-ES \cite{hansen2016cma}.
The one used in this work is also inspired from Hansen's work \cite{hansen2016cma}, namely the \emph{stagnation criterion}, which stops the emitter when there is no more improvement on the reward.
To do so the history of the rewards is tracked over the last $120 + 20 * n / M_{\mathcal{E}}$ emitter's generations, where $n$ is the size of the parameter space $\Theta$ and $M_{\mathcal{E}}$ is the emitter's population size.
The emitter is terminated if either the maximum or the median of the last 20 rewards is not better than the maximum or the median of the first 20 rewards.
Before starting the new emitter evaluation, $N_Q$ policies from the terminated emitter's \emph{novelty candidates buffer} $\mathcal{Q}_{\text{Cand\_Nov}}$ are uniformly sampled to be added to the novelty archive $\mathcal{A}_{\text{Nov}}$.
In addition to saving particularly novel solutions as part of the final result, this prevents the exploration phase from re-exploring areas covered by emitters during the exploitation phase.

The whole \emph{exploitation phase} is detailed in Algorithm \ref{alg:ser_emitter}.
\section{Experiments}
\label{sec:ser_exps}
This section testes if \gls{serene} can efficiently deal with sparse reward settings, finding all disjoint reward areas, and optimizing the reward in each of them.
For the evaluation, the following four sparse rewards environments - illustrated in Fig. \ref{fig:ser_environments} - are considered:
\begin{figure}[h]
    \includegraphics[width=\linewidth]{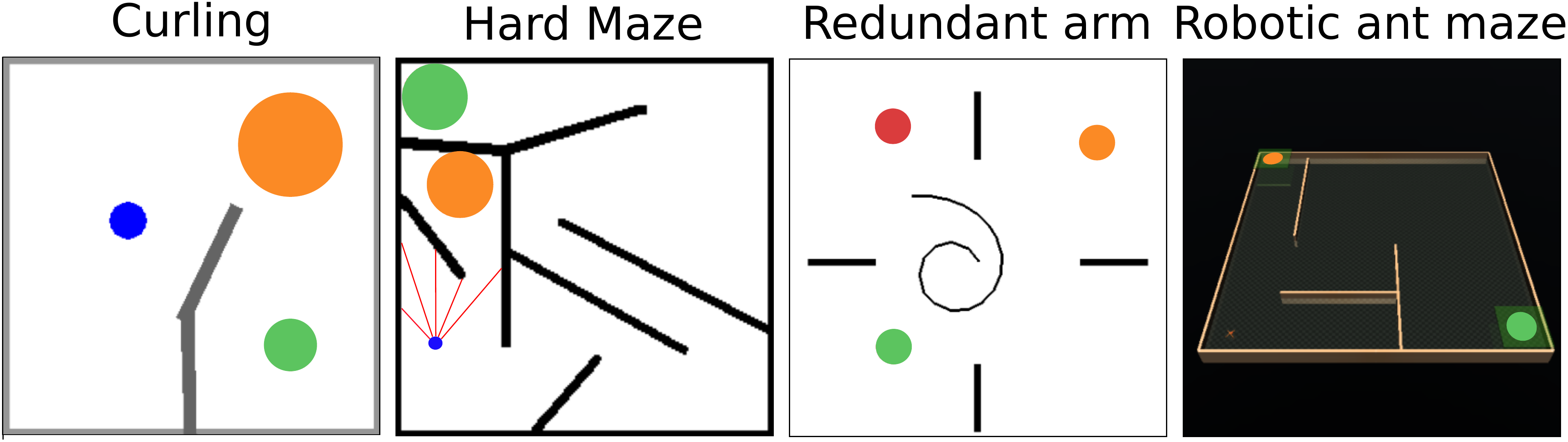}
    \caption[SERENE test environments]{Testing environments: Curling, HardMaze, Redundant arm, Robotic ant maze. 
    }
    \label{fig:ser_environments}
\end{figure}

\noindent\subsubsection*{Curling}
A two \gls{dof} robotic arm \cite{paolo2020billiard} controlled by a 3 layers \gls{nn} with each layer  of size $5$. 
The arm has to push the blue ball into one of the two goal areas shown in orange and green. 
A reward is provided only if the ball stops in one of the two areas. 
Moreover, the closer the ball is to the center of the reward area, the higher the reward is. 
The controller takes as input a 6-dimensional vector containing the ball pose $(x,y)$, and the two joints angles and velocities. 
The output of the controller is the speed of each joint at the next timestep. 
The size of the parameter space $\Theta$ is 94, and each policy is run in the environment for $500$ timesteps.

\noindent\subsubsection*{Hardmaze} 
Inspired by the Hardmaze environment introduced by Lehman and Stanley \cite{Lehman2008NS}, it consists of a two-wheeled robot, in blue, whose task is to navigate the maze and reach either one of the green and orange areas. 
The reward is only given if the robot stops in one of the two areas, and is higher the closer the robot is to its center. 
The robot is controlled by a 2-layers \gls{nn} with each layer of size $5$.
The controller takes as input the reading of the 5 distance sensors mounted on the robot; shown in red in Figure \ref{fig:ser_environments}.
Its output is the 2-dimensional vector containing the speed of the 2 wheels at the next timestep. 
The size of the parameter space $\Theta$ is 63, and each policy is run in the environment for $2000$ timesteps.

\noindent\subsubsection*{Redundant arm}
A 20-\gls{dof} robotic arm in which the arm's end-effector has to reach one of the 3 colored goal areas. 
The reward is maximal in the center of the areas, and the arm is controlled by a \gls{nn} with 2 layers of size 5. 
This environment is based on the one introduced by Loviken and Hemion \cite{loviken2017online}.
The controller takes as input the 20-dimensional vector of each joint's position, and outputs the 20-dimensional joint's torque vector. 
The size of the parameter space $\Theta$ is 228, and each policy is run in the environment for $100$ timesteps.

\noindent\subsubsection*{Robotic ant maze}
Based on the setup introduced by Cideron et al. \cite{cideron2020qd}, it consists of a 4-legged robotic ant in a maze. 
There are two goal areas and the task is for the ant to navigate the maze and reach the center of one of them.
The robot is controlled by a 3-layers \gls{nn}, with each layer of size 10. 
The input of the controller is the 29-dimensional observation returned by the environment at each step, while its output is the 8-dimensional joint's torque control.
The size of the parameter space $\Theta$ is 574, and each policy is run in the environment for $3000$ timesteps.
\newline

For all environments, the reward is given only if inside the reward area, and as a continual value in the $[0, 1]$ range. 
The reward varies with the distance to the center of the area and is highest directly at the center.
It can be expressed as:
\begin{equation*}
    r(\theta) = \begin{cases} 0, & \mbox{if } d_r > radius \\ \frac{radius - d_r}{radius}, & \mbox{if } d_r \leq radius \end{cases}
\end{equation*}
where $d_r$ is the distance from the center and $radius$ is the radius of the reward area.

Structuring the reward in this way allows to have a reward gradient once inside the reward area, providing the possibility to improve on it and to highlight the advantages provided by the emitters.
Were the reward to be binary, there would be no need to use emitters because there is no improvement to be done on the reward once discovered, thus vanilla \gls{ns} would be enough.

\newpage
\subsection*{Baselines}
\gls{serene} is compared against 5 different baselines:
\begin{itemize}
    \item \textbf{\gls{ns}}\cite{Lehman2008NS}: vanilla \gls{ns}, that performs pure exploration and does not attempt to improve on the reward;
    \item \textbf{\gls{moo}-NR}\cite{deb2002fast}: a multi-objective evolutionary algorithm optimizing both the novelty and the reward; 
    \item \textbf{CMA-ME}\cite{fontaine2020covariance}: the original algorithm introducing emitters that combines \gls{me} with emitters over a $50 \times 50$ grid covering the behavior space of all environments. 
    Among the various emitters proposed with CMA-ME \cite{fontaine2020covariance}, the ``optimizing'' emitter was selected;
    \item \textbf{ME}\cite{mouret2015illuminating}: vanilla MAP-Elites that uses a $50 \times 50$ grid to cover the behavior space of every environment;
    \item \textbf{RND}: pure random search in which no selection happens, and every policy is sampled from a normal distribution $\mathcal{N}(0, I)$.
\end{itemize}

For each experiment the total evaluation budget is $Bud=500000$, with the chunk size set to $K_{Bud}=1000$.
The population size is $M=100$, and each policy generates $m=2$ offspring.
The mutation parameter is set to $\sigma=0.5$, while the number of policies uniformly sampled to be added to the novelty archive is $N_Q=5$.
\gls{serene} uses an emitter population size of $M_{\mathcal{E}}=6$, with a bootstrap phase for each emitter of $\lambda=6$ generations.
The implementation of CMA-ME uses the same parameters used in Fontaine et al. \cite{fontaine2020covariance}: 15 emitters, each one with a population size of 37.
In every experiment, the policies parameters are clipped in the $[-5, 5]$ range.
Finally, the statistical results are computed over 15 runs for each experiment.
\section{Results}
\label{sec:ser_discussion}
This section discusses the results obtained during the experiments.
\subsection{Budgeting}
\label{sec:ser_budget}
Balancing the exploration of the search space and the exploitation of the reward is an aspect of paramount importance for reward-based algorithms.
Even more so in sparse reward environments.
This balance can be studied by analyzing the amount of evaluation budget dedicated to either one of the two aspects.
The exploration budget consists of all the evaluated policies that did not get any reward.
On the contrary, the exploitation budget is obtained by counting all the evaluated policies that collected some reward from one of the reward areas.

\begin{figure}[!h]
    \centering
    \includegraphics[width=\linewidth]{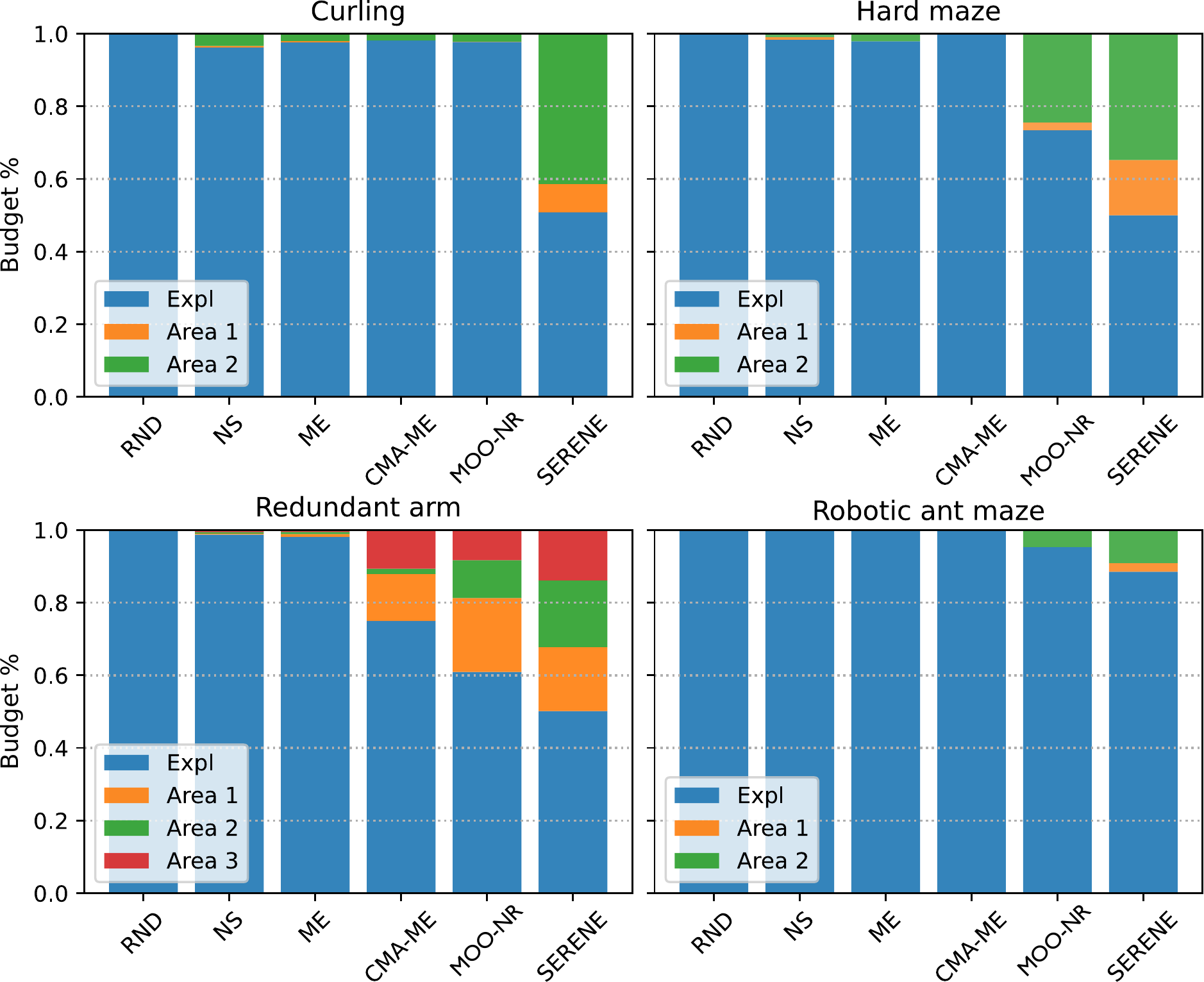}
    \caption[SERENE budgeting results]{Average budget percentage between the exploration of the search space (in blue) and the exploitation of each reward areas (other colors).
    }
    \label{fig:ser_focus}
\end{figure}

As Figure \ref{fig:ser_focus} shows, \gls{serene} has a more balanced budget split between exploration (in blue) and exploitation (other colors) compared to the other baselines.
In situations in which exploration is harder, a bigger part of the budget is assigned to exploration rather than exploitation of the reward.
This is the case for the robotic ant maze environment.
Additionally, due to the way emitters are selected, the algorithm can shift its exploitation focus among the different reward areas.
Figure \ref{fig:ser_focus} shows that most of \gls{serene}'s exploitation budget is assigned to the green reward area in the Curling, Hard maze and Robotic ant maze environments.
As it can be seen in Figure \ref{fig:ser_environments}, this area is more difficult to discover and to reach with respect to the orange area.
This makes the exploitation of the orange reward area faster, having both the novelty and the improvement go to zero rapidly.
On the contrary, the novelty of the harder to reach green area remains higher for longer, making \gls{serene} more likely to select emitters focused on it.
The effect can also be seen in Figure \ref{fig:ser_rew}, where the reward for area 1 quickly reaches higher values compared to the one of reward area 2. 
At the same time, in the Redundant arm environment where the 3 reward areas are equally easy to discover and to reach, this effect is less present and the exploitation budget is more evenly split between them. 
The ability to switch its focus is similar to intrinsic motivation based methods \cite{gottlieb2013information, blaes2019control} and allows \gls{serene} to reach high rewards in all reward areas.
Other baselines exhibit a less balanced distribution of the evaluation budget, as they do not explicitly separate exploration from exploitation.

\subsection{Exploration}
\label{sec:ser_exploration}
\begin{figure*}[!h]
    \centering
    \includegraphics[width=\linewidth]{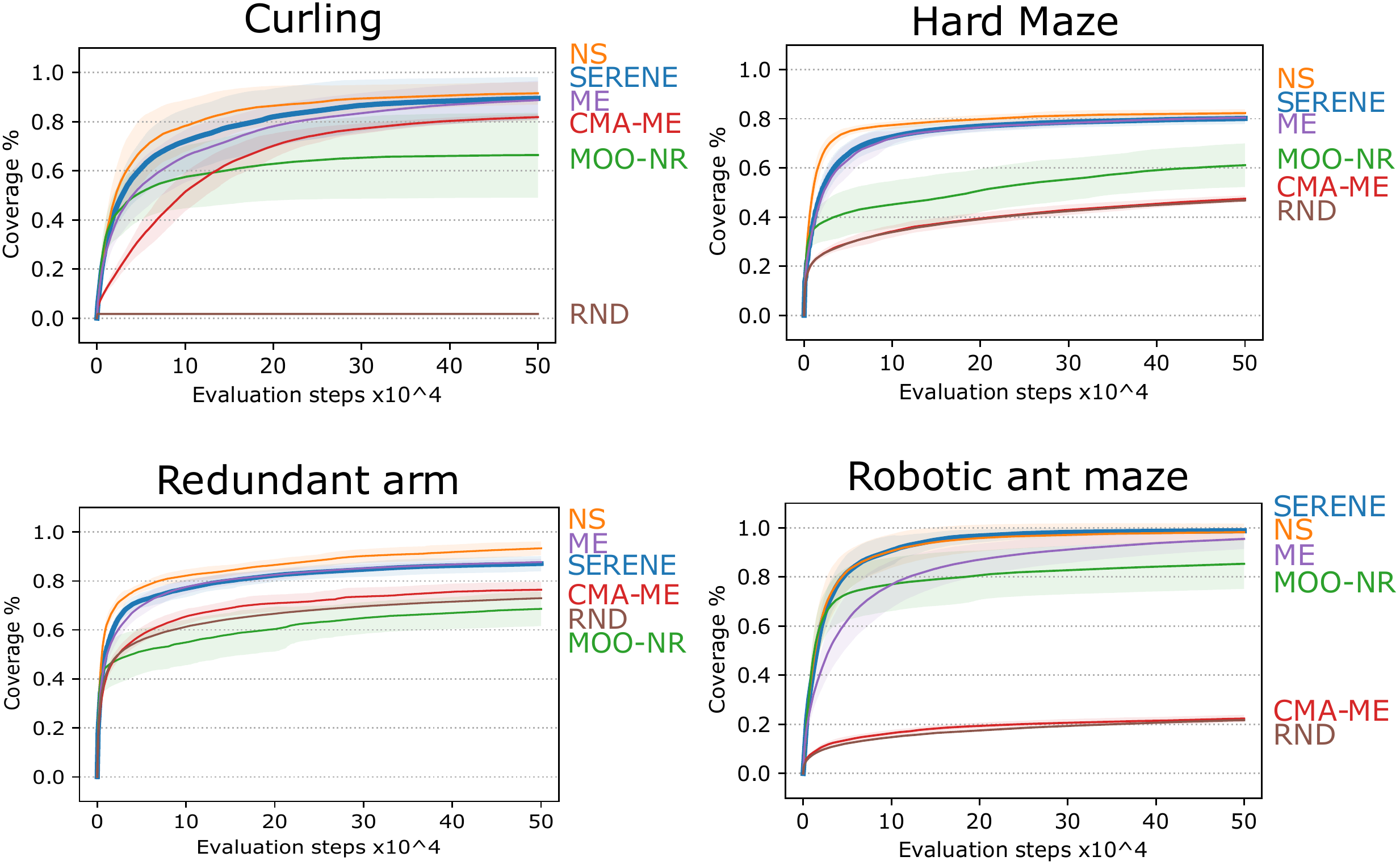}
    \caption[SERENE coverage results]{Average coverage with respect to the given evaluation budget. The shaded areas represent one standard deviation.}
    \label{fig:ser_cvg}
\end{figure*}
Performing good exploration in situations of sparse rewards is fundamental in order to discover all the possible rewarding areas of the search space.
In the experiments, the exploration capacity of each of the tested algorithms was measured through the \emph{coverage metric} \cite{mouret2015illuminating, paolo2019unsupervised}.
It is evaluated by discretizing the search space in a $50 \times 50$ grid and calculating the percentage of cells occupied by the policies found during the search.
This metric does not include any measure of the performance of the solutions in the cells.

The plots in Figure \ref{fig:ser_cvg} show that \gls{serene} can perform exploration with an efficiency comparable to \gls{ns}, notwithstanding the lower budget assigned to exploring the search space.
At the same time, Figure \ref{fig:ser_cvg} shows that the final coverage obtained by \gls{me} is similar to the one of \gls{ns} and \gls{serene}.

On the contrary, although based on \gls{me}, CMA-ME results are more variable across all environments, and exhibit lower exploration compared to \gls{me}.
This effect is likely due to the reliance on emitters for exploration, leading to more local exploration in the parameter space $\Theta$.
It can prove useful in environments like Curling or Redundant arm, where a small change in parameters leads to big behavioral changes, increasing the probability of finding a reward.
On the contrary, environments like Hard Maze or Robotic ant maze in which this does not happen can prove more challenging to explore.

At the same time, the exploration performance of \gls{moo}-NR is poor.
In the Redundant arm environment, exploration is even lower than the random search baseline.
This result is likely due to the multi-objective approach of optimizing both novelty and reward through Pareto fronts.
With time, finding more novel parts of the environment becomes increasingly more difficult.
This is due to the rewarding solutions dominating all policies outside of the reward areas that have no reward.
The non-rewarding solutions - that can foster exploration towards unexplored areas of the search space - will be then less likely to be selected.
This biases the algorithm towards the exploitation of the already discovered rewards rather than the exploration of the search space.

\subsection{Exploitation}
\label{sec:ser_exploitation}
\begin{figure*}[!hp]
    \centering
    \includegraphics[width=\textwidth]{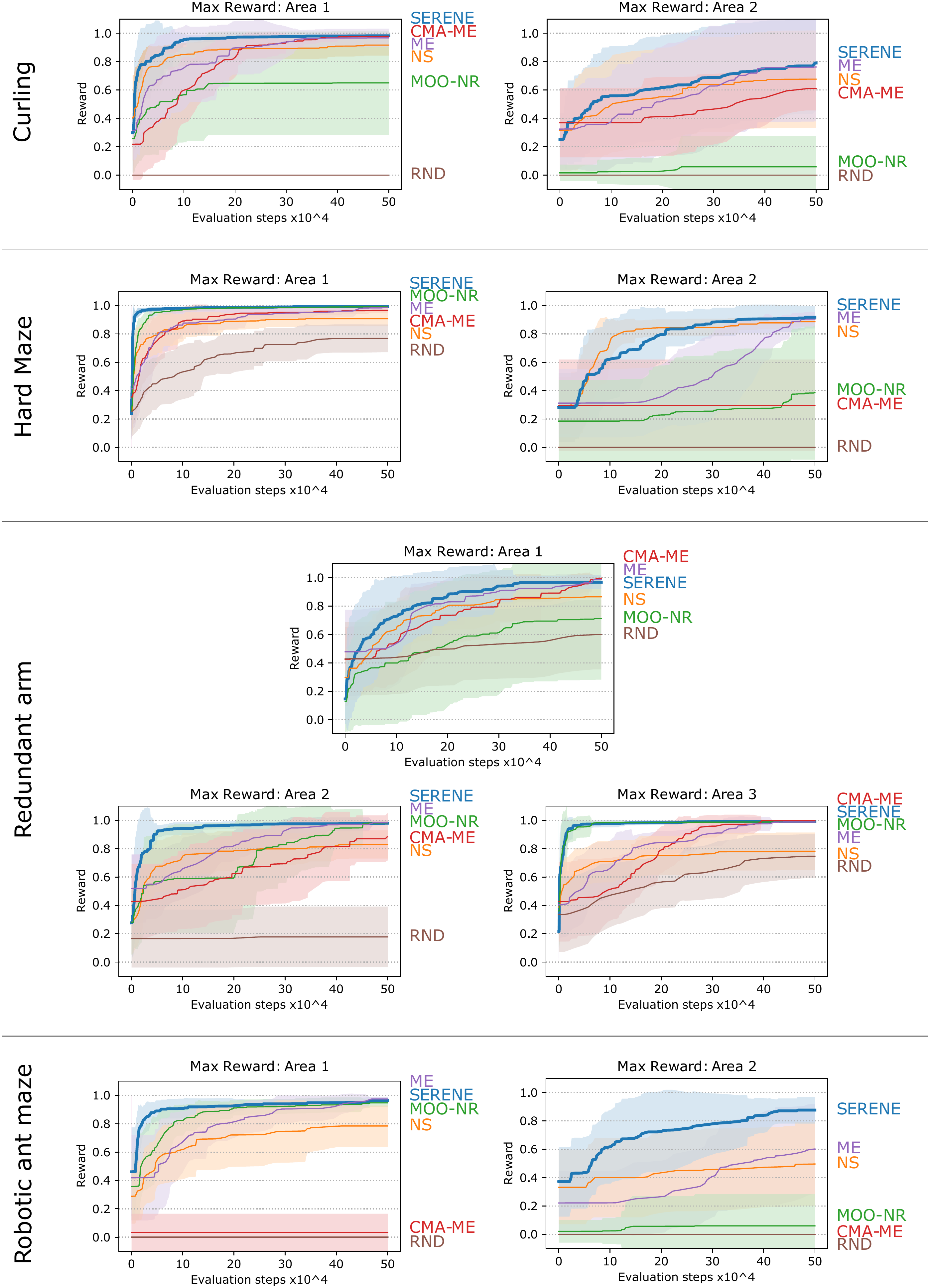}
    \caption[SERENE reward results]{Average maximum reward reached in all the reward areas. The shaded areas represent one standard deviation.
    }
    \label{fig:ser_rew}
\end{figure*}
Fig. \ref{fig:ser_rew} shows the average maximum reward achieved by the algorithms in the reward areas of all environments.
Emitters solely focusing on exploiting the reward allow \gls{serene} to reach almost the maximum reward on the easiest to reach reward areas in less than $10^5$ evaluations.
High rewards are also achieved on the harder to reach areas, even if the required time is higher.
On the contrary, \gls{me} improves on the reward at a much slower pace.
This is likely due to the random selection of policies from the archive to generate new policies.
In a sparse reward environment in fact, the probability of selecting a rewarding policy is proportional to the ratio between the rewarding and non-rewarding areas.
The sparser the reward is, i.e. the smaller the reward area is, the lower the probability of selecting a rewarding policy from the archive is, and the slower the exploitation gets. 
A similar trend is exhibited by CMA-ME: even if able to reach high rewards on the discovered reward areas, it is slow in its optimization.
At the same time, even \gls{ns} reached high rewards on almost all environments, but without any explicit reward optimization it did not exploit the reward areas to the maximum.
The multi-objective approach \gls{moo}-NR can always find at least one of the multiple reward areas, but then tends to extensively focus on it, instead of also exploring other areas.
For this reason only the easiest reward area is exploited to high values in all environments, while the harder reward area is seldom exploited.

\subsection{Final archive distribution}
\begin{figure*}[!hp]
    \includegraphics[width=\textwidth]{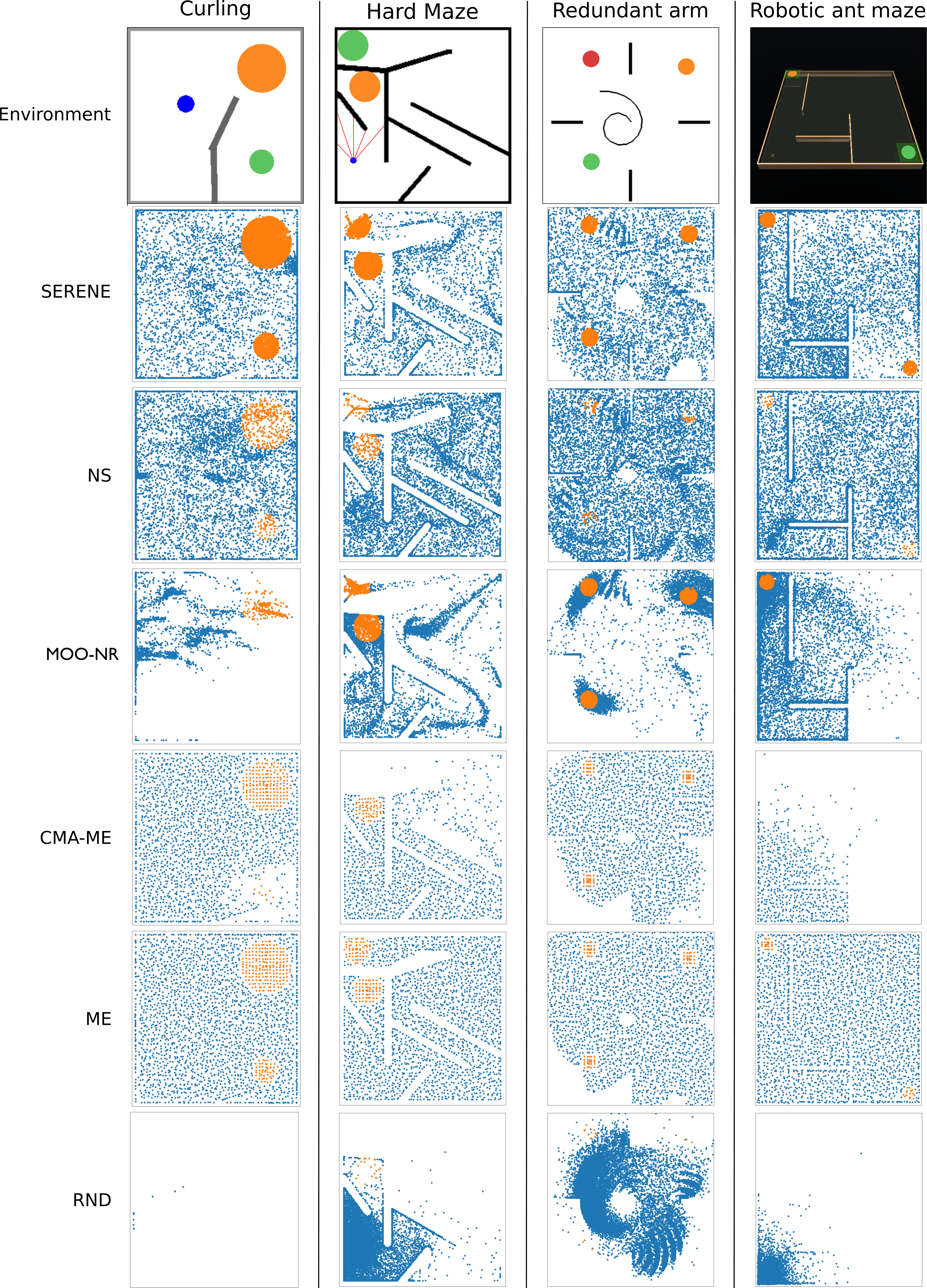}
    \caption[SERENE final archives]{Distribution of the behavior descriptors of the archived policies. On each column are shown the results for an environment, while on each row is shown the distribution for each experiment. The archive plotted are from the runs achieving highest coverage. In blue are the policies with no reward, in orange the policies with a reward. For \gls{serene} in blue are the policies in the \emph{novelty archive} and in orange the policies in the \emph{reward archive}. }
    \label{fig:ser_archives}
\end{figure*}

Fig. \ref{fig:ser_archives} shows the distribution of the behaviors of the policies in the final archive.
Each point represents a different policy.
In blue are the policies that do not get any reward, thus considered \emph{exploratory}, while in orange are rewarding policies, considered \emph{exploitative}.
For \gls{serene} the \emph{exploratory policies} are the ones in the \emph{novelty archive} $\mathcal{A}_N$, while the \emph{exploitative policies} are the ones in the \emph{rewarding archive} $\mathcal{A}_R$.

From the figure, it is possible to see that \gls{serene} covers the search space well, in a fashion similar to \gls{ns}, with the exception of the reward areas, where \gls{serene}'s emitters allow to have a denser exploration (and thus exploitation of the reward).
A similar dense exploration of the reward areas is performed by \gls{moo}-NR but at the cost of exploration.
This is likely due to the fact that once a reward is discovered, the reward scales more than the novelty, making the algorithm focus more on the reward improvement than the exploration.
In the Curling and Robotic ant maze environments this effect is so strong that it prevents the discovery of all the reward areas.
At the same time, and contrary to \gls{ns} based methods, both \gls{me} and CMA-ME can explore the search space in a much more uniform fashion, thanks to the discretization of the search space performed by the algorithm.
While this discretization is enough to push the exploration to cover the whole space for \gls{me}, this is not the case for CMA-ME, as it can be seen especially in the hardest to explore environment: the Robotic ant maze.
This is the result of relying on emitters for both reward exploitation and exploration.
This strategy forces emitters to both diversify the behavior of the discovered policies, trying to cover as much search space as possible, and optimize the policies performances, focusing on a very narrow area of the whole search space.

\section{Conclusion}
\label{sec:ser_conclusion}
In this chapter \gls{serene} was introduced, a method that efficiently deals with sparse reward environments by augmenting \gls{ns} with emitters.
Contrary to similar methods using emitters, \gls{serene} keeps exploration and exploitation of the reward as two distinct processes.
Exploration is carried out by taking advantage of \gls{ns} to discover all the reachable reward areas.
These areas are then exploited by using local instances of population-based optimization algorithms called emitters.
By using a meta-scheduler, \gls{serene} can automatically assign the evaluation budget to either exploration or exploitation.
This is advantageous also in situations in which no reward is present: in the absence of reward to exploit, \gls{serene} performs exactly like \gls{ns}.

\gls{serene} has been tested on four different sparse reward environments, reaching high performances on all of them.
At the same time, many kind of emitters can be used to address different kind of problems \cite{fontaine2020covariance, cully2020multi}.
The implementation of various types of reward-based algorithms as emitters and their combination can prove an exciting line of work to extend the current method to new domains. 

Notwithstanding these encouraging results, the method still suffers from the same limitations as other \gls{qd} methods, and first and foremost from the prior hand-design of the behavior space $\mathcal{B}$.
The next chapter will present a method that can learn a behavior descriptor useful for exploration while exploiting any discovered reward through emitters.
This is done by building on the ideas introduced through \gls{taxons} and \gls{serene}.
\clearpage\null\thispagestyle{empty}

\chapter{STAX}
{\hypersetup{linkcolor=black}\minitoc}
\label{chap:stax}
\section{Introduction}
\label{sec:stax_intro}
Previous chapters addressed two of the major shortcomings of \gls{ns} when dealing with sparse reward systems.
Chapter \ref{chap:taxons} introduced \gls{taxons} to reduce the amount of prior information needed to be specified at design time.
This is done by learning the behavior descriptor during the search in an autonomous way through an \gls{ae}.
Then, in chapter \ref{chap:serene}, the other limitation of \gls{ns}, even more relevant in sparse rewards settings, was addressed: the inability to optimize any reward discovered during the search.
When discovering a reward, in fact, the algorithm should be able to shift its focus on it in order to find the best possible solutions.
\gls{serene} does that through the use of emitters, by performing local search around any possible reward discovered.
The two methods address separate but complementary limitations of \gls{ns}.
In this chapter, the two algorithms are brought together by introducing \gls{name}.
This algorithm uses \gls{taxons} to perform exploration while learning a low-dimensional representation of the search space; the discovery of a reward triggers the instantiation of an emitter around the rewarding policy in order to locally explore and exploit the reward.
\gls{name} performs the exploitation of the reward in the same way \gls{serene} does: through emitters and through the alternating of the exploration and exploitation steps thanks to a meta-scheduler.

The advantages of this method are twofold: it removes the need of hand-designing a \gls{bs} by directly learning a low-representation of the search space from high-dimensional observations of the policies behaviors.
At the same time, it also removes the reward-related limitation of \gls{ns}-based approaches by swiftly exploiting any reward discovered through emitters.
All of this allows \gls{name} to efficiently deal with sparse reward environments with minimum prior information required about the task at design time.

\section{Method}
\label{sec:stax_method}
As stated in the previous section, \gls{name} deals with the limitations of \gls{ns} for \emph{sparse rewards} settings by separating the search process in two alternating sub-processes: one performing \emph{exploration} of the search space and another performing \emph{exploitation} of any discovered reward.
This allows \gls{name} to find different high reward policies with minimal prior information about the task.
This is done through a \emph{meta-scheduler} whose task is to split the total evaluation budget $Bud$ in small chunks of size $K_{Bud}$ and assigning them to either one of the two sub-processes, in the same way the \gls{serene} algorithm discussed in Chapter \ref{chap:serene} does.

While \gls{serene} and \gls{name} perform \emph{exploitation} in the same way, that is through emitters spawned around rewarding solutions, as shown in Alg. \ref{alg:ser_emitter}, the main difference between the two algorithms lies in the way \emph{exploration} is performed.
Rather than relying on \gls{ns} to explore an hand-designed \gls{bs}, as \gls{serene} does, \gls{name} takes advantage of \gls{taxons} to learn a \gls{bs} representation while performing the search in it.
This reduces the amount of prior information needed to solve the task by removing the requirement of hand-designing the \gls{bs}.

As explained in Chapter \ref{chap:taxons}, \gls{taxons} learns a \gls{bs} through the use of an \gls{ae} that is trained online on the data generated by the evaluation of the policies $\theta_i \in \Theta$.
The encoder part of the \gls{ae} can then be used as \emph{observation function} and its \emph{feature space} $\mathcal{F}$ as behavior space $\mathcal{B}$, also called \emph{outcome space}.
The way the different spaces are connected through the \gls{ae} can be defined as in Eq. \eqref{eq:tax_outcome_space}, reported here in Eq. \eqref{eq:outcome_space}:
\begin{equation}
\label{eq:outcome_space}
\begin{split}
    E: \mathcal{O} \rightarrow \mathcal{F} \equiv \mathcal{B}\\
    D: \mathcal{F} \rightarrow \mathcal{O}
\end{split}
\end{equation}
where $\mathcal{O}$ is the observation space, $E$ is the encoder of the \gls{ae} and the $D$ is the decoder.
There are two main differences between the exploration performed by \gls{taxons} and \gls{name}: the policy selection and the \gls{ae}'s training regime.

\subsection{Policy Selection}
\label{sec:stax_policy_selection}
As discussed in Chapter \ref{chap:taxons}, \gls{taxons} drives the search for novel policies through two different metrics: \emph{novelty} and \emph{surprise}.
The first one is calculated as the average distance between the policy $\theta_i$ and the $|J|$ closest other policies in the learned outcome space, as define in Eq. \eqref{eq:tax_ae_novelty}.
The \emph{surprise} is calculated as the \gls{ae}'s \emph{reconstruction error} over the observations generated by the policy $\theta_i$, as defined in Eq. \eqref{eq:tax_surprise_ae}.
A higher surprise on an outcome $o_T$ implies that the \gls{ae} has not seen that outcome very often, thus the area of the observation space $\mathcal{O}$ close to $o_T$ has not yet been properly explored.
This means that by selecting policies whose outcome has higher surprise the algorithm can push towards more exploration.

\gls{name} also uses these two metrics, but instead of using only the last observation $o_T$ to calculate them, multiple observations sampled along the trajectory are used.
As discussed in Chapter \ref{chap:signatures}, this allows to remove the assumption of the last observation encoding enough information to describe the whole behavior of a policy $\theta_i$.
This requires some modifications to the way the behavior descriptor is calculated with respect to \gls{taxons}.
\gls{name} builds this descriptor by concatenating the \gls{ae}'s low-dimensional representations of the sampled observations.
Eq. \eqref{eq:tax_phi} can then be rewritten as:
\begin{equation}
    \label{eq:stax_phi}
    f(\theta_i) = [\dots, E(o_{t_k}), \dots, E(o_{t_K})],
\end{equation}
where $o_{t_k}$ is the observation generated by the policy $\theta_i$ at time-step $t_k$.

\noindent At the same time, the surprise is calculated as the \emph{sum} of the reconstruction errors over each observation.
Meaning that Eq. \eqref{eq:tax_surprise_ae} can be rewritten as:
\begin{equation}
\label{eq:stax_surprise_ae}
    s(\theta_i) = \sum_{k \in K}\big|\big|o_{t_k}^{(\theta_i)} - D\big(E(o_{t_k}^{(\theta_i)})\big)\big|\big|^2,
\end{equation}
where $K$ is the list of indexes of the selected time-steps along the trajectory.

Other than the way novelty and surprise are calculated, \gls{name} differs from \gls{taxons} in the way these metrics are exploited.
At each generation $g$, \gls{taxons} randomly chooses only one between novelty and surprise to select the best policies.
On the contrary, during \gls{name}'s exploration step, the two metrics are combined through the NSGA-II multi-objective approach \cite{deb2002fast} detailed in Sec. \ref{sec:related_moo}.
This means that the algorithm can optimize at the same time both the novelty and the surprise.
The whole exploration process is shown in Alg. \ref{alg:stax_exploration}.

\begin{algorithm}[!h]
\caption{\gls{name} Exploration Phase}\label{alg:stax_exploration}
\textbf{INPUT:} budget chunk $K_{Bud}$, number of offspring per parent $m$, mutation parameter $\sigma$, novelty archive $\mathcal{A}_{\text{Nov}}$, candidate emitters buffer $\mathcal{Q}_{\text{Cand\_Em}}$, population $\Gamma_g$, number of policies $N_Q$, autoencoder \gls{ae}\;
\While{$K_{Bud}$ not depleted}{
    Generate offspring $\Gamma^m_g$ from population $\Gamma_g$\;
    Evaluate $\theta_i, ~~ \forall \theta_i \in \Gamma^m_g$\;
    Calculate $b_i= \phi(\theta_i) = [\dots, E(o_{t_k}), \dots, E(o_{t_K})] ~~ \forall\theta_i \in \Gamma^m_g$\;
    Calculate $\eta(\theta_i) = \frac{1}{|J|}\sum_{j \in J}\text{dist}(b_i, b_j), ~~ \forall\theta_i \in \Gamma^m_g$\;
    Calculate $s(\theta_i) = \sum_{k \in K}\big|\big|o_{t_k}^{(\theta_i)} - D\big(E(o_{t_k}^{(\theta_i)})\big)\big|\big|^2 ~~ \forall\theta_i \in \Gamma^m_g$\;
    $\mathcal{A}_{\text{Nov}} \leftarrow N_Q \text{ samples from } \Gamma^m_g$\; 
    \If{$\phi(\theta_i) \in \mathcal{B}_{\text{Rew}}$}{
        $\mathcal{Q}_{\text{Cand\_Em}} \leftarrow \theta_i$
    }
    \CommentSty{/* NSGA-II based policy selection */}\\
    Calculate non dominated fronts $F_j, ~~ \forall\theta_i \in \Gamma^m_g \bigcup \Gamma_g$\;
    Sort fronts according to \emph{non domination}\;
    Generate $\Gamma_{g+1}$ from most non dominated solutions $\theta_i \in F_j$\;
    \If{If last front $F_J$ is partially selected}{
        Calculate \emph{crowding distance} $\forall\theta_i \in F_J$\;
        Complete filling up $\Gamma_{g+1}$ with less crowded solution $\theta_i \in F_J$\;
    }
}
\end{algorithm}

\subsection{Training of the autoencoder}
In \gls{name} the exploration is driven thanks to the \gls{ae}.
This means that the way the \gls{ae} is trained is fundamental.
In order to meaningfully look for diversity in the learned feature space $\mathcal{F}$ used as \gls{bs}, the \gls{ae} has to be trained on the data collected during the search for policies itself.
This is done in a fashion similar to what described in Chapter \ref{chap:taxons} for the \gls{taxons} algorithm, with a few differences.
\gls{taxons} built the dataset $DS$ used to train the \gls{ae} with the last observations $o_T$ generated by the policies of the last $I$ generations.
On the contrary, \gls{name} takes advantage of the alternating two-step process inherited from \gls{serene}.
During this process, the algorithm generates two collections of policies: the \emph{reward archive} $\mathcal{A}_{\text{Rew}}$ and the \emph{novelty archive} $\mathcal{A}_{\text{Nov}}$.
The dataset $DS$ is then formed with the observations generated by the policies in both archives, plus the observations generated by the last population $\Gamma_g$ and the last offspring population $\Gamma_g^m$.
The data of the archives provides a \emph{curriculum}, preventing the search to cycle back to already explored areas.
This also prevents any possible destabilization of the training process due to the ever changing data distribution used in \gls{taxons}.
Another destabilizing factor can be the training of the \gls{ae} on data collected thanks to the \gls{ae} itself.
This problem should be attenuated by training the model also on the data from policies in $\mathcal{A}_{\text{Rew}}$.
These policies are in fact selected not according to the novelty calculated thanks to the \gls{ae}, but only for their reward, that is a factor independent from the feature space.
At the same time, adding the observations from the most recent population to the training dataset helps the \gls{ae} to better represent the frontier of the explored space, towards which the search is to be pushed.
Moreover, for each policy, the observations used to form the training dataset are all the observations used to generate the behavior descriptor defined in Eq. \eqref{eq:stax_phi}.
This is different from what done for \gls{taxons}, where only the last observation $o_T$ was used.

Once the dataset $DS$ has been collected, it is split into two sub-datasets: \emph{training dataset} $DS_{\text{Train}}$ and a \emph{validation dataset} $DS_{\text{Val}}$.
For each training episode, the \gls{ae} is trained on the $DS_{\text{Train}}$.
At the end of each training epoch on $DS_{\text{Train}}$, the model validation error is calculated on $D_{\text{Val}}$.
The training episode is stopped if the error increases for 3 consecutive epochs.
Contrary to \gls{taxons}, here the training episodes do not happen at regular intervals.
Instead, the \gls{ae} is trained less and less frequently the longer the search is performed; this is the same strategy employed in the AURORA method \cite{cully2019aurora}.
Training according to this strategy allows to adapt the frequency of the training to the maturity of the learned \gls{bs}.
In fact, after the first training episodes, the learned representation is mature enough for the \gls{ae} to start focusing on its refinement.
This means that there is no reason anymore to train the \gls{ae} too often, allowing to save time and computational resources.
Moreover, by training less frequently, the possible overfitting of the \gls{ae} on the data present in the archives is limited.
This shifting training regime is obtained by performing the training process every $TI$ exploration steps.
At the beginning of the search, \gls{name} sets $TI = 1$.
Its value is then increased by $1$ every time a training episode is performed.

Finally, at the end of each training episode, the behavior descriptor of all the policies present in the archives and in the populations is updated with the new descriptors generated by the retrained \gls{ae}.
This allows to keep the behavior descriptors and the novelty measurements of the policies consistent and meaningful.
The complete \gls{name} algorithm is shown in Alg. \ref{alg:stax_main}.
\begin{algorithm}[!h]
\caption{\gls{name}}\label{alg:stax_main}
\textbf{INPUT:} evaluation budget $Bud$, budget chunk size $K_{Bud}$, population size $M$, emitter population size $M_{\mathcal{E}}$, offspring per policy $m$, mutation parameter $\sigma$, number of policies added to novelty archive $Q$, \gls{ae} training interval $TI$, randomly initialized \gls{ae}\;
\textbf{RESULT:} Novelty archive $\mathcal{A}_{\text{Nov}}$, rewarding archive $\mathcal{A}_{\text{Rew}}$, trained \gls{ae}\;
$\mathcal{A}_{\text{Nov}} = \emptyset$\;
$\mathcal{A}_{\text{Rew}} = \emptyset$\;
$\mathcal{Q}_{\text{Em}} = \emptyset$\;
$\mathcal{Q}_{\text{Cand\_Nov}} = \emptyset$\;
$\mathcal{Q}_{\text{Cand\_Em}} = \emptyset$\;
$D = 0$\;
Initialized training counter $TI_C = 0$\;
Sample population $\Gamma_0$\;
Split $Bud$ in chunks of size $K_{Bud}$\;
\While{$Bud$ not depleted}{
    \If{$\Gamma_0$}{
        Evaluate $\theta_i, ~~ \forall \theta_i \in \Gamma_0$\;
    Calculate $b_i = \phi(\theta_i) \in \mathcal{B}, ~~ \forall\theta_i \in \Gamma_0$\;
    }
    \emph{Exploration Phase} ($K_{Bud}$, $m$, $\sigma$, $\mathcal{A}_{\text{Nov}}$, $\mathcal{Q}_{\text{Cand\_Em}}$, $\Gamma_g$, $Q$, \gls{ae})\;
    $TI_C = TI_C + 1$\;
    \If{$TI_C == TI$}{
    $DS = $ \emph{Extract dataset}($\mathcal{A}_{\text{Nov}}$, $\mathcal{A}_{\text{Rew}}$, $\Gamma_g$, $\Gamma^m_g$)\;
    \emph{Train Autoencoder} (\gls{ae}, $DS$)\;
    \emph{Update descriptors} (\gls{ae}, $\Gamma_g$, $\Gamma^m_g$, $\mathcal{A}_{\text{Nov}}$, $\mathcal{A}_{\text{Rew}}$, $\mathcal{Q}_{\text{Em}}$, $\mathcal{Q}_{\text{Cand\_Nov}}$, $\mathcal{Q}_{\text{Cand\_Em}}$)\;
    $TI = TI + 1$\;
    $TI_C = 0$\;
    }
    
    \If{$\mathcal{Q}_{\text{Cand\_Em}} != \emptyset$ \textbf{or} $\mathcal{Q}_{\text{Em}} != \emptyset$}{
    \emph{Exploitation Phase} ($K_{Bud}$, $\mathcal{Q}_{\text{Cand\_Em}}$, $\lambda$, $m$, $\mathcal{Q}_{\text{Em}}$, $\mathcal{A}_{\text{Nov}}$, $\mathcal{A}_{\text{Rew}}$, $M_{\mathcal{E}}$)\;
    }
}
\end{algorithm}

\subsection{Reward exploitation in a learned space}
\label{stax:exploitation}
\begin{figure}[!h]
    \centering
    \includegraphics[width=\textwidth]{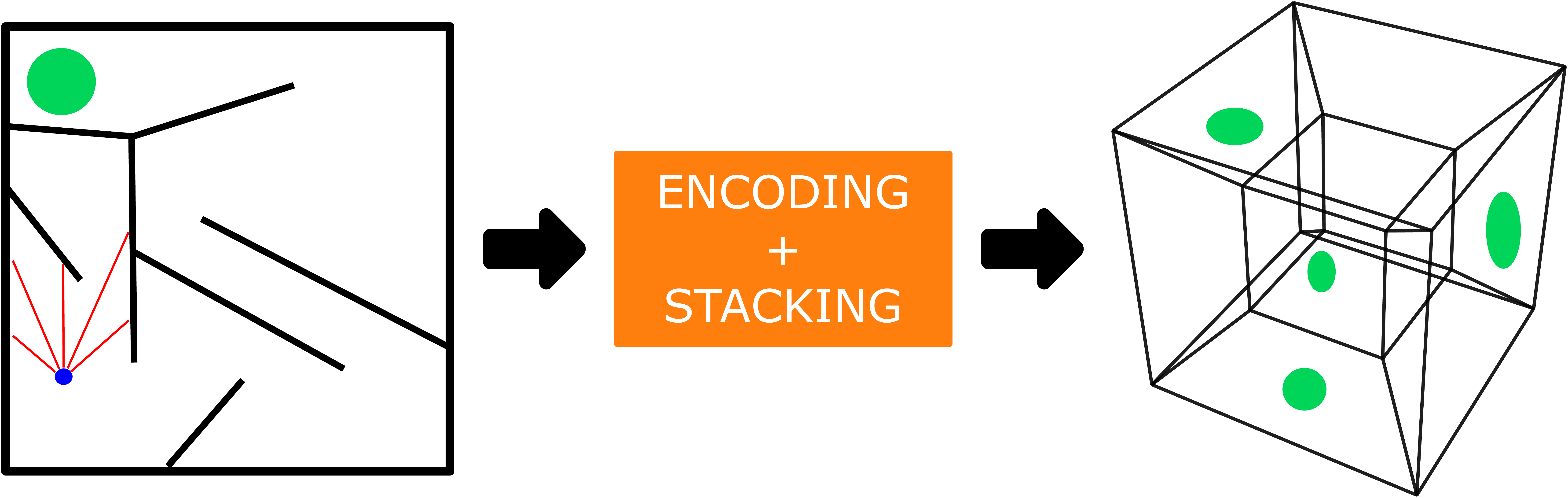}
    \caption[Example of multiplication of reward areas]{The behavior space generated by stacking the learned representations from multiple observations generated during the search can contain multiple reward areas. 
    Even if the original ground-truth space contained only one.}
    \label{fig:stax_reward_areas}
\end{figure}

As it was the case for \gls{serene}, \gls{name} exploits any discovered reward through emitters.
The power of this kind of approach has already been discussed in Chapter \ref{chap:serene}, where thanks to simple elitist reward-based emitters, \gls{serene} could easily exploit any reward area discovered during the search.
The ability to disjointly optimize multiple reward areas in an efficient way is even more fundamental for an approach like \gls{name}.
In hand-designed \gls{bs}, like the ones \gls{serene} was designed to deal with, the engineer has total control over the \gls{bs} itself.
This can help in reducing the disjointedness of the reward areas.
This is not the case when the behavior descriptor is generated by stacking multiple learned representations extracted from high-dimensional observations, as done by \gls{name}.
In this kind of settings there is no guarantee that the new \gls{bs} will have the same structure of the reward areas as the ground-truth hand-designed \gls{bs}. 
It can happen that this space will have multiple reward areas, even if only one is present in the ground-truth \gls{bs}.
This can be explained by considering the Hard Maze environment with just one reward area shown in Fig. \ref{fig:stax_reward_areas}.
In this example, the environment has a 2D ground-truth \gls{bs} consisting of the $(x, y)$ position of the robot at the end of a trajectory.
On the contrary, the \gls{bs} obtained by stacking the \gls{ae} representations extracted from multiple RGB high-dimensional observations has higher dimensionality.
This can lead to multiple areas of this higher dimensional space representing the ground-truth reward area.
The effect is even more likely in the first phases of the search, when the \gls{ae} is not yet properly trained and its feature space not completely mature.
For these reasons, using an emitter-based approach as \gls{name} capable of focusing on multiple reward areas can give a strong advantage in situations where the \gls{bs} representation is so complex.
This approach can be even more useful in settings in which the reward is difficult to express in the ground-truth \gls{bs} but easy to observe from the high-dimensional observations of the trajectory.
Being able to see the rewarding situations from these observations means that it will be likely for them to be represented in the learned \gls{bs}.
This will create zones in this space in which the reward can be achieved, allowing the algorithm to use emitters to exploit it.
The whole exploitation phase is shown in Alg. \ref{alg:stax_emitter}.

\begin{algorithm}
\caption{\gls{name} Exploitation Phase}\label{alg:stax_emitter}
\textbf{INPUT:} budget chunk $K_{Bud}$, candidate emitters buffer $\mathcal{Q}_{\text{Cand\_Em}}$, number of bootstrap generations $\lambda$, emitter population size $M_{\mathcal{E}}$, number of offspring per policy $m$, emitters buffer $\mathcal{Q}_{\text{Em}}$, rewarding archive $\mathcal{A}_{\text{Rew}}$, novelty archive $\mathcal{A}_{\text{Nov}}$\;
\CommentSty{/* Bootstrap step */}\\
\While{$\nicefrac{K_{Bud}}{3}$ not depleted}{
Select most novel policy $\theta_i$ from $\mathcal{Q}_{\text{Cand\_Em}}$\;
Calculate $\sigma_i$\;
Initialize: $\mathcal{E}_i$, $\mathcal{Q}^i_{\text{Cand\_Nov}} = \emptyset$, and $P_0$\;
\For {$\gamma \in \{0, \dots, \lambda\}$}{
    \If{$P_0$}{
        Evaluate $\Tilde{\theta}_j$, $\forall \Tilde{\theta}_j \in P_0$\;
    }
    Generate offspring population $P^m_{\gamma}$ from $P_{\gamma}$\;
    Evaluate $\Tilde{\theta}_j$, $\forall \Tilde{\theta}_j \in P^m_{\gamma}$\;
    Generate $P_{\gamma+1}$ from best $\Tilde{\theta}_j \in P^m_{\gamma} \bigcup P_{\gamma}$\;
}
Calculate $I(\mathcal{E}_i)$\;
\If{$I(\mathcal{E}_i) > 0$}{
    $\mathcal{Q}_{\text{Em}} \leftarrow \mathcal{E}_i$\;
}
}
\CommentSty{/* Emitters step */}\\
Calculate pareto fronts in $\mathcal{Q}_{\text{Em}}$\;
\While{$\nicefrac{2}{3}K_{Bud}$ not depleted}{
Sample $\mathcal{E}_i$ from \emph{non-dominated emitters} in $\mathcal{Q}_{\text{Em}}$\;

\While{\textbf{not} $terminate(\mathcal{E}_i)$}{
Generate offspring population $P^m_{\gamma}$ from $P_{\gamma}$\;

Evaluate $\Tilde{\theta}_j$, $\forall \Tilde{\theta}_j \in P^m_{\gamma}$\;

$\mathcal{A}_{\text{Rew}} \leftarrow \Tilde{\theta}_j, ~~ \forall \Tilde{\theta}_j \in P^m_{\gamma} \mid r(\Tilde{\theta}_j) > R_{\gamma}$\; 
$\mathcal{Q}^i_{\text{Cand\_Nov}} \leftarrow \Tilde{\theta}_j, ~~ \forall \Tilde{\theta}_j \in P^m_g \mid \eta(\Tilde{\theta}_j) > \eta_i$\;

Generate $P_{\gamma+1}$ from best $\Tilde{\theta}_j \in P^m_{\gamma} \bigcup P_{\gamma}$\;
Update $I(\mathcal{E}_i)$ and $R_{\gamma}$\;

\If{$terminate(\mathcal{E}_i)$}{
$\mathcal{A}_{\text{Nov}} \leftarrow N_Q \text{ samples from } \mathcal{Q}^i_{\text{Cand\_Nov}}$\;
Discard emitter $\mathcal{E}_i$\;
}
}
}
\end{algorithm}


\section{Experiments}
\label{stax:experiments}
This section studies how \gls{name} can discover highly rewarding policies while exploring an outcome space learned on the fly.
All of this with minimal previous information about the task at hand and the environment.
The performances of \gls{name} will be compared against various baselines to study which ones are the most important aspects of the method.

\begin{figure}[!h]
    \centering
    \includegraphics[width=.7\textwidth]{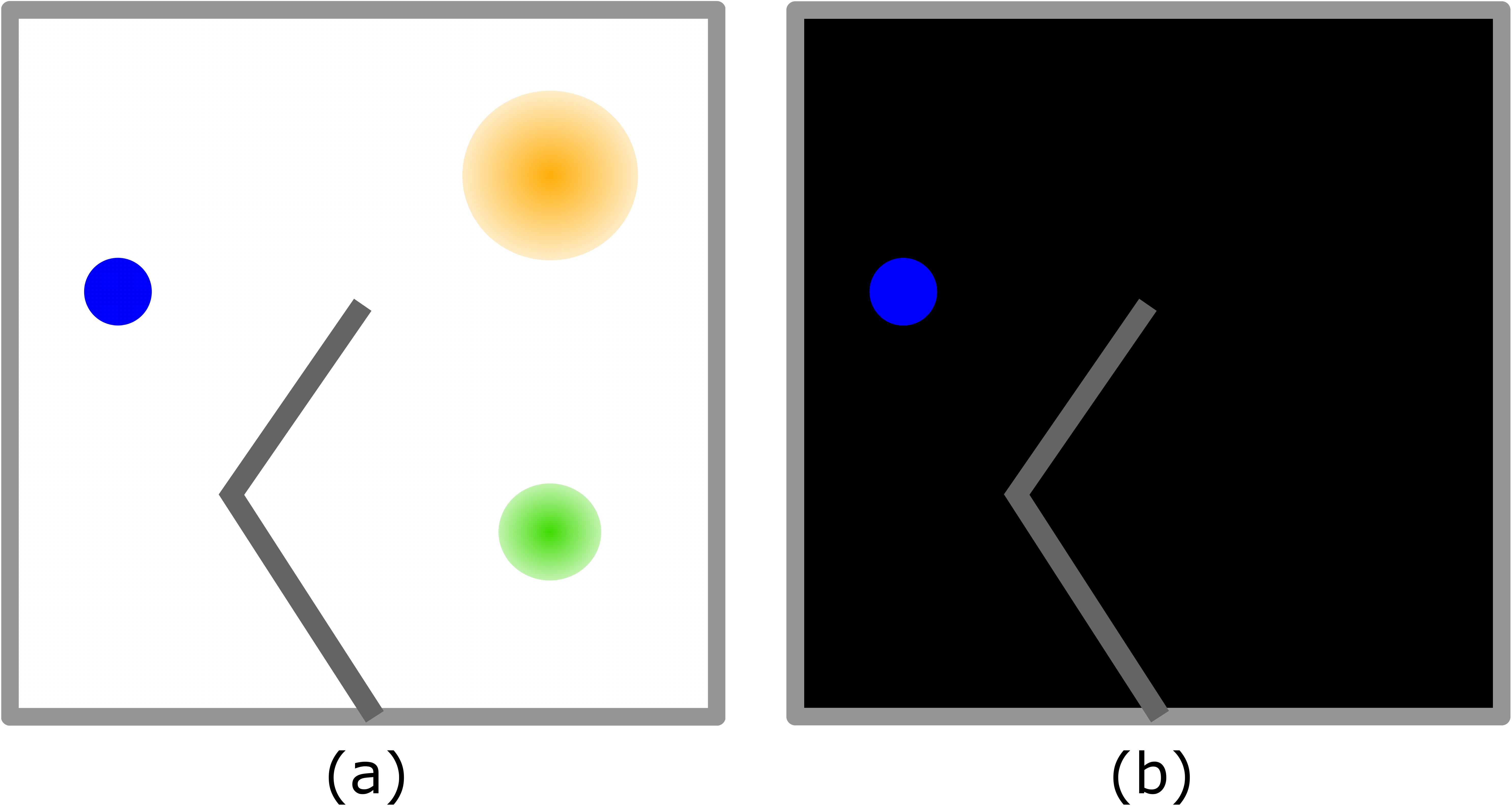}
    \caption[Curling environment]{Curling environment. (a) Real environment with reward areas. (b) RGB observation of the environment as seen by the \gls{ae}.}
    \label{fig:stax_curling}
\end{figure}

In order to perform this analysis \gls{name} is evaluated on 3 of the sparse rewards environments presented in Chapter \ref{chap:serene}:

\textbf{Curling}: it consists of a 2 \gls{dof} arm pushing a ball over a table \cite{paolo2021sparse}.
The arm is controller by a 3 layers \gls{nn} with each layer of size 5.
The input of the controller is a 6-dimensional array containing the $(x,y)$ ball pose and the two joints angles and velocities.
The controller outputs a 2-dimensional array containing the speeds of the two joints at the next time-step.
Each policy is run in the environment for 500 timesteps.
The reward is given only if the ball is in one of the two rewarding areas and is higher the closer it is to the center of the area.
The ground truth behavior descriptor used by methods that do not learn the \gls{bs} representation is the $(x,y)$ position of the ball.
The environment is shown in Fig. \ref{fig:stax_curling}. 
Fig. \ref{fig:stax_curling}.(b) shows the $64 \times 64$ RGB image the \gls{ae} sees during the algorithm execution.

\begin{figure}[!h]
    \centering
    \includegraphics[width=.7\textwidth]{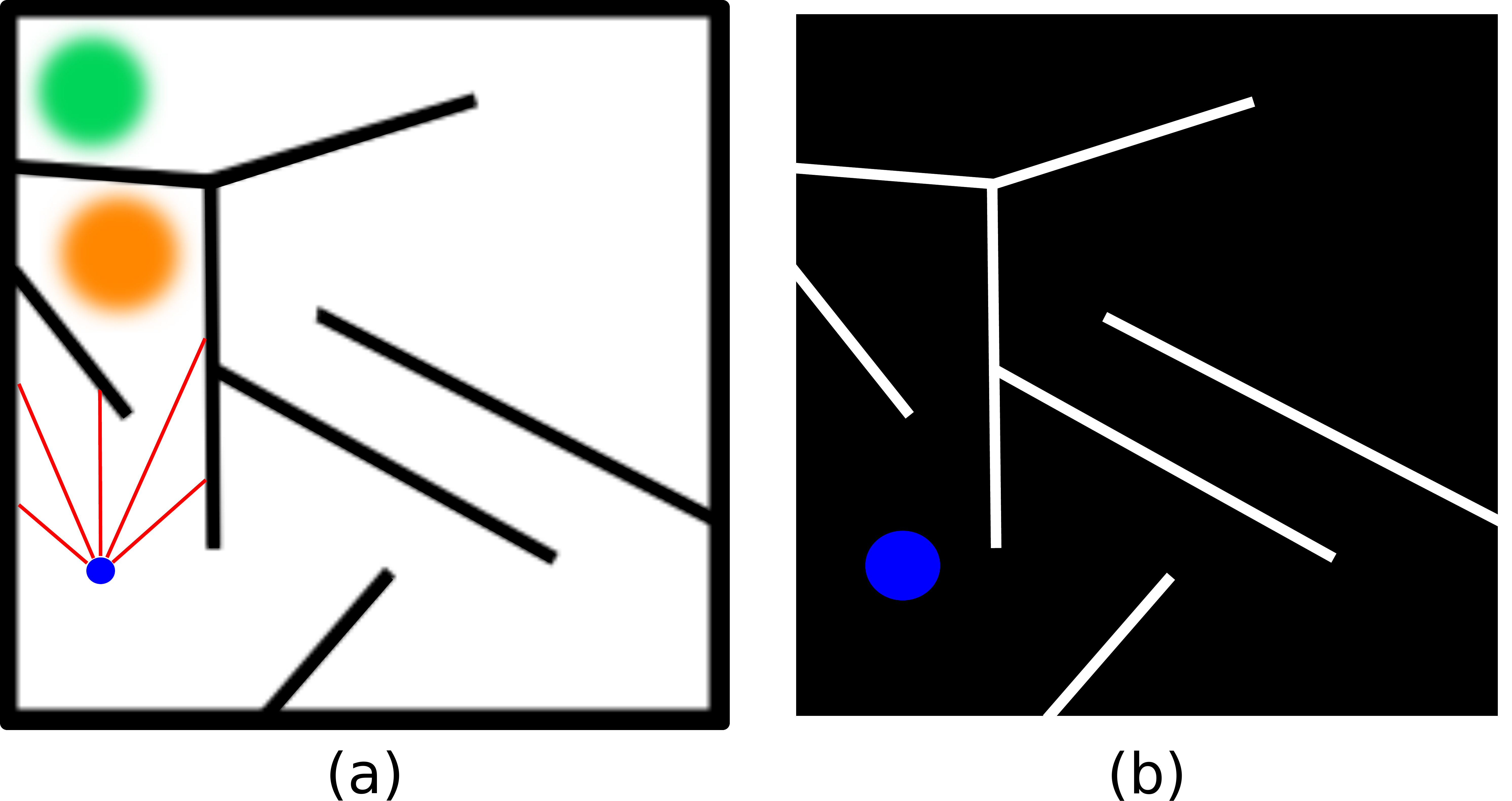}
    \caption[HardMaze environment]{HardMaze environment. (a) Real environment with reward areas. (b) RGB observation of the environment as seen by the \gls{ae}.}
    \label{fig:stax_maze}
\end{figure}

\textbf{HardMaze}: it consists of a 2-wheeled robot whose goal is to navigate a maze with the aid of 5 distance sensors \cite{Lehman2008NS}.
The robot, in blue in Fig. \ref{fig:stax_maze}, is controlled by a 2-layers \gls{nn} with each layer of size 5.
The controller receives as inputs the reading of the 5 distance sensors, shown in red in Fig. \ref{fig:stax_maze}.(a), and outputs the speed of the wheels for the next timestep.
The agent receives a reward if the robot reaches one of the 2 reward areas, with the reward being higher the closer to the center the robot stops.
Each policy is run in the environment for 2000 timesteps.
The ground truth behavior descriptor used by methods that do not learn the \gls{bs} representation is the $(x,y)$ position of the robot.
The whole environment is shown in Fig. \ref{fig:stax_maze}.(a), while in Fig. \ref{fig:stax_maze}.(b) is shown the $64 \times 64$ RGB observation fed to the \gls{ae}.  

\begin{figure}[!h]
    \centering
    \includegraphics[width=.7\textwidth]{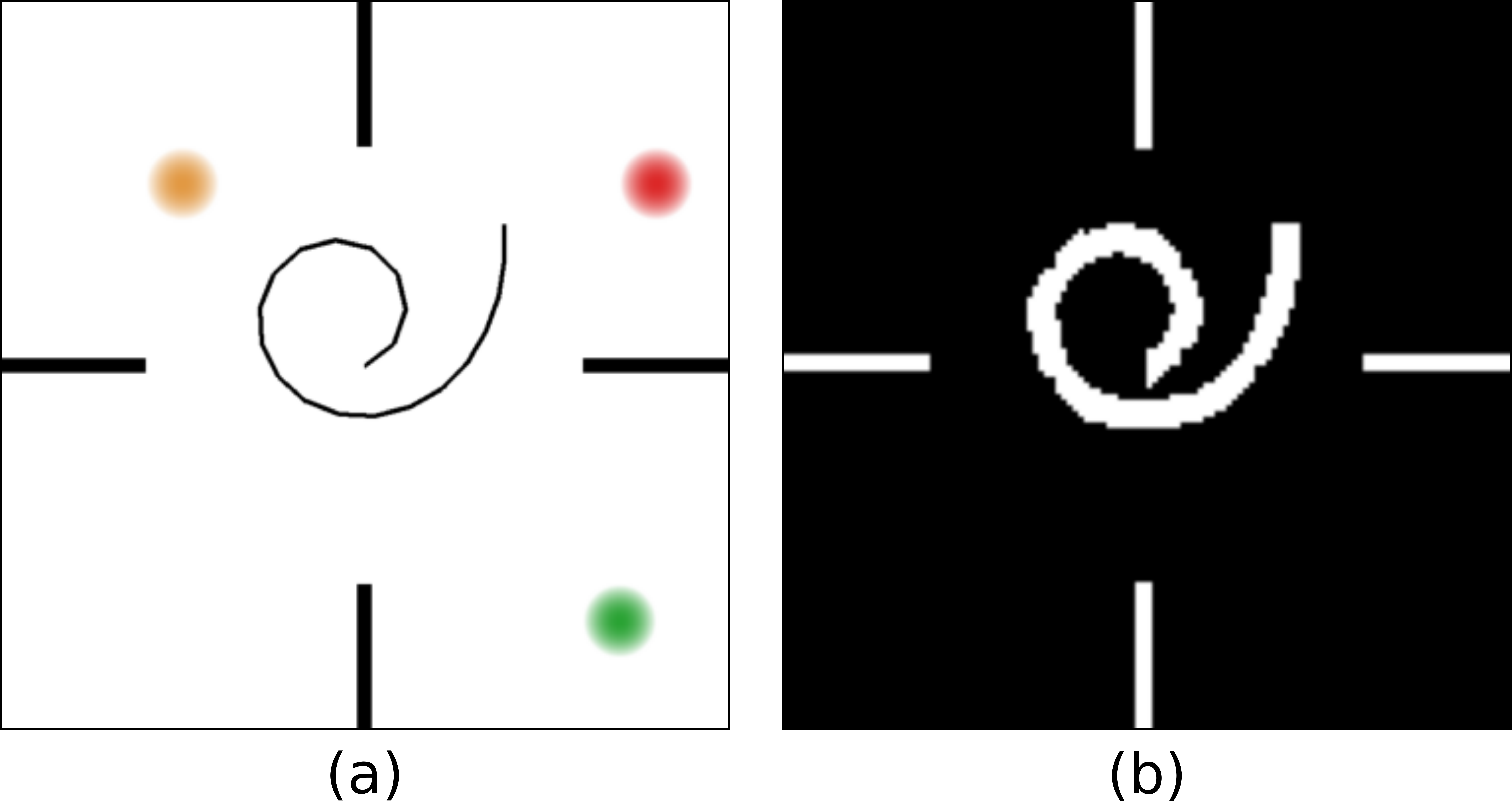}
    \caption[Redundant Arm environment]{Redundant Arm environment. (a) Real environment with the 3 reward areas. (b) RGB observation of the environment as seen by the \gls{ae}.}
    \label{fig:stax_arm}
\end{figure}

\textbf{Redundant Arm}: it consists of a 20-DoF arm moving on a 2 dimensional plane \cite{loviken2017online}.
The arm is controlled by a \gls{nn} with 2 layers, each one of size 5.
The input of the controller is the 20-dimensional vector of each joints' positions, while the output consists in the 20-dimensional joints' torque vector.
The policies are run for 100 timestep each, or until the arm collides either with the wall or itself.
The ground truth behavior descriptor used by methods that do not learn the \gls{bs} representation is the $(x,y)$ position of the end effector.
The reward is given if the end effector reaches one of the three highlighted areas, with the reward being higher the closer the effector is to the center of the reward area.
The environment is shown in Fig. \ref{fig:stax_arm}.(a).
Fig. \ref{fig:stax_arm}.(b) shows the $64 \times 64$ RGB image that the \gls{ae} uses to build the behavior descriptors.

In all of these environments, \gls{name} builds the behavior descriptors by stacking the low-dimensional representations extracted by the \gls{ae} from multiple high-dimensional observations. 
To this end, 5 samples collected at regular intervals along the trajectories are used during the experiments.

\subsection*{Baselines}
\gls{name} is compared against the following baselines:
\begin{itemize}
    \item \textbf{\gls{ns}} \cite{Lehman2008NS}: vanilla \gls{ns}, that performs pure exploration and does not attempt to improve on the reward;
    \item \textbf{\gls{me}} \cite{mouret2015illuminating}: vanilla MAP-Elites that uses a 50 × 50 grid to cover the behavior space of every environment;
    \item \textbf{\gls{moo}-NR} \cite{deb2002fast}: the multi-objective evolutionary algorithm introduced in Sec. \ref{sec:related_moo}. It optimizes both the novelty and the reward of the policies;
    \item \textbf{\gls{taxons}} \cite{paolo2019unsupervised}: it performs pure exploration by learning the behavior descriptor through an \gls{ae} trained during the search process;
    \item \textbf{\gls{serene}} \cite{paolo2021sparse}: it performs exploration through \gls{ns}, exploiting any discovered reward through emitters.
\end{itemize}

\begin{figure}[!h]
    \centering
    \includegraphics[width=\textwidth]{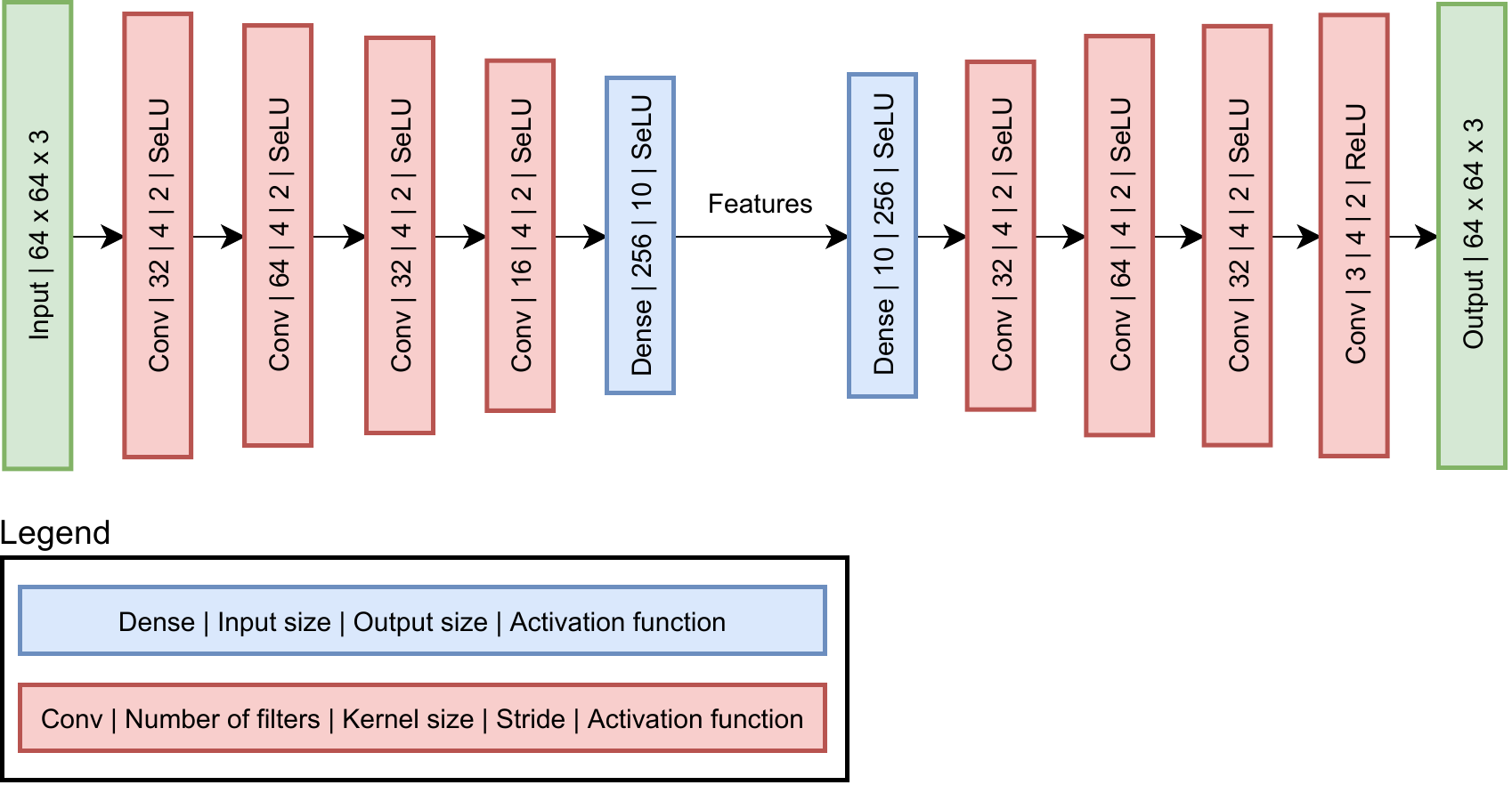}
    \caption[STAX AE structure]{\gls{ae} structure. The input and output of the network are represented in green; in red are the convolutional layers, while in blue are the fully connected layers.}
    \label{fig:stax_ae}
\end{figure}

For each experiment the given evaluation budget is $Bud = 500000$, with a chunk size of $K_{Bud}=100$.
The population used has a size of $M=100$ and each policy generates $m=2$ offsprings.
This is done by using a mutation parameter of $\sigma =0.5$.
At each generation, the number of policies sampled to be added to the novelty archive is $N_Q=5$.
The emitters have a population size of $M_{\mathcal{E}} = 6$ with a bootstrap phase of $\lambda = 6$.
For every experiment the policies parameters are bounded in the $[-5, 5]$ range.
All approaches using an \gls{ae} to represent the behavior descriptor use the same structure for it, shown in Fig. \ref{fig:stax_ae}.
The \gls{ae} consists of an encoder $E(\cdot)$ with 4 convolutional layers of sizes [32, 64, 32, 16], followed by a linear layer projecting the 256-dimensional vector returned by the last convolutional layer into the 10-dimensional feature space.
Each convolutional operation has a kernel of size 4, with a stride of 2 and a padding of 1.
Every layer is followed by a SeLU activation function \cite{Klambauer2017selu}, allowing the self-normalization of the \gls{nn}.
On the contrary, the decoder $D(\cdot)$ starts with a linear layer projecting the 10-dimensional feature vector into a 256-dimensional vector. 
Then it is followed by 4 convolutional layers of sizes [32, 64, 32, 3], each one using a kernel of size 4, a stride of 2 and a padding of 1.
Every layer uses a SeLU activation function, with the exception of the last convolutional one using a ReLU, in order to force the non-negativity of the output value.
The \gls{ae} is trained with the Adam optimizer \cite{kingma2014adam} with an learning rate of 0.001.
Finally,the statistical results are computed over 15 runs for each experiment.
\section{Results}
\label{sec:stax_results}
In this section, the results obtained during the experiments are discussed.

\subsection{Exploration}
\label{sec:stax_exploration}
\begin{figure}[!h]
    \centering
    \includegraphics[width=0.6\textwidth]{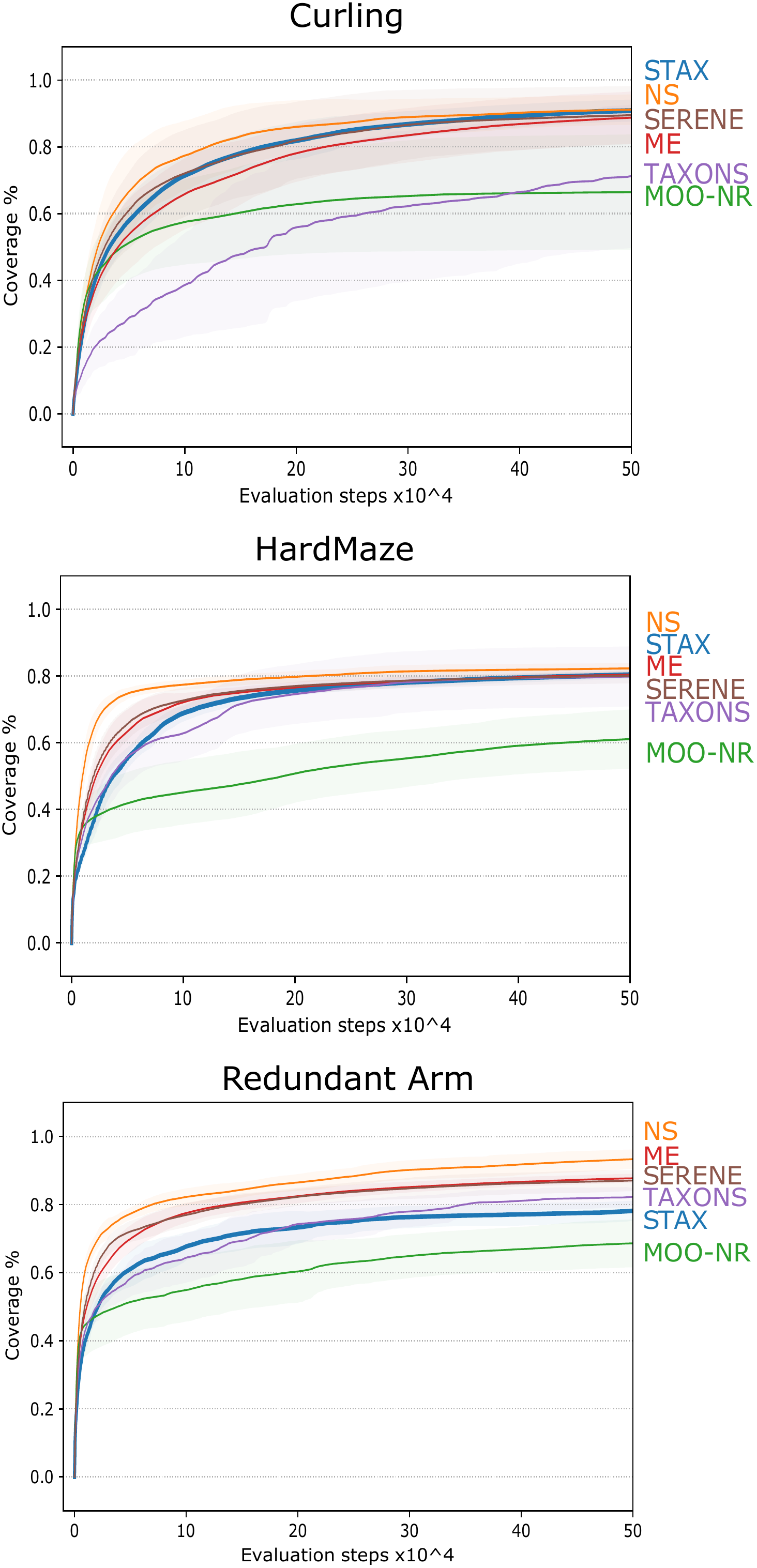}
    \caption[STAX coverage results]{Average coverage with respect to the given evaluation budget reached by \gls{name} against the different baselines. The shaded areas represent one standard deviation.}
    \label{fig:stax_cvg_baselines}
\end{figure}

Performing good exploration in situations of sparse rewards is fundamental but is not an easy task, even more so with a minimal amount of prior information, as \gls{name} does.
This section studies how well \gls{name} can explore.
This is done by measuring the \emph{coverage metric} obtained in the \emph{ground truth} \gls{bs} that are defined in Sec. \ref{stax:experiments} for each one of the tested environments.
As done in previous chapters, the coverage metric is calculated by dividing said ground truth space into a $50 \times 50$ grid and calculating the percentage of cells occupied during the search.
A cell is considered occupied if a policy reaches it at the end of its evaluation.
Note that, while the coverage is calculated in the ground-truth space, \gls{name} has no access to this space at search time.
The algorithm has to learn a representation from a collection of high-dimensional observations in order to perform the exploration.
This means that the method can also explore in areas of the learned space that are not considered by the coverage metric in the ground-truth space.

Fig. \ref{fig:stax_cvg_baselines} shows the coverage reached by our method and all the tested baselines.
It can be seen that \gls{name} can perform exploration on a level comparable with \gls{ns} on all the environments, except on the Redundant Arm, in which the coverage is lower. 
All of this while having minimal information about the environment and the task.
At the same time, the methods using the hand-designed ground-truth \gls{bs} to drive the search, that is \gls{me} and \gls{serene}, reach high levels of coverage too.
This is expected given that both method perform the search in the same space in which the coverage metric is computed.
This is not the case for \gls{moo}-NR, which struggles in all environments, likely because of the interference of the multi-objective optimization between reward and diversity maximization.

\gls{taxons} also obtains high coverage, with the notable exception of the Curling environment.
This is in contrast with the results obtained in Chapter \ref{chap:taxons}, where the coverage of \gls{taxons} in the Billiard environment was comparable to the one of \gls{ns}.
The culprit of this loss of performance is likely the presence of the 2-Dof arm in the image fed to the \gls{ae}, as shown in Fig. \ref{fig:stax_curling}.
The arm can act as a distractor and was not present in the experiments done in Chapter \ref{chap:taxons}.
At the same time, the presence of the arm is not an hindrance to the performances of \gls{name}.
This is likely be due to both the more efficient selection of new policies according to the pareto based approach, performed by \gls{name}, and the training of the \gls{ae} on the observations from policies in the reward archive.
This last element allows \gls{name} to observe many more situations in which the ball is in different positions.

\subsection{Exploitation}
\begin{figure}[!h]
    \centering
    \includegraphics[width=\textwidth]{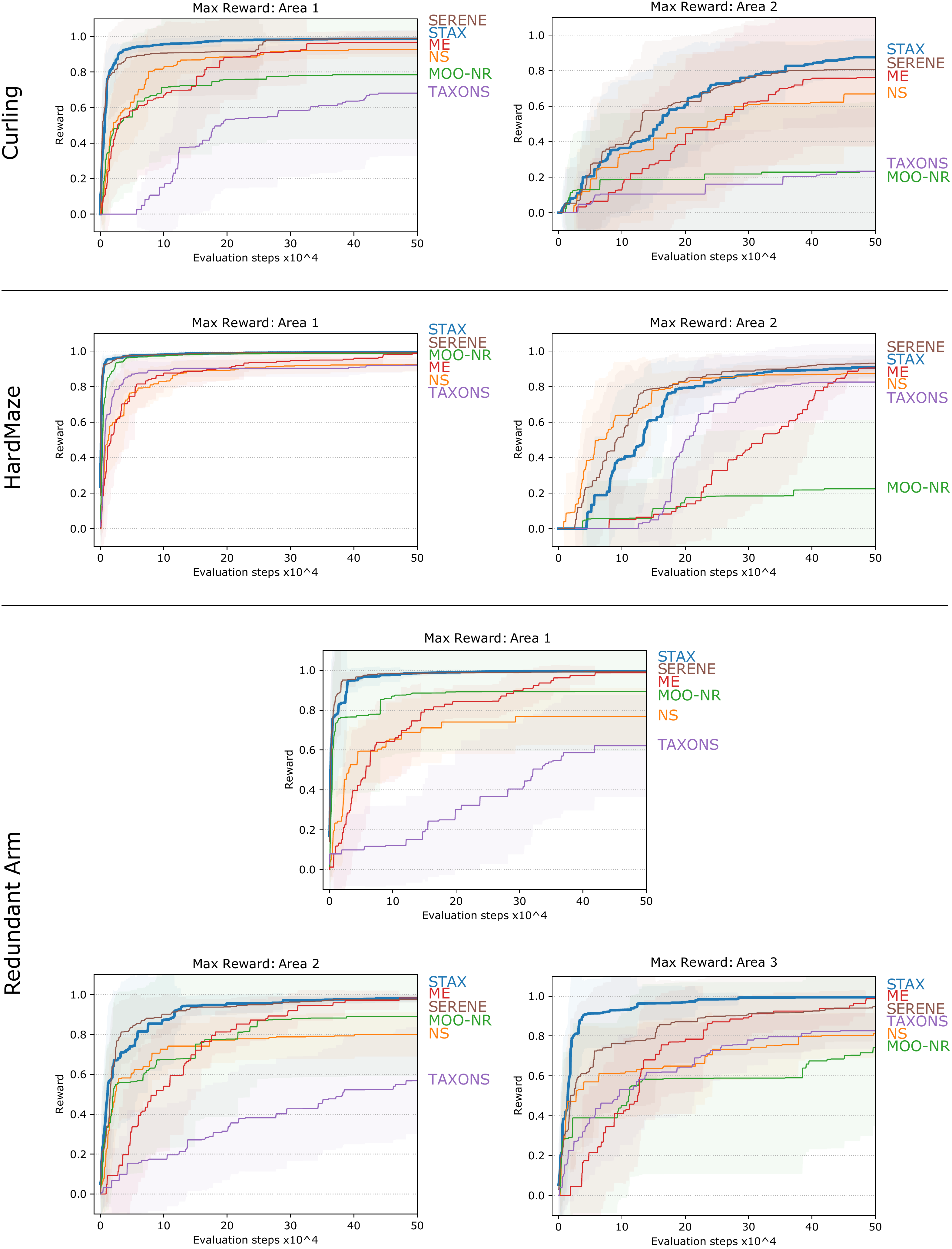}
    \caption[STAX reward results]{Average maximum reward reached in all the reward areas by \gls{name} against the different baselines. The shaded areas represent one
standard deviation.}
    \label{fig:stax_cvg_reward}
\end{figure}
Fig. \ref{fig:stax_cvg_reward} shows the maximum reward achieved by the algorithms in all the reward areas.
Using emitters to exploit the reward allows \gls{name} to reach almost the maximum reward in very few evaluations. 
These performances are very similar to the ones obtained by \gls{serene}, thanks to the fact that the reward exploitation performed by the emitters does not rely on any behavior descriptor.
Among the other baselines performing reward improvement, the best performing one is \gls{me}, capable of reaching high values on all reward areas, but with much slower pace than \gls{name}.
This is not the case for the multi-objective approach, which suffers from the same problems highlighted in Chapter \ref{chap:serene}, maximizing only the easiest to reach reward area.
On the contrary, while \gls{ns} and \gls{taxons} can perform good exploration, they cannot reach high reward levels very quickly, with \gls{taxons} being consistently worse in this regard.
This is even more noticeable on the redundant arm environment, where even if \gls{taxons} can reach similar coverage levels than \gls{name}, the absence of any reward improving mechanism leads to very low performances on all reward areas.

\subsection{Final archives distribution}
\begin{figure}[p]
    \centering
    \includegraphics[height=\dimexpr
  \textheight-5\baselineskip-\parskip-.1em-
  \abovecaptionskip-\belowcaptionskip\relax]{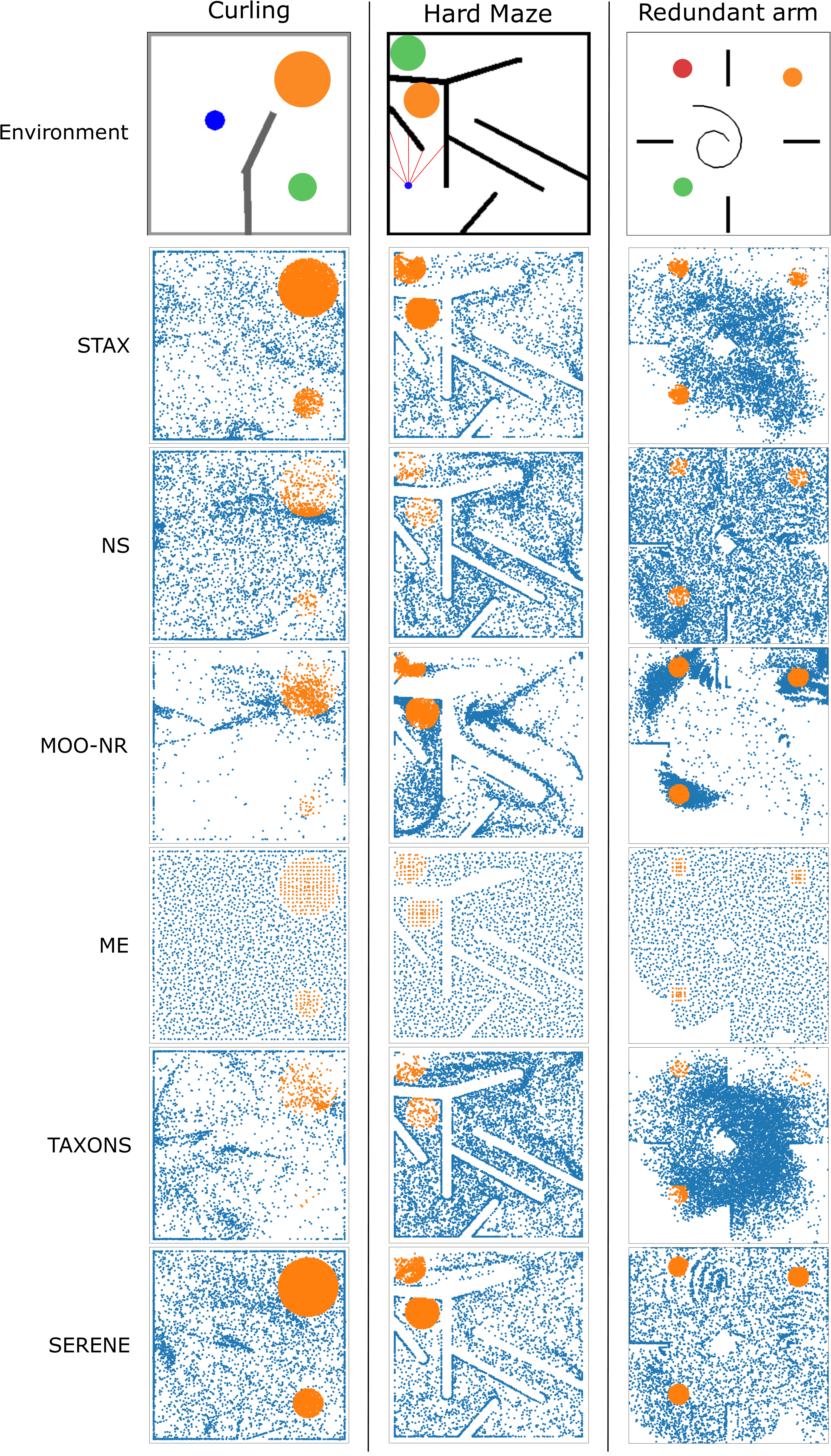}
    \caption[STAX final archives]{Distribution of the behavior descriptors of the archived policies. On each column are shown the results for an environment, while on each row is shown the distribution for each experiment. The archive plotted are from the runs achieving highest coverage. In blue are the policies with no reward, in orange the policies with a reward. For \gls{name} and \gls{serene} in blue are the policies in the novelty archive and in orange the policies in the reward archive.}
    \label{fig:stax_final_arch}
\end{figure}

Fig. \ref{fig:stax_final_arch} shows the final distribution of the behaviors representations for the policies in the final archives.
Each point represents a different policy.
In blue are shown the policies present in the novelty archive $\mathcal{A}_{\text{Nov}}$, while in orange are the policies in the reward archive $\mathcal{A}_{\text{Rew}}$.
For the baselines not using the double archives structure, the blue points represent the policies that did not receive any reward, thus considered \emph{exploratory}, while the orange points represent the rewarding policies.

The coverage of the reward areas for \gls{name} and \gls{serene} is similar, both approaches using emitters to exploit the rewards.
At the same time, \gls{name} tends to cover the search space less densely than \gls{serene} and \gls{ns}, due to not knowing the ground-truth \gls{bs} at search time.
The coverage of the space for \gls{name} more closely resembles the one obtained by \gls{taxons}, the other method learning the \gls{bs} representation at search time, with the exception of the reward areas, better covered by \gls{name}.
Similar coverage of the reward areas is obtained by \gls{moo}-NR, but as said in Chapter \ref{chap:serene}, this comes at the cost of exploration of the rest of the search space.
\gls{me} also obtains a uniform distribution over the whole space, thanks to its discretization.

\subsection{Exploration ablation studies}
\label{sec:stax_ablation}
This section studies what are the contributing factors to the exploration results obtained by \gls{name}.
The study focuses on two aspects of the algorithm: the multi-objective approach for policy selection and the multiple observations used to generate the behavior descriptor of a policy.
Four ablated variants of \gls{name} are considered:
\begin{itemize}
\item \textbf{\gls{name}\_multi}: it is the vanilla version of \gls{name}. It uses both the multi-objective policy selection between novelty and surprise and the 5 observations sampled along the policy trajectory to generate the behavior descriptor;
\item \textbf{\gls{name}\_single}: this variant still uses the multi-objective policy selection strategy, but the behavior descriptor is calculated only from the last observation;
\item \textbf{\gls{name}-ALT\_multi}: this variant uses the same strategy used by \gls{taxons} to select between novelty and surprise, sampling either one of the two at each generation. The behavior descriptor is generated by using 5 observations sampled at regular intervals along the trajectory;
\item \textbf{\gls{name}-ALT\_single}: as the previous variant, here the \gls{taxons} policy selection strategy is used. Moreover, the behavior descriptor is generated by only the last observation of the trajectory.
\end{itemize}
Both the coverage metric, calculated as in previous sections, and the maximum reward reached by each variant over each reward areas are analyzed.
It can seem counterintuitive to analyze the maximum reward obtained by algorithms separating exploration from exploitation when performing an ablation study on the factors fostering exploration.
The importance of this analysis lies in the fact that \gls{name} and its variants select emitters also according to the novelty of the original policy.
This means that the way the novelty is calculated has an influence - albeit minimal - on the reward exploitation.
Moreover, being in a setting of sparse rewards, the way exploration is performed has a direct influence on the speed each reward area is discovered and optimized.

\begin{figure}[!h]
    \centering
    \includegraphics[width=0.7\textwidth]{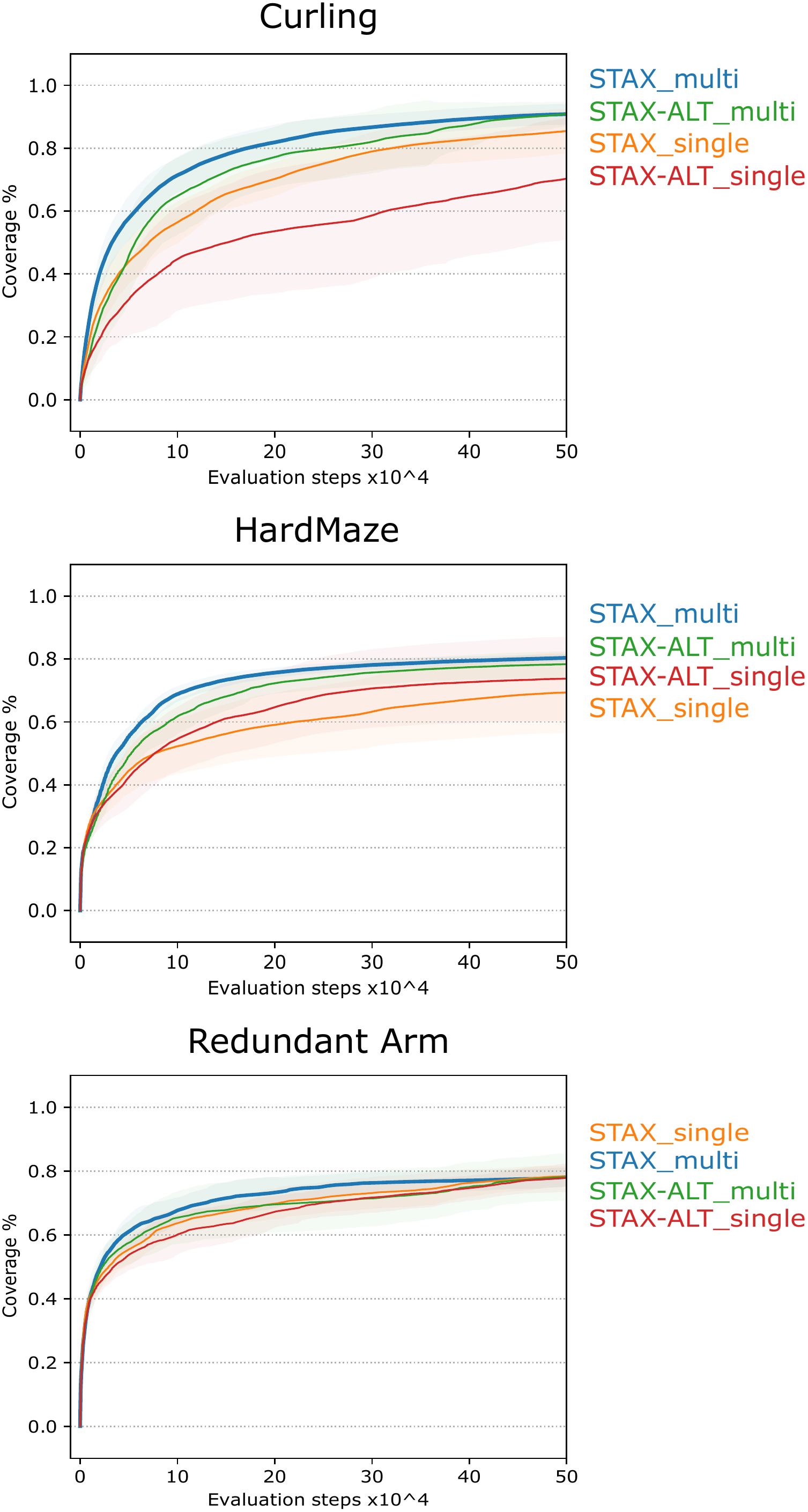}
    \caption[STAX ablation experiments - coverage results]{Average coverage with respect to the given evaluation budget reached by \gls{name} against the ablated versions of the algorithm. The shaded areas represent one standard deviation.}
    \label{fig:stax_ablation_cvg}
\end{figure}

The results on the average coverage reached by the algorithms are shown in Fig. \ref{fig:stax_ablation_cvg}.
From it, it is possible to see that the variants using multiple observations of the trajectory tend to perform consistently better on all environments.
This is also the case for the Redundant arm environment, in which while the final coverage of the algorithms is equivalent, the two variants using multiple observations tend to reach higher levels quicker.
A possible explanation for this is due to the \glspl{ae} of these variants being trained on 5 times more data than the ones of the variants using a single observation. 

\begin{figure}[!h]
    \centering
    \includegraphics[width=\textwidth]{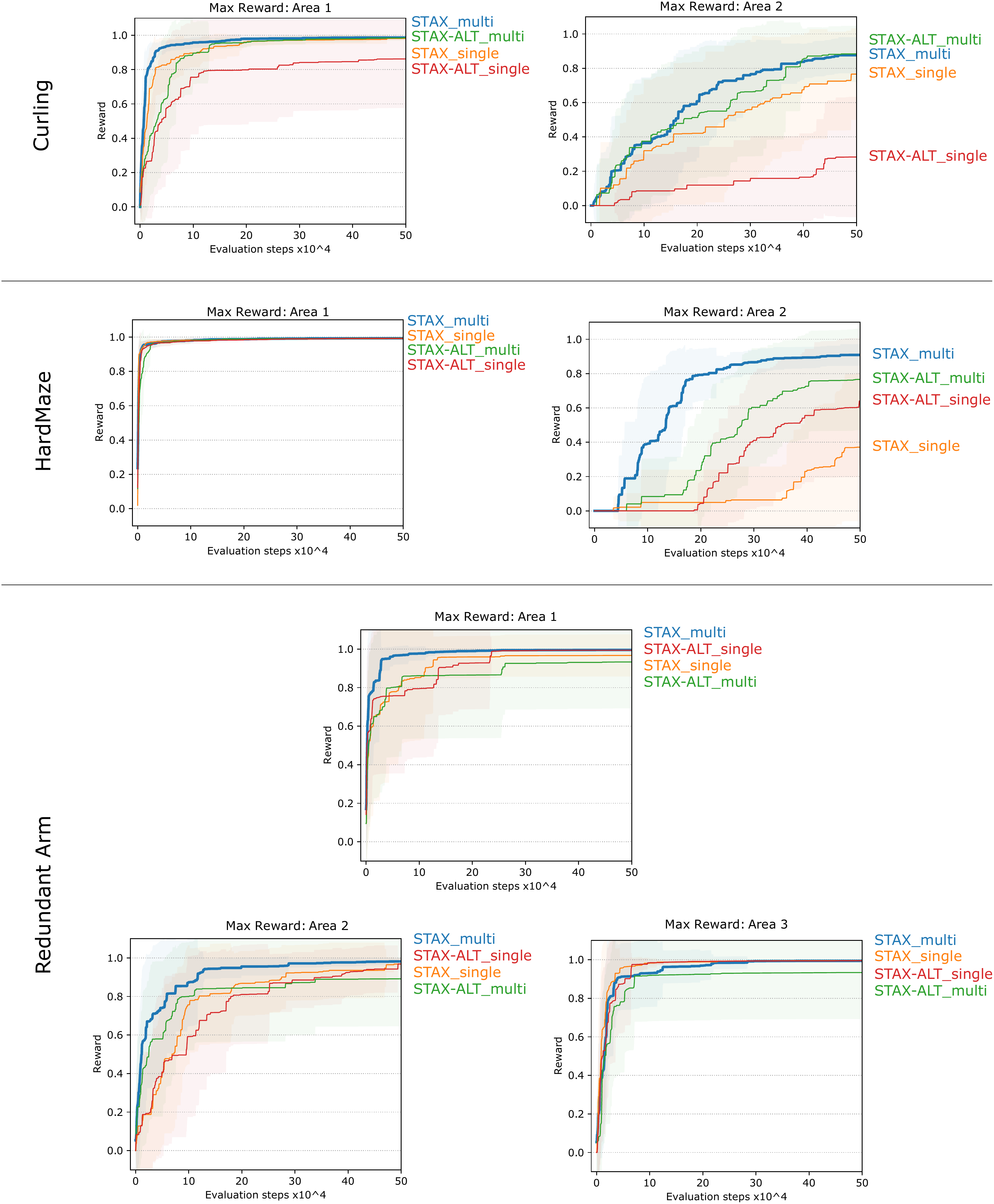}
    \caption[STAX ablation experiments - reward results]{Average maximum reward reached in all the reward areas by \gls{name} against the ablated versions of the algorithm. The shaded areas represent one standard deviation.}
    \label{fig:stax_ablation_rew}
\end{figure}

The improved performance provided by using multiple observations can be seen also when analyzing the maximum reward reached in the environments, as shown in Fig. \ref{fig:stax_ablation_rew}.
In each of the reward areas of all environments, \gls{name}\_multi reaches the highest performances in the quickest fashion.
In general the methods using only the last observation to extract a description of the behavior of a policy performs the worst.
At the same time, the multi-objective policy selection method, used by both \gls{name}\_multi and \gls{name}\_single, has a weaker but non negligible effect on both exploration and the exploitation.
It can be seen in fact that the version using both multiple observations and the multi-objective policy selection strategy performs consistently better than all the other variants.

\subsection{Autoencoder training regime}
\label{sec:stax_learned_bs}
This section analyzes how the way the \gls{bs} is learned through the \gls{ae} influences the search.
In this regard the study focuses on two aspects.
The first one concerns how important it is to learn the representation versus just using a random one.
The second aspect is the importance of continuously training the \gls{ae} during the whole search process.
This training strategy produces a \emph{curriculum effect} over the borders of the explored space due to the training on the last generation of the population and offsprings.
The curriculum effect is also given by training the \gls{ae} over the archives.
However, during the first iterations of the search, when the archives are still small, the \gls{ae} will mainly be trained on the data coming from the last population and offsprings.
This means that at the beginning of the search the biggest contribution to the "memory" of already explored areas for the \gls{ae} comes from the continuous training of the \gls{ae}.

\begin{figure}[!h]
    \centering
    \includegraphics[width=.7\textwidth]{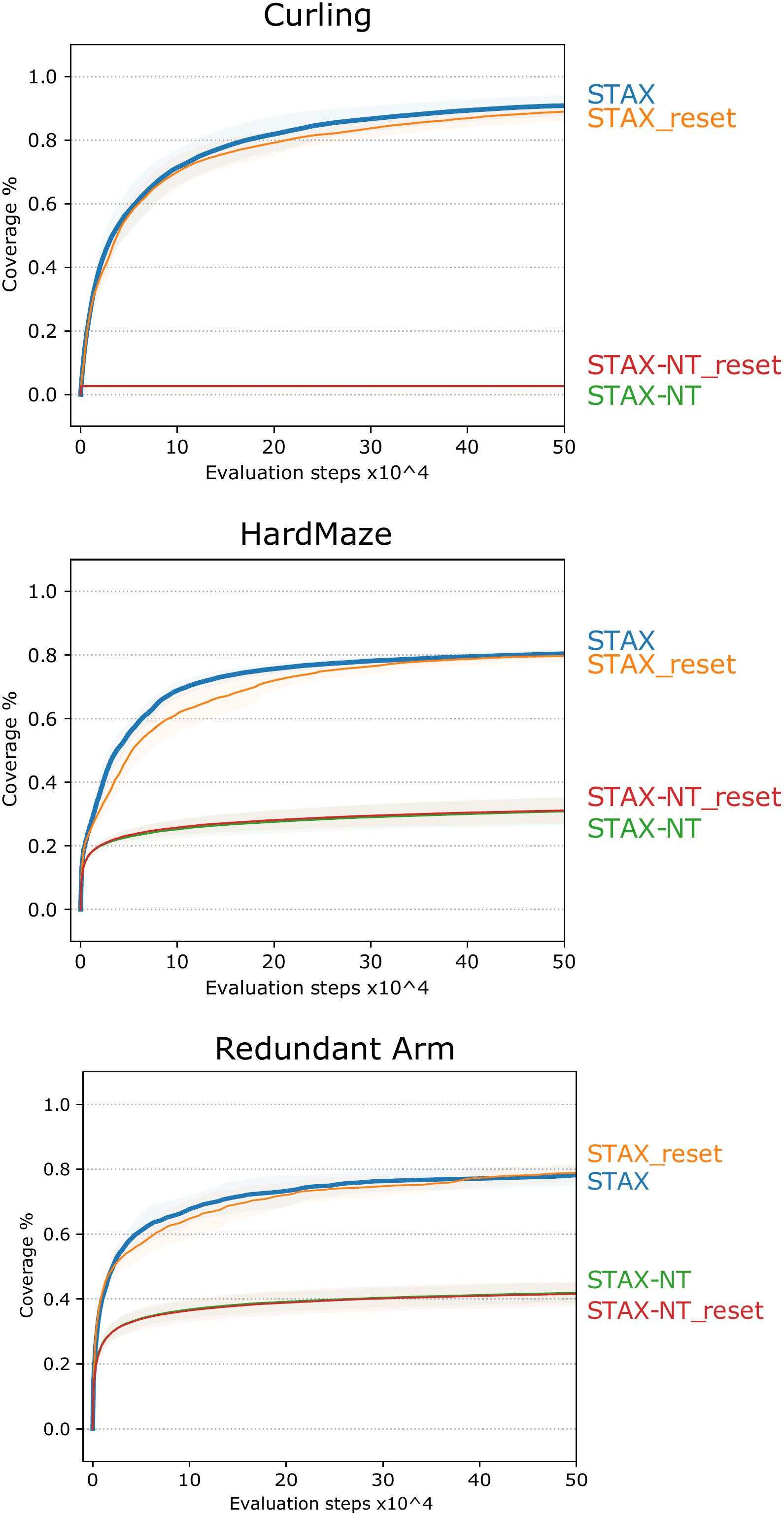}
    \caption[STAX learned BS experiments - coverage results]{Average coverage with respect to the given evaluation budget reached by \gls{name} against the other versions of the algorithm. The shaded areas represent one standard deviation.}
    \label{fig:stax_reset_cvg}
\end{figure}

To analyze these two aspects,\gls{name} is compared against 3 variants:
\begin{itemize}
    \item \textbf{\gls{name}-NT}: in which the search is driven through an \gls{ae} whose weights are randomly sampled at the beginning of the search and not modified anymore;
    \item \textbf{\gls{name}-NT\_reset}: in which the search is driven through an \gls{ae} whose weights are randomly sampled every $TI$ exploration steps. This means that every time the vanilla version of \gls{name} would train the \gls{ae}, this version randomly samples new weights for the \gls{ae};
    \item \textbf{\gls{name}\_reset}: in which the weights of the \gls{ae} are randomly resampled before each training episode.
    This effectively removes any memory from previous iterations from the \gls{ae}.
\end{itemize}
Thanks to the first two variants, it is possible to analyse if a random but constant representation is better than a continuously changing random representation to drive the search.
The last variant allows to study the importance of the curriculum effect given by the continuous training of the \gls{ae}.
Note that the only change among all these version of \gls{name} is the \gls{ae} training regime. 
The behavior descriptor is still generated by stacking the representations extracted from 5 frames sampled along the trajectory, as done in Sec. \ref{sec:stax_exploration}.
The coverage results for the 3 tested environments are shown in Fig. \ref{fig:stax_reset_cvg}. 
Not surprisingly, the results show that training the \gls{ae} rather than using a randomly generated one really pushes exploration. 
This means that the random representations are not enough to discover all the areas of the ground truth \gls{bs}, even if said representations change during the search, as is the case for the \gls{name}-NT\_reset variant.
At the same time, the results show how the continuous training of the \gls{ae} does not have a big effect on the coverage in any of the environments.
This means that the archive can provide enough of a curriculum when learning a representation of the \gls{bs}.
However, \gls{name} has a much smaller execution wall time compared to \gls{name}\_reset, not having to retrain the \gls{ae} from scratch every time.

\begin{figure}[p]
    \centering
    \includegraphics[width=\textwidth]{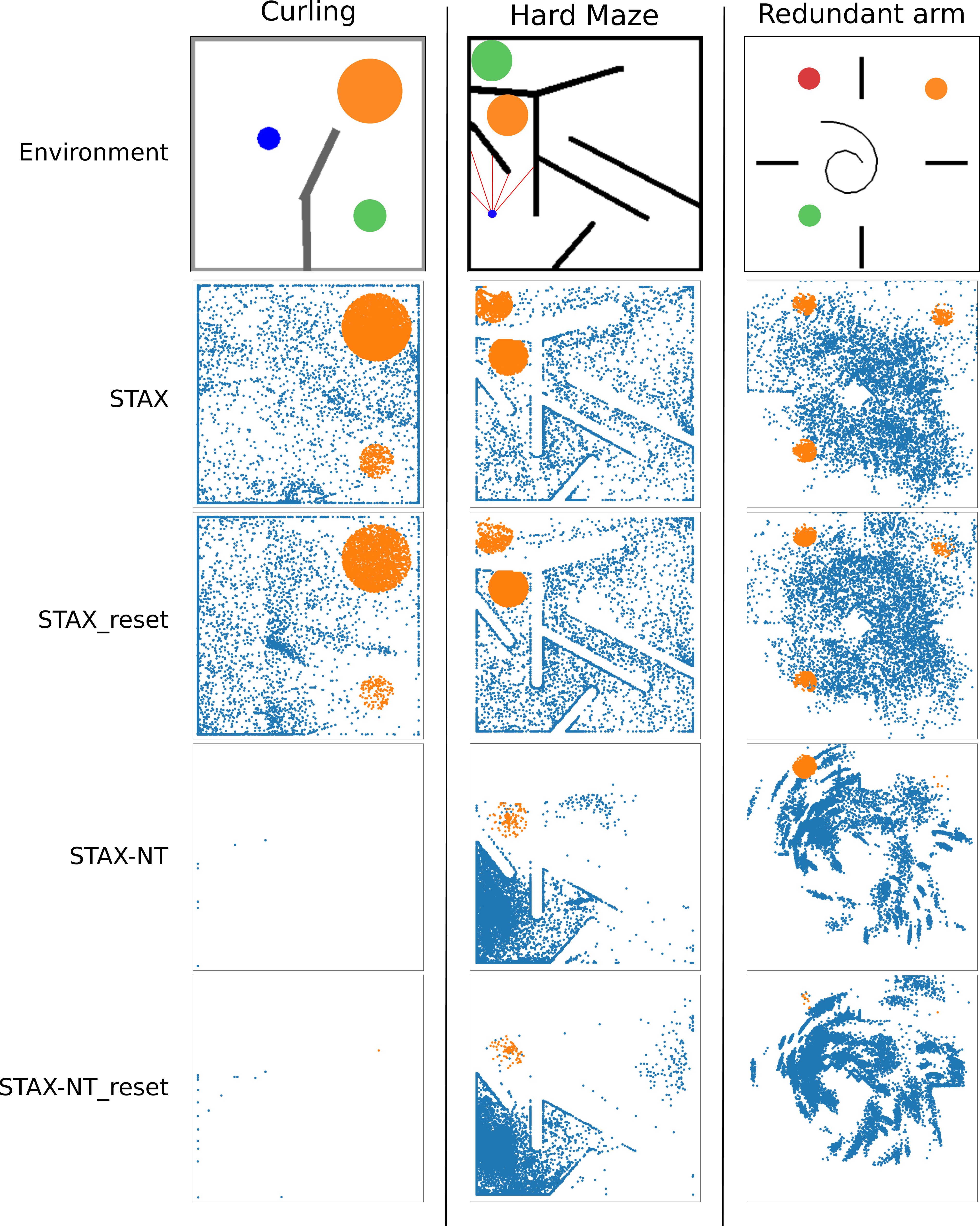}
    \caption[STAX learned BS experiments - final archives]{Distribution of the behavior descriptors of the archived policies. On each column are shown the results for an environment, while on each row is shown the distribution for each experiment. The archive plotted are from the runs achieving highest coverage. In blue are the policies in the novelty archive and in orange the policies in the reward archive.}
    \label{fig:stax_reset_final_arch}
\end{figure}

Fig. \ref{fig:stax_reset_final_arch} shows the final distribution of the archived policies behavior descriptors.
The results show how the variants in which the \gls{bs} representation is not learned really struggle to explore a big part of the space.
This effect is extreme in the Curling environment in which, in order to obtain good exploration it is not enough to randomly move the arm, but it is necessary to properly hit the ball.
In the HardMaze and the Redundant Arm environments the non trained versions can explore the easier to reach areas of the space, but not farther.

These experiments clearly show that each environment has different dynamics when it comes to exploration.
This strengthens our assumption that hand-designing a \gls{bs} in order to properly explore can be difficult and require adaptations to each single situation.
For this reason, it is important to design algorithms like \gls{name} that can learn said \gls{bs} online while starting with minimal prior information.
These algorithms should adapt to all environment dynamics by taking advantage as much as possible of the data generated during the search.

\subsection{Learned behavior space}
\begin{figure}[!h]
    \centering
    \includegraphics[width=\textwidth]{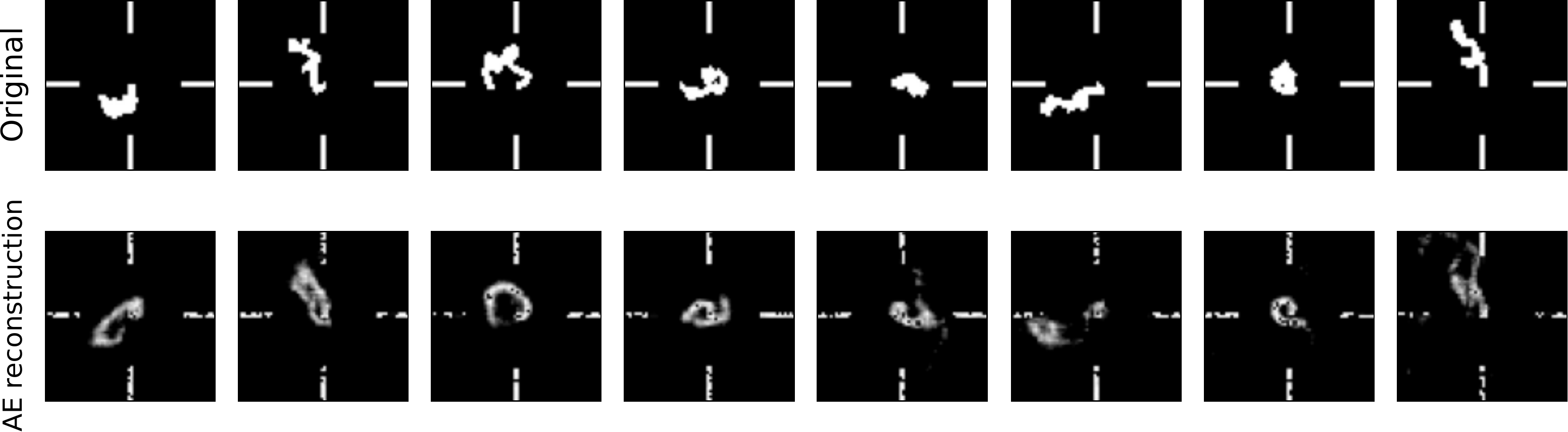}
    \caption[STAX AE reconstruction]{Reconstruction of the \gls{ae} trained during the search performed by \gls{name}. The first row shows the original $64 \times 64 \times 3$ images. The second row shows the reconstructions of the images produced by the \gls{ae}.}
    \label{fig:stax_ae_reconstruction}
\end{figure}

This section studies how well the trained \gls{ae} can represent the \gls{bs} and how close these learned representations are to the ground truth one.
Given that the results are comparable among all the environments, the section will focus mainly on the harder to explore Redundant Arm environment. 
Fig. \ref{fig:stax_ae_reconstruction} shows how well \gls{name}'s learned \gls{ae} can reconstruct the observations collected during the evaluation of the policies.
The first row shows the $64 \times 64 \times 3$ final observation of the trajectories of a set of policies sampled form the final archives.
In the second row are shown the reconstructions produced by the trained \gls{ae}.
While from this reconstruction it is possible to understand the position of the arm, the image is not perfect.
Nonetheless, this level of reconstruction accuracy seems to be enough to push for good exploration in the environment, as seen in Sec. \ref{sec:stax_exploration}.

\begin{figure}[p]
    \centering
    \includegraphics[height=\dimexpr
  \textheight-5\baselineskip-\parskip-.2em-
  \abovecaptionskip-\belowcaptionskip\relax]{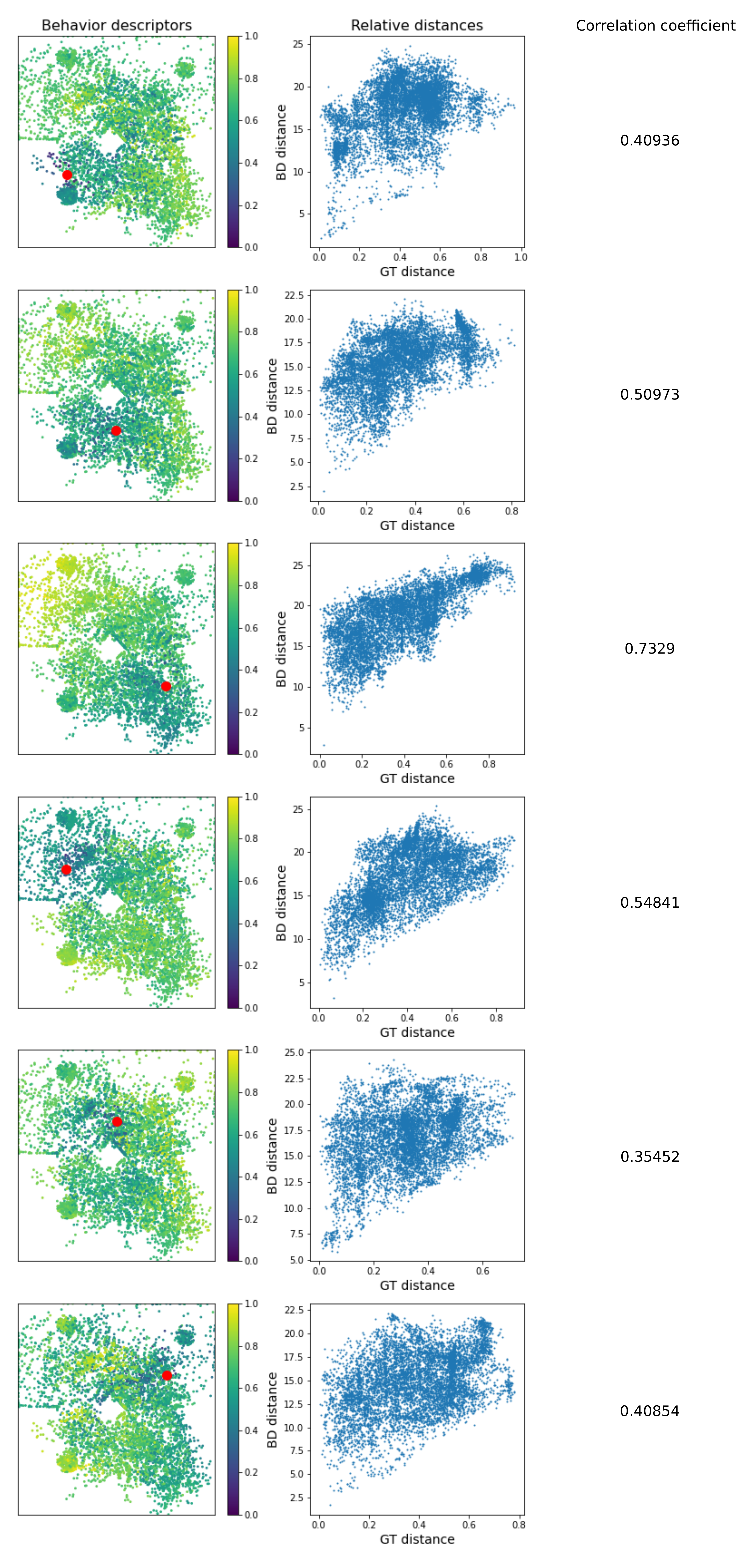}
    \caption[STAX AE behavior descriptors]{Representation of the proximity between the learned \gls{bs} and ground truth one. The first column represents the points in the ground truth \gls{bs} color coded according to the distance in the learned \gls{bs}. The red circle is the sampled descriptor for which the distance from the other descriptors is calculated. 
    The second column represent the distances in the learned space with respect to the ground truth \gls{bs}.
    Finally, on the third column is shown the correlation coefficient calculated between said distances in the two spaces.}
    \label{fig:stax_ae_distance}
\end{figure}

From this, the question on how close the learned \gls{bs} representation is to the ground truth one arises naturally.
This has been studied by sampling 6 policies form the archives at 6 different positions in the ground truth \gls{bs}.
Then, the distance between the learned behavior descriptors of the sampled policies and the ones of the other policies in the archives is calculated.
The results can be seen in Fig. \ref{fig:stax_ae_distance}.
Each row shows the results with respect to one of the 6 sampled policies, whose ground truth descriptor is highlighted in red in the plot in the first column.

The first column contains the policies' ground-truth descriptors plotted by color coding them according to their distance in the learned \gls{bs}.
Closer points in this space are represented in darker colors, farther ones in lighter colors.
The second column represents the distances in the learned \gls{bs} with respect to the distances in the ground-truth space, while on the third column it is shown the Pearson correlation coefficient \cite{pearson1895vii} between these distances.

From the figure it is possible to see that closer points in the learned \gls{bs} are closer in the ground-truth space, and the farther this points are in the GT space, the farther they become in the learned space.
This is also shown by analyzing the correlation between the distances in the two spaces. 
The correlation coefficient shows that there is moderate to high correlation between these distances, confirming that closer points in the ground-truth space tend to be closer in the learned space and vice-versa.
This means that the \gls{ae} has learned a meaningful representation that can be used to push for exploration by calculating distances in the learned space, proving the efficacy of this approach. 

\newgeometry{top=0.3cm,bottom=0.1cm, left=1cm, right=1cm}
\restoregeometry
\section{Conclusion}
\label{sec:stax_conclusion}
This chapter introduced \gls{name}, a method that reduces the amount of prior information about a task to minimal levels.
It does so by combining the representation learning ability of \gls{taxons} when dealing with unknown \gls{bs} and the capacity to focus on interesting areas of the search space of \gls{serene} through emitters.
In addition to what \gls{taxons} does when learning the \gls{bs}, \gls{name} uses multiple observations sampled along the trajectory generated by the policies to extract their behavior descriptor. 
This allows to overcome the limitation of dealing with environments where the final observation needs to be descriptive enough to distinguish between the policies.
Moreover, by using a multi-objective approach to combine the two metrics of novelty and surprise, \gls{name} can perform better exploration compared to \gls{taxons}.
As discussed in Sec. \ref{stax:exploitation}, performing exploitation through emitters that can deal with multiple disjoint reward areas when learning a \gls{bs} representation can prove extremely useful.
This is due to the fact that there is no guarantee that the learned \gls{bs} will represent all the rewards in a single connected area.

\gls{name} has been tested on three different sparse rewards environments, reaching high performances in all of them, both from the point of view of exploration and exploitation of the reward.
These results are comparable to the ones obtained by \gls{serene} notwithstanding \gls{name} being provided much less prior information about the task to solve.
Moreover, thanks to the representation learning capability provided by \gls{taxons}, \gls{name} is capable of overcoming the main limitation of \gls{serene} discussed in Sec. \ref{sec:ser_conclusion}.

To properly study how the aspects of policy selection and \gls{bs} learning of \gls{name} influence the exploration process, and the discovery and exploitation of rewards, multiple ablation experiments have been performed.
The results show that the combination of using multiple observations collected during the trajectory and the multi-objective policy selection strategy are important in obtaining good coverage of the ground-truth search space.
Moreover, the continuous training of the \gls{ae} during the whole search is shown to provide an useful curriculum effect, in addition to the one provided by training on the data from the archives.
Finally, the Chapter shows how the learned \gls{bs} has a similar structure to the ground truth \gls{bs}, allowing the algorithm to perform good exploration in both.

Notwithstanding the multiple shortcomings of the original \gls{ns} algorithm addressed by \gls{name}, there are still many aspects of this method that can be further studied and improved.
As for \gls{serene}, \gls{name} uses a simple scheduler to alternate between the exploration and the exploitation processes.
Applying more complex and adaptive approaches to perform the switch between the two processes can be an interesting line of work in improving the method even more.
Another stimulating direction of research is the one initiated by Cully in \cite{cully2020multi}, where multiple kind of emitters are combined through a multi-armed bandit approach.

\clearpage\null\thispagestyle{empty}

\chapter{Discussion}
{\hypersetup{linkcolor=black}\minitoc}
\label{chap:discussion}
This manuscript introduced methods to augment the capacity of \gls{ns}, and \gls{qd} algorithms in general, when dealing with the problem of sparse rewards.
The main approach considered is to focus on exploration while reducing the amount of prior information needed by the algorithm.
The contributions can be grouped in two main categories:
\begin{itemize}
    \item Autonomously learning the behavior space while performing the search;
    \item Giving the methods the ability to focus on the most interesting areas of the search space without sacrificing global exploration.
\end{itemize}
This chapter discusses these contributions in light of the introduced methods.
Their limitations are also analysed, with possible solutions and extensions, and the research directions they open.

\section{Learning the behavior space}
The biggest contribution towards autonomously learning the search space is the one provided by \gls{taxons}, discussed in Chapter \ref{chap:taxons}.
By learning the space in which the novelty of the policies is evaluated, this method greatly reduces the amount of engineering effort needed at design time.
At the same time, it extends the range of domains in which \gls{ns} and \gls{qd} algorithms can be applied. 
For many problems it is not easy to properly hand-design a \gls{bs} that can foster good exploration.
For example, while trying to discover policies capable of changing the position of an object, this position needs to be tracked.
This usually implies having multiple cameras capable of triangulating the location of said object, or a dedicated software.
Learning the behavior space through high-dimensional RGB images removes this need, at the cost of not being able to control the representation of the \gls{bs} anymore.
Sec. \ref{sec:tax_results} showed how well \gls{taxons} can perform exploration by observing the environment through said RGB images.

Most of the prior information provided to \gls{taxons} comes from the selection of the final observation of the trajectory to generate the behavior descriptor.
This requires the final state of the whole system to contain information about the whole behavior of the policy.
A study in order to address this limitation was conducted in Chapter \ref{chap:signatures}.
The focus of the study was the signature transform \cite{fermanian2021embedding}, a mathematical operator capable of encoding a whole stream of data into a single vector.
This could have been combined with \gls{taxons} through a two step process.
First, the \gls{ae} would represent each high-dimensional RGB image generated during the trajectory in its learned feature space, generating a trajectory of encoded representations.
Then the signature could be used on this trajectory to encode it even more, in a single, fixed length vector.
Before continuing on this path, the signature was tested as a way to encode a trajectory of the simple ground-truth representations generated during the policy evaluation.
The results reported in Sec. \ref{sec:sign_results} show that even if signatures can foster good coverage along the whole trajectory of states, simply sampling few observations from the trajectory allows to obtain similarly good results.
The probable reason is the exponential increase in dimensionality of the signature with respect to its order and to the size of the datapoints.
It has been shown that in high-dimensional spaces, distance metrics lose most of its relevance \cite{Aggarwal2002distance}.
This can hinder the exploration power of \gls{ns} based algorithms that rely on distances in the \gls{bs} to push for diversity.
For this reason, when \gls{name} was introduced in Chapter \ref{chap:stax}, the method relied on a subsampling and stacking of multiple observations along the trajectory to generate the \gls{bs}.
Even if relatively simple, this approach proved effective enough in greatly reducing the amount of prior information with minimal overhead.
Sec. \ref{sec:stax_results} shows that thanks to this approach, \gls{name} can autonomously learn a behavior descriptor from the whole trajectory, fostering good exploration in the unknown ground-truth \gls{bs}.

However, autonomously learning a representation of the \gls{bs} does not come without costs.
The training of the \gls{ae} adds significant overhead in both execution time and computation power required to execute the algorithm. 
Using an \gls{ae}, and even more so one based on \gls{cnn}, requires the presence of a dedicated GPU to speed up training and inference, requirement not usually necessary for \gls{qd} algorithms with an hand-crafted \gls{bs}.
Moreover, after every training episode of the \gls{ae}, the descriptors of the policies in all archives and current populations need to be updated with the features extracted by the newly trained \gls{ae}.
This makes the methods scale badly with the number of generations, given the continuously increasing number of policies in the archives that need to be updated.
The problem is much more important for \gls{taxons} than for \gls{name}: the former trains the \gls{ae} at regular intervals while the latter performs the training episodes less frequently with the passing of generations.

At the same time, these are not the biggest shortcomings.
As said, by autonomously learning the \gls{bs} the designer loses control on what features will be present in the descriptors.
The \gls{ae} will try to learn a representation of everything present in the RGB observations.
This includes distractors.

\subsection{Distractors}
\label{sec:distractors}
Stone et al. define distractors as ``variations in the input that are irrelevant for the task''  \cite{stone2021distracting}.
They can appear in multiple forms.
It is possible to have distractors controlled by the algorithm but not directly related to the task, or distractors whose presence or actions are independent from the learning algorithm.
An example of the first kind of distractors is the arm in the Curling environment used in Chapter \ref{chap:stax} to test \gls{name}.
While directly controlled by the policies, the position of the arm at any given moment is not important with respect to the task of having the ball in different locations on the table.
At the same time, being represented in the RGB observations, it can greatly influence the novelty of the corresponding behavior descriptor extracted from these observations.
In the case of the Curling experiments, said distractor did not prove to be a limitation for \gls{name}, as it can be seen from the results reported in Chapter \ref{chap:stax}.
However, this is more likely due to the simplicity of the environment rather than the power of the algorithm itself.
Nonetheless, learning to represent this kind of distractors can be beneficial to the generalization capabilities of the algorithm.
The final archive, selected according to diversity in a space in which the distractor is encoded, can then be used to address tasks for which what before was a distractor now is fundamental.

The second kind of distractors, the ones on which the algorithm has no control, can be more difficult to deal with.
These can be of different types, going from an object randomly moving in the environment to the background changing color. 
These aspects of the environment can be problematic for any algorithm using intrinsic rewards to foster exploration.
This is due to the great source of randomness provided by these elements, rendering already visited states novel enough for the algorithm to keep revisiting them.
The problem has been formalized as a thought experiment and given the name of \emph{Noisy TV problem} \cite{burda2018exploration}.
The experiment considers an agent in an environment trying to maximize the novelty of its observed states. 
The environment contains a TV in which random images are continuously displayed.
This TV would be able to attract the focus of the agent forever, thanks to the continuous novelty provided by the unpredictability of these random images.
In these situations, the agent would need an external signal in order to understand what is interesting and what is not.
This is especially true when the representation is learned in an unsupervised fashion as done for \gls{taxons} and \gls{name}.
The importance of the problem of distractors and the increased attention to it are highlighted by the amount of recent publications proposing methods aiming at dealing with distractors \cite{laversanne2018curiosity, burda2018exploration, agarwal2021contrastive, bodnar2021metric, fu2021learning, hadeep, schott2021improving}.
Moreover, Stone et al.\cite{stone2021distracting} extended the classical DeepMind control suite \cite{tassa2018deepmind} with visual distractors in order to benchmark existing algorithms against this kind of problem.

While the methods introduced in this manuscript cannot deal with them, studying a way to address distractors in the framework of \gls{qd} algorithms can be an exciting line of future work.
This could also lead to a much wider range of application of this family of algorithms.

\subsection{Disentangled representations}
An interesting approach in dealing with distractors is to learn a set of disentangled representations of the \gls{bs} \cite{thomas2017independently, pandey2020disentangled}.
Disentangled representation learning allows to separate, or \emph{disentangle}, each independent factor of variation in the input vector as a small set of variables in the learned feature space.
In other terms, it allows to separate the factor of variations in the inputs on different axis in the feature space.
Being able to do so would allow the designer to easily select on which aspects of the learned \gls{bs} the algorithm should focus, thus also being able to steer the search in the desired direction.

There are many ways in which this problem can be tackled, all falling under the umbrella of state representation learning.
This field is rapidly growing and attracting the interest of many researchers \cite{lesort2018state}.
At the same time, there are many aspects to consider.
It was shown that through pure unsupervised learning it is impossible to obtain a completely disentangled representation of the factors of variation \cite{locatello2019challenging}.
This means that some other factors need to be taken in account while trying to learn this type of representations.
In this regard, an interesting approach is HOLMES \cite{etcheverry2020hierarchically}, in which the authors learn a hierarchy of representations to foster exploration.
While the approach is designed for morphogenetic systems, it should be possible to apply it to settings similar to the ones addressed in this manuscript.
A related problem is that the autonomous learning of a \emph{single representation space}, in which exploration is performed, can be limiting on more complex problems, where multiple aspects may need to be explored.
The collection of policies can then be biased toward a single one of these aspects.
MC-AURORA \cite{cazenille2021ensemble} addresses this problem by generating multiple collections.
For each of these collections, the policies are selected according to a different learned representation.
This helps in having a more diverse final collection of policies, covering many of the aspects of a same problem.
Generating multiple archives thanks to multiple learned representations can be an interesting direction of research also for methods like \gls{taxons} and \gls{name}.

\section{Focusing on the interesting parts of the search space}
\gls{ns}'s limitation of not focusing on any interesting part of the search space was addressed in this manuscript in Chapters \ref{chap:serene} and \ref{chap:stax}.
The main contribution is the one introduced by \gls{serene}, which can exploit any possible reward discovered during the search by using emitters.
This idea has then been extended with \gls{name}, a method in which the exploration is driven not by \gls{ns} but by \gls{taxons}, while still exploiting the reward through emitters.
This way of exploiting the reward areas has proven to be effective in quickly obtaining high rewards on all the disjoint reward areas, as shown in Sec. \ref{sec:ser_discussion} and in Sec. \ref{sec:stax_results}.
An important aspect of these two methods, greatly contributing to such results is the separation of the exploration and exploitation steps in two alternating processes.
The switch between the two is performed thanks to a scheduler assigning computation budget to either one of them.
The scheduler used in this manuscript is fairly simple: if any reward is discovered, it assigns the budget to the two processes in an alternating fashion, otherwise it completely focuses on exploration.
Sec. \ref{sec:ser_budget} analyzed how this simple strategy is effective in balancing the budget between exploration and exploitation of the different reward areas.
Nonetheless, there is still room for improvement.
Once all the discovered rewarding areas have been densely explored and optimized upon, it would be best to focus more on the exploration in order to discover more reward areas.
At the same time, after much of the search space has been explored, there is no need anymore to spend half the budget on exploration, but rather focusing on improving on the rewards could prove to be more beneficial.
For this reason, having a scheduling strategy capable of adapting itself to the dynamics of the search process could help in optimizing the efficiency of the search even more.
This issue is what is usually referred as \emph{exploration-exploitation trade-off} \cite{sutton2018reinforcement}.
A possible way to optimize the scheduler strategy in these situations could be to use a \gls{mab} approach \cite{slivkins2019introduction}, capable of selecting the best option among the two at each given moment.
At the same time, to apply a \gls{mab} algorithm there is the need to define a metric to optimize between exploration and exploitation.
\glspl{mab} can in fact be described as stateless \gls{rl} algorithms.
This means that they learn a policy by optimizing a reward function.
In order to define a \gls{mab} reward function for an emitter based approach like \gls{serene} or \gls{name}, there is the need for a way to measure the exploration progress and the reward exploitation in comparable ways.
Even if this looks like an easy to address issue, it is not the case due to the fundamentally different nature of the two aspects.

Another possible direction of development for \gls{serene} based algorithms, regards the type of emitters used.
In this manuscript only a single type of elitist-based algorithm has been used as emitter, but this needs not to be the case.
Different kind of emitters were already introduced in Fontaine's work \cite{fontaine2020covariance}, even if a single type was used at the time.
Cully improved on this by using multiple kind of emitters concurrently, selecting the best one at each given time through a \gls{mab} \cite{cully2020multi}.
Moreover, emitters do not have to be necessarily based on \glspl{ea}.
On the contrary, any reward based algorithm can be used as an emitter, \gls{rl} ones included.
Following this path of research could help in improving the efficiency of emitter based algorithms and allow their application to more complex problems.

\section{Noisy environments}
Another limitation of \gls{ns} based algorithms, and more in general all \gls{qd} ones, is the inability to deal with noisy environments.
This is due to the way the behavior representation of each policy is compared against all the others.
In order to have a meaningful comparison, the environment in which the policies act needs to be deterministic, having as only reason for the difference among behaviors the policy itself.
The requirement for deterministic environments greatly hinders the range of problems to which this kind of algorithms can be applied.
Many real world environments are in fact noisy by nature.
For this reason, being able to deal with noisy environments is an important aspect to take into account, even more in comparison with the \gls{rl} literature, for which stochastic environments are common.
Yet, while some work has been done in dealing with noisy fitness functions for standard \glspl{ea} \cite{jin2005evolutionary, rakshit2017noisy}, not so much effort has been done with respect to \gls{qd} algorithms.
The most straightforward approach to the problem is to perform multiple evaluations of the same policy to strengthen its performance estimation when calculating its behavior descriptor \cite{cully2018hierarchical}.
The same idea has also been used in conjunction with applying adaptive sampling to increase the efficiency of the evaluation \cite{justesen2019map}.
Yet, these approaches come with the added cost of testing multiple times the same policies, greatly increasing the computational requirements of the algorithm.

A different strategy is the one introduced with DG-MAP-Elites \cite{flageat2020fast}, in which the usual \gls{me} grid is extended in depth in order to host multiple policies in the same cell.
This allows a better estimate of the performance of a given elite of solutions without the need for multiple evaluations.
Said estimate leads to an higher stability of the archive with respect to noise.
At the same time, while this strategy works for a grid based algorithm as \gls{me}, it might be difficult to adapt to \gls{ns} based algorithms.
In this light, a different way to approach the problem could be to use a multi-objective approach in which policies are selected not only for their novelty but also for their robustness to noise.
While this would still require multiple evaluations for each policy, it could lead to more robust and generalizable solutions.

Nonetheless, the problem is still far from solved, and working in this direction could lead to great improvements in the range of applicability and recognition from other communities of \gls{qd} algorithms.

\clearpage\null\thispagestyle{empty}

\chapter{Conclusion}
\label{chap:conclusion}

Throughout this manuscript, algorithms capable of addressing the issue of sparse rewards with minimal prior information about the task have been introduced and evaluated.
The thesis started by framing the problem of sparse rewards in the \gls{rl} framework, proposing to address it by focusing on \emph{exploration}.
Doing so allows to efficiently analyze the whole search space and discover any obtainable reward, giving us the opportunity to later exploit it.
At the same time, the search space needs to be properly defined in order to perform efficient exploration and discover these rewards. 
Collecting the necessary prior information about the task to solve and designing the search space itself requires a lot of engineering effort.
This limits the range of applicability and the generalization ability of said policy learning algorithms.
For this reason, reducing the amount of prior information needed at design time can prove beneficial.
As a step in this direction, methods that can autonomously learn the search space in an online fashion have been introduced.

An overview of the literature on the subject of sparse rewards in Chapter \ref{chap:related}.
It starts by describing the framework of \gls{rl} and how sparse reward problems can affect the performances of \gls{rl} algorithms.
The best case scenario in which to apply \gls{rl} algorithms is in fact one in which the reward is dense. 
If it is not the case, other strategies need to be considered.
A technique usually employed in these situations is to have the algorithm generate its own rewards. 
However, this can be complicated or lead the search to get stuck on local minima, as for example in the noisy TV problem discussed in Sec. \ref{sec:distractors}.
In light of this, the work focused on \glspl{ea} as an alternative to \gls{rl} for learning policies in situations of sparse rewards.
Being episode-based rather than step-based means that \glspl{ea} can more naturally deal with sparse rewards settings compared to \gls{rl} methods.
This is due to \glspl{ea} needing the reward only at the end of the policy evaluation and not after each single action.
At the same time, there can be situations in which the reward is extremely sparse and the policies do not get any reward.
To deal with this, the \emph{divergent search} family of algorithms has been considered.
These \glspl{ea} mainly focus on exploration.
They do so by looking for a set of policies covering a given search space, called \glsfirst{bs}, as much as possible.
Among all the possible divergent search \glspl{ea}, the work focused mainly on \gls{ns}.
The reason behind this choice lies in the ability of this method to quickly explore the whole \gls{bs}, tending towards an uniform coverage of it.
All of this is done while completely ignoring any reward and without the discretization of the \gls{bs} required by other methods in the same family.

Notwithstanding the great exploration abilities of \gls{ns}, it is still limited by two main shortcomings.
First, the search space needs to be hand-designed, which requires a lot of effort from the designer point of view, while also limiting the range of problems to which this algorithm can be applied.
Second, by completely discarding rewards, \gls{ns} ignores important information that can be useful for solving the task at hand.
This thesis started by addressing the two issues separately.
This has been done with the introduction of two algorithms: \gls{taxons} and \gls{serene}.
Once proven the efficacy of these methods, an approach merging the complementary aspects of \gls{taxons} and \gls{serene} in a single algorithm was introduced: \gls{name}.
\gls{name} can thus perform exploration of the \gls{bs} while exploiting any discovered reward, with minimal prior information.
This allows to address both limitations of \gls{ns} at the same time.
Let us recap how each of the mentioned methods work and what contributions they brought.

\subsubsection{TAXONS}
\gls{taxons}, introduced in Chapter \ref{chap:taxons}, addresses the issue of hand-defining the search space in which \gls{ns} evaluates the novelty of each policy.
It does so by autonomously learning said space during the search process.
This is done by taking advantage of the representation capabilities of an \gls{ae}, while performing the search directly in its learned feature space.
This space is learned online on the data generated by the policies during the search itself.
The results show how, by learning the \gls{bs} directly from high-dimensional RGB observation of the environment, it is possible to perform exploration almost as well as \gls{ns}.
Doing so greatly reduces the amount of prior information needed.
At the same time, \gls{taxons} relies on the assumption that the last image of the trajectory contains enough information to extract the behavior descriptor of a policy.
In this regard, Chapter \ref{chap:signatures} discussed the Signature transform as a possible way to remove this assumption by representing the whole trajectory as a single vector.
The aim was to be able to combine this method with \gls{taxons} in order to calculate the behavior descriptor of a policy from the trajectory of high-dimensional RGB observations.
This approach was compared against the simpler strategy of obtaining the behavior descriptor by just stacking few observations sampled along the trajectory.
Eventually, this simpler strategy proved as effective as the more complex signature transform.
The possible reason behind this is the high dimensionality of the descriptor extracted by the signature.
Euclidean distances on high-dimensional vectors lose most of their relevance, lowering the effectiveness of the novelty metric in pushing for exploration.

\subsubsection{SERENE}
The second limitation of \gls{ns}, the discarding of any information coming from the reward, has been addressed with the introduction of \gls{serene} in Chapter \ref{chap:serene}.
This algorithm performs exploration through \gls{ns}, then once rewards are discovered it deploys emitters to exploit them.
As explained in Sec. \ref{sec:ser_emitters}, emitters are instances of reward-based algorithms performing local search around the policies used to initialize the emitters themselves.
This local search allows to quickly optimize on the reward without interfering with the exploration process.
To this end, \gls{serene} keeps the exploration of the search space and the exploitation of the rewards separated through a two alternating-steps process.
This strategy allows the method to find a good compromise for the exploration-exploitation trade-off.
\gls{serene} has been shown to perform exploration in a fashion similar to \gls{ns}.
At the same time, it can efficiently optimize the reward for each discovered area, overcoming \gls{ns}'s limitation.
The reward optimization has been shown to be more efficient than other \gls{qd} algorithms as \gls{me} or CMA-ME that perform exploration and reward exploitation at the same time.
This proves that the separation of the exploration process from the exploitation can be advantageous in situations of particularly sparse rewards.

\subsubsection{STAX}
As said, \gls{taxons} and \gls{serene} address two different problems of \gls{ns}.
These two approaches can be considered complementary in dealing with \gls{ns} limitations and thus combined.
This has been done in Chapter \ref{chap:stax} with the introduction of the \gls{name} algorithm.
The method augments \gls{serene} with \gls{taxons} by using the latter to perform the search in the exploration step of the former.
This strategy allows \gls{name} to explore an unknown search space without relying on any prior information about the task.
Then, once a reward is discovered, an emitter is initialized and evaluated to exploit said reward.
At the same time, the \gls{taxons}'s requirement for the last observation of the trajectory to be informative enough with respect to the whole policy behavior has been removed in \gls{name}.
This has been done by taking advantage of the lessons learned in Chapter \ref{chap:signatures}.
Multiple high-dimensional observations are sampled along a trajectory and their low-dimensional representations, generated by the \gls{ae}, are stacked to form the behavior descriptor of the policy.
The results show how this approach can foster exploration on levels similar to \gls{ns}, with the minimal amount of prior information given to the algorithm.
All of this without any loss in performance on the exploitation of the rewards.
\gls{name} is in fact able to discover and quickly exploit all reward areas present in the environment.

The advantages and contributions introduced by the methods developed during this thesis have been discussed in Chapter \ref{chap:discussion}.
Starting from \gls{ns} and developing first \gls{taxons} then \gls{serene}, highlighted a path towards the introduction of a family of algorithms that can seamlessly deal with sparse reward environments.
All the while requiring minimal knowledge about the task.
This could potentially lead to the introduction of algorithms capable of great generalization, reducing to the minimum the intervention of the engineer when moving from one setting to the other.
The introduction of \gls{name} represents a promising step in this direction, unifying the two aspects of learning the \glsfirst{bs} and exploiting any discovered reward.
Nonetheless, the path towards a solution of the sparse rewards problem is still long.
Many improvements and extensions can be developed to augment the ability of \gls{name}, and \gls{qd} algorithms in general, to better address this problem.
At the same time, while framing the problem of sparse rewards from an \gls{rl} point of view, we did not take advantage of the many innovations introduced in the field of \gls{rl}.
We believe that a lot can be learned from the union of the two fields of \gls{rl} and divergent search \glspl{ea}, thanks to the complementarity of the two approaches.
This could lead to the development of many innovative approaches, the solution of many of the still unresolved problems in both fields and the application of learning algorithm to new and exciting fields of research.

\clearpage\null\thispagestyle{empty}

\chapter*{Bibliography}
\addstarredchapter{Bibliography}
\bibbysegment[heading=subbibliography]

\appendix
\include{tex/appendix}

\backmatter
\include{tex/thanks}

\end{document}